\documentclass[acmtog]{acmart}

\usepackage{booktabs} 

\makeatletter
\def\runningfoot{\def\@runningfoot{}}
\def\firstfoot{\def\@firstfoot{}}
\makeatother
\renewcommand\footnotetextcopyrightpermission[1]{} 
\setcitestyle{square}

\usepackage[ruled]{algorithm2e} 

\SetAlFnt{\small}
\SetAlCapFnt{\small}
\SetAlCapNameFnt{\small}
\SetAlCapHSkip{0pt}
\IncMargin{-\parindent}

\setcopyright{none}

\makeatletter
\newcommand\footnoteref[1]{\protected@xdef\@thefnmark{\ref{#1}}\@footnotemark}
\makeatother

\usepackage{steinmetz}
\usepackage{physics}
\usepackage{graphicx}
\usepackage{amssymb}
\usepackage{gensymb}
\usepackage{mathtools}
\usepackage{multirow}
\usepackage[english]{babel} 

\begin{document}

\title{Learning to Predict Indoor Illumination from a Single Image}

\author{Marc-Andr\'e Gardner}
\affiliation{Universit\'e Laval}
\email{marc-andre.gardner.1@ulaval.ca}
\author{Kalyan Sunkavalli}
\author{Ersin Yumer}
\author{Xiaohui Shen}
\author{Emiliano Gambaretto}
\affiliation{Adobe Research}
\author{Christian Gagn\'e}
\author{Jean-Fran\c cois Lalonde}
\affiliation{Universit\'e Laval}

\renewcommand\shortauthors{}

\begin{abstract}
We propose an automatic method to infer high dynamic range illumination from a single, limited field-of-view, low dynamic range photograph of an indoor scene.
In contrast to previous work that relies on specialized image capture, user input, and/or simple scene models, we train an end-to-end deep neural network that directly regresses a limited field-of-view photo to HDR illumination, without strong assumptions on scene geometry, material properties, or lighting. We show that this can be accomplished in a three step process: 1) we train a robust lighting classifier to automatically annotate the location of light sources in a large dataset of LDR environment maps, 2) we use these annotations to train a deep neural network that predicts the location of lights in a scene from a single limited field-of-view photo, and 3) we fine-tune this network using a small dataset of HDR environment maps to predict light intensities. This allows us to automatically recover high-quality HDR illumination estimates that significantly outperform previous state-of-the-art methods. Consequently, using our illumination estimates for applications like 3D object insertion, produces photo-realistic results that we validate via a perceptual user study.


\end{abstract}

%

%
%
%
%

%
%


\begin{teaserfigure}
\centering
\setlength{\fboxsep}{0pt}%
\setlength{\fboxrule}{1pt}%
  \includegraphics[width=0.24\textwidth]{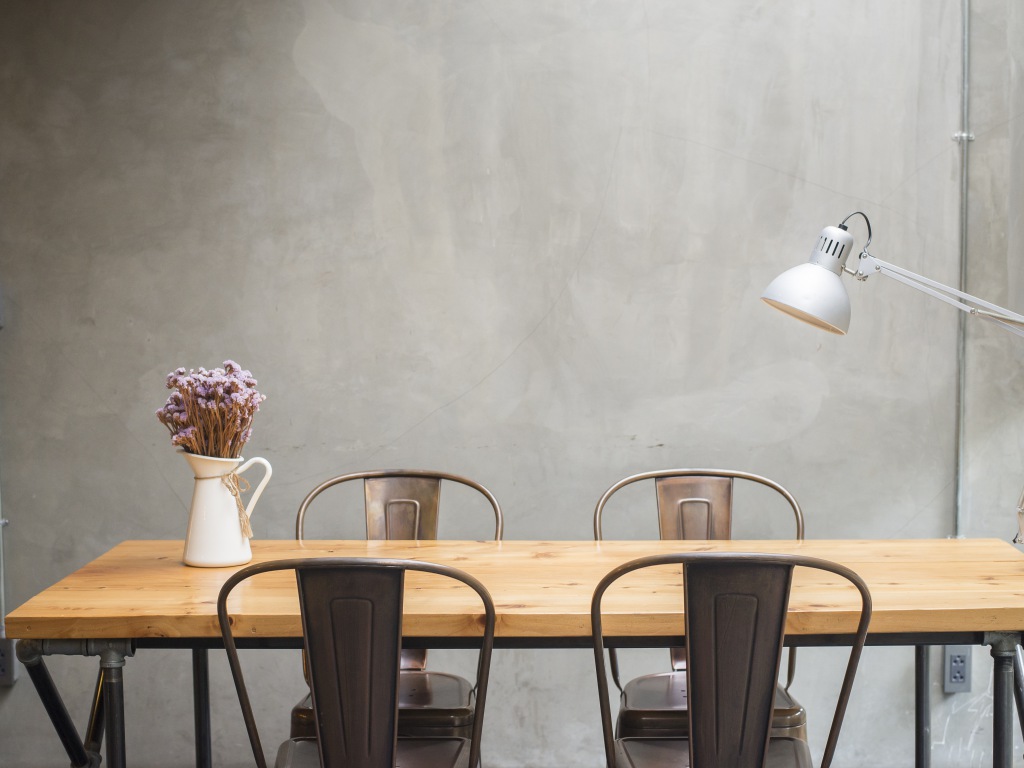}
	\begin{picture}(0,0)
		\put(-126,63){\fcolorbox{white}{white}{\includegraphics[height=1cm]{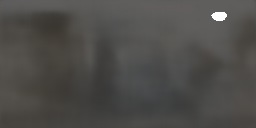}}}
	\end{picture}
  \includegraphics[width=0.24\textwidth]{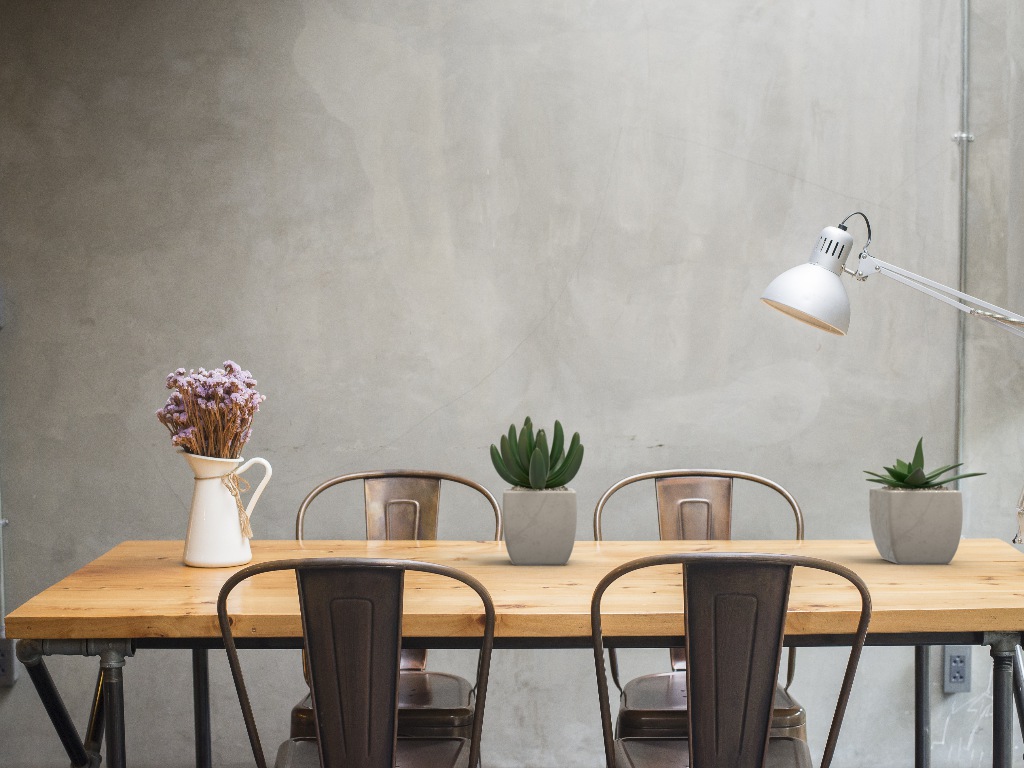} 
	\hspace{0.2cm}
  \includegraphics[width=0.24\textwidth]{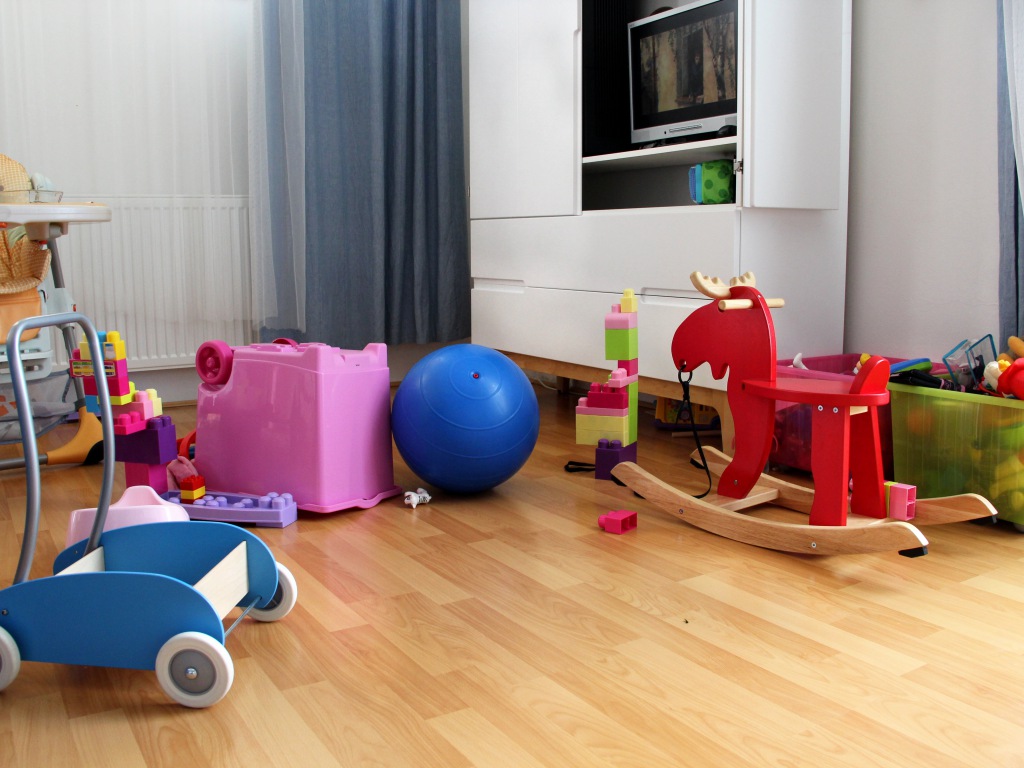} 
	\begin{picture}(0,0)
		\put(-126,63){\fcolorbox{white}{white}{\includegraphics[height=1cm]{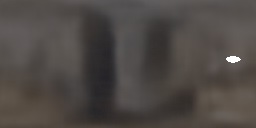}}}
	\end{picture}
  \includegraphics[width=0.24\textwidth]{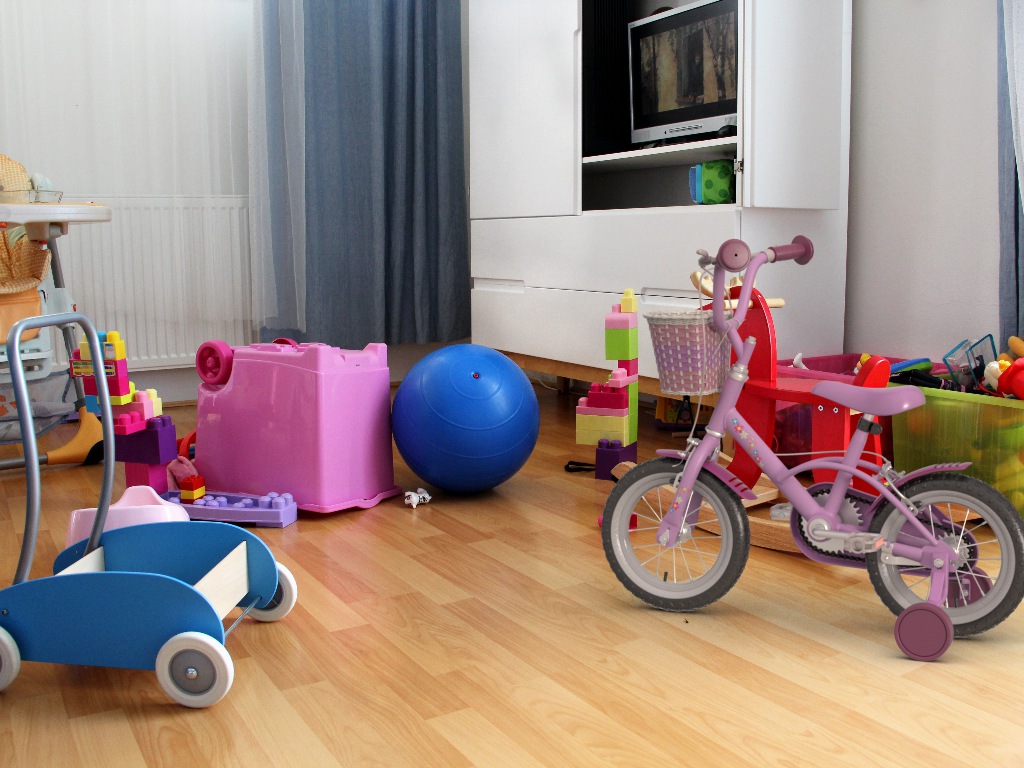}
\caption{Given a single LDR image of an indoor scene, our method automatically predicts HDR lighting (insets, tone-mapped for visualization). Our method learns a direct mapping from image appearance to scene lighting from large amounts of real image data; it does not require any additional scene information, and can even recover light sources that are not visible in the photograph, as shown in these examples. Using our lighting estimates, virtual objects can be realistically relit and composited into photographs.}
\label{f:teaser}
\end{teaserfigure}
 
\maketitle

\vspace{15em}
\section{Introduction}
\label{sec:introduction}

\begin{figure}[!t]
\centering
\includegraphics[height=2.5cm]{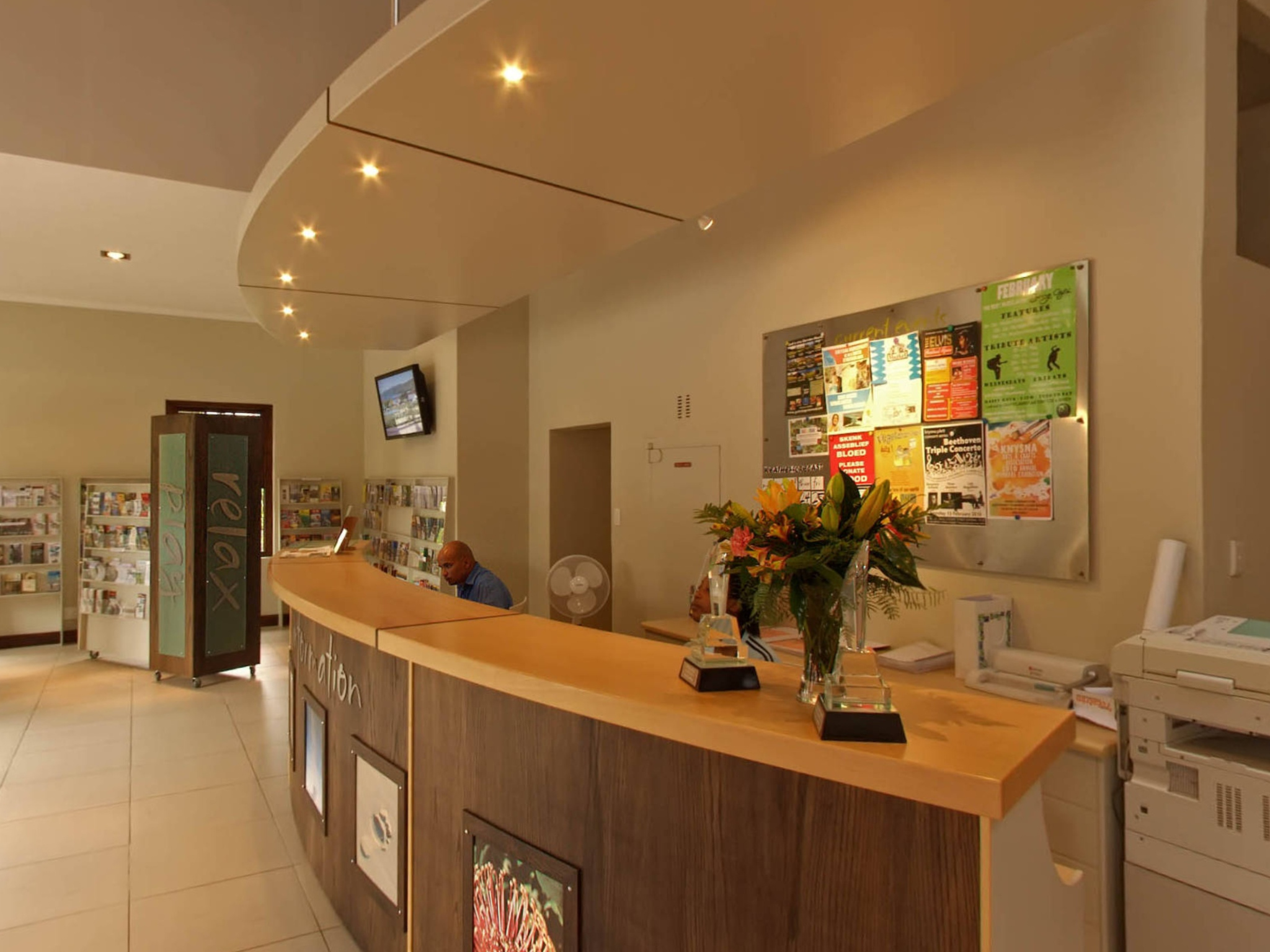}
\includegraphics[height=2.5cm]{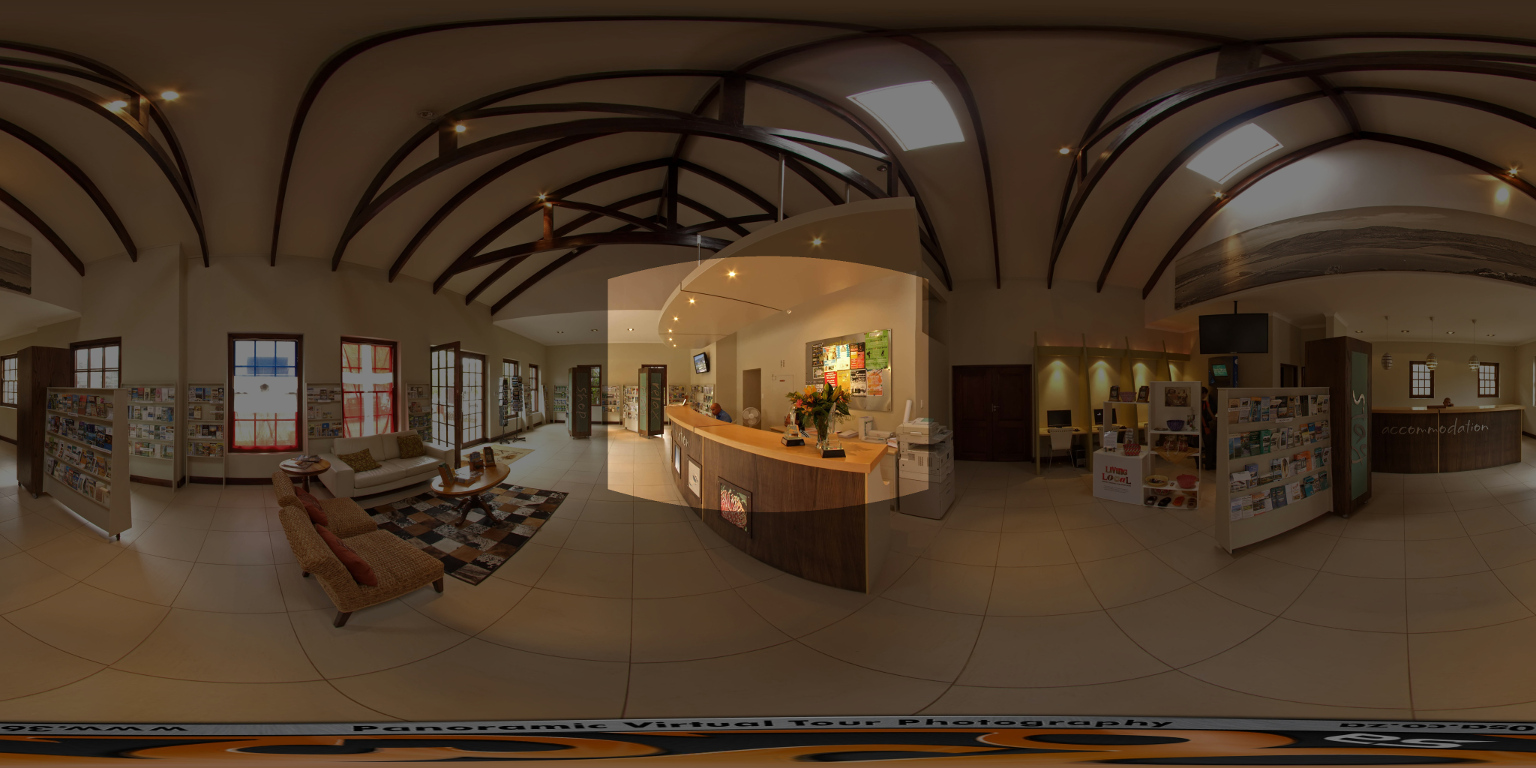}
\caption{Extracting crops from panoramas: (left) normal lens crop image, (right) spherical panorama.}
\label{f:makethepoint}
\vspace{-.5em}
\end{figure}

Inferring scene illumination from a single photograph is a challenging problem. The pixel intensities observed in an image are a complex function of scene geometry, materials properties, illumination, the imaging device, and subsequent post-processing. Disentangling one of these factors from another is an ill-posed inverse problem. This is especially hard from a \emph{single limited field-of-view image}, since many of the factors that contribute to the scene illumination are not even directly observed in the photo (Fig.~\ref{f:makethepoint}). This problem is typically addressed in two ways: first, by assuming that scene geometry (and/or reflectance properties) is given (either measured using depth sensors, reconstructed using other methods, or annotated by a user), and second, by imposing strong low-dimensional models on the lighting (e.g., low-frequency spherical harmonics). 

While we have made significant progress in single-image geometric reconstruction~\cite{eigen-iccv-15,bansal2016marr} and reflectance estimation~\cite{bell2015minc,zhou2015intrinsic}, state-of-the-art techniques are still significantly error-prone. These errors can then propagate into lighting estimates when they are directly used in a rendering-based optimization. Fundamentally, indoor lighting varies widely in its geometric and photometric properties; for example, the same scene can have large windows, bright spot lights, and diffuse lamps. This wide range of illuminants typically cannot be accurately represented by low-dimensional lighting models. 

This paper proposes a method to infer high dynamic range (HDR) illumination from a single, limited field-of-view, low dynamic range (LDR) photograph of an indoor scene. Our goal is to be able to model the range of typical indoor light sources, and choose a spherical environment map (or IBL) representation that is often used to represent real-world illumination~\cite{debevec-sig-98}. We also want to make this inference robust to errors in geometry, surface reflectance, and scene appearance models. To this end, we introduce an end-to-end learning based approach, that takes images as input and predicts illumination using deep neural networks.

Deep neural networks have been successfully applied to closely related problems such as depth estimation~\cite{eigen-iccv-15,bansal2016marr}, reflectance map estimation~\cite{rematas-cvpr-16}, and intrinsic images~\cite{zhou2015intrinsic}. In the vein of this previous work, we propose training a deep neural network to learn a representation for image appearance in terms of illumination. However, training such a network would require a large database of image-HDR illumination pairs. Such a dataset does not currently exist and would require a significant amount of time and effort to assemble. Instead, we resort to the large database of 360 LDR panoramas of Xiao et al.~\shortcite{xiao-cvpr-12}. However, using LDR panoramas as training data poses an additional challenge: the light sources are not explicitly available. Hence, we also introduce a method to detect light sources on a given LDR panorama image, which significantly outperforms the state-of-the-art. We use the results of this algorithm as the output for the training pairs, and images cropped from the corresponding panoramas as the input. This allows us to predict the location of light sources in a scene, but not the intensities of the lights since this information is not accurately captured in LDR panoramas. We resolve this by capturing a new dataset consisting of 2100 HDR environment maps. Fine-tuning the network trained on LDR panoramas using this HDR data allows us to train a network that directly regresses an LDR, limited field-of-view photo to the true HDR scene illumination. 

Unlike previous work, our technique does not require special image capture or user input. Nor does it rely on any assumptions on scene appearance, geometry, material properties or lighting. Instead, we can automatically recover illumination estimates from images of indoor scenes truly captured ``in the wild''. Our work significantly outperforms previous state-of-the-art methods. Consequently, using these illumination estimates for applications like 3D object insertion lead to results that are photo-realistic (e.g. Fig.~\ref{f:teaser}). We demonstrate this over a large set of examples and via a perceptual user study that indicates that objects lit by our illumination estimates are almost indistinguishable from those lit by ground truth illumination. We believe that this represents a significant step forward on an important, and challenging scene understanding problem.

\paragraph{Contributions} Our main contributions are:

1. An end-to-end illumination estimation method that leverages a deep convolutional network to take a limited-field-of-view image as input and produce an estimation of HDR illumination.

2. A state-of-the-art light source detection method for LDR panoramas and a panorama warping method, that help generate training data for the end-to-end illumination estimation network.

3. A new HDR environment map dataset that can be used to train and evaluate illumination estimation or other scene inference tasks.

4. A benchmarking of the state-of-the-art in single image scene illumination estimation by means of a perceptual user study.









\section{Related work}

Reconstructing a scene---and all its properties including geometry, surface reflectance, and illumination---from a single image is one of the long-standing goals of computer vision and graphics, and has been extensively studied in the literature. Because these properties are intrinsically tied to each other, estimating each of them often relies on reconstructing the others too~\cite{barron-pami-15}. In this related work, we will specifically focus on techniques for recovering illumination.

The seminal work of Debevec~\shortcite{debevec-sig-98} on image-based lighting demonstrated that capturing several photographs of a mirrored sphere using different exposures can be used to reconstruct a physi\-cally-correct, omnidirectional HDR radiance map, which can then be used to realistically render novel objects into the scene. Follow-up work has demonstrated that the same can be done from a single shot, provided there is also a diffuse sphere in the scene~\cite{reinhard-book-10}, or a metallic/diffuse hybrid sphere~\cite{debevec-sslp-12}.

\begin{figure*}[!t]
\includegraphics[width=0.94\linewidth]{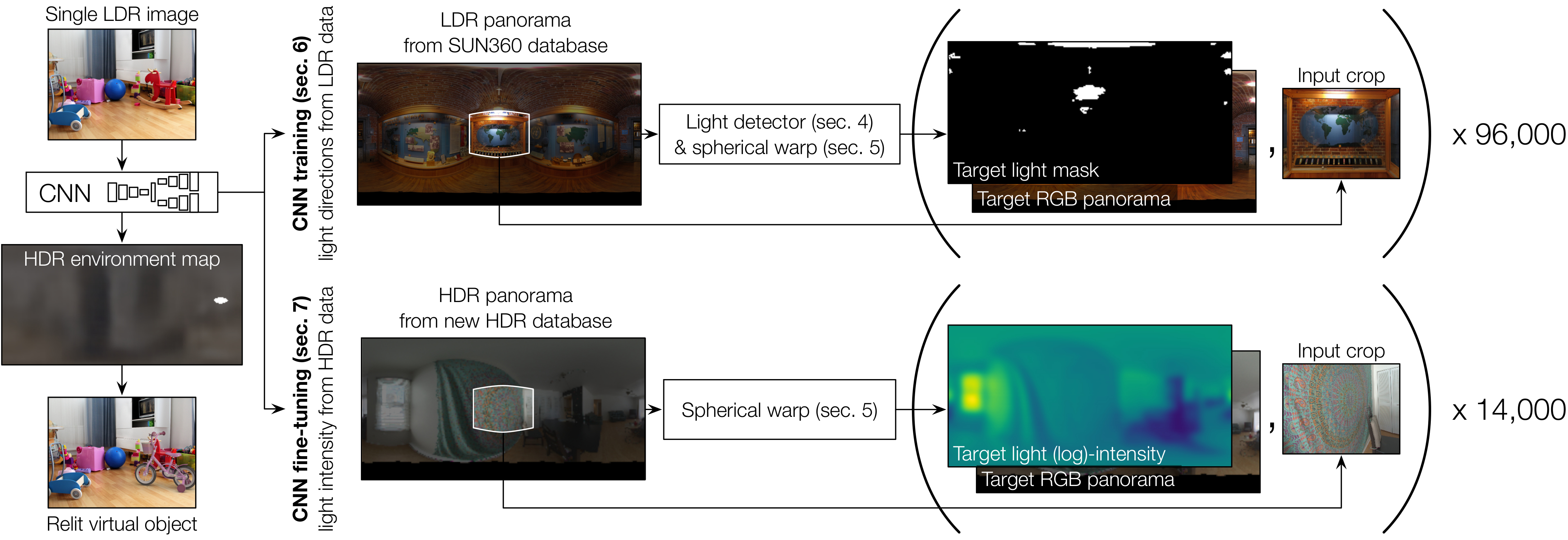}
\caption{Overview of the paper. Our method automatically predicts the HDR lighting conditions from a single photograph (left). To do so, it relies on a deep CNN that is trained in two stages. First, we rely on a database of LDR panoramas~\cite{xiao-cvpr-12}. To compensate for the low dynamic range, light sources are detected and the panoramas are warped to generate target light masks, which, combined with crops extracted from the panoramas, can be used to train the CNN to predict light directions. Second, the network is fine-tuned on a novel dataset of HDR panoramas, which allows it to learn to predict light intensities.}
\label{f:overview}
\end{figure*}

Previous work on illumination estimation typically models scene appearance as a function of the scene geometry, reflectance properties, and illumination, and optimizes for values that best explain the captured image. This appearance model is typically Lambertian shading under a low-dimensional lighting model like low-order spherical harmonics. This is combined with known geometry captured using depth sensors~\cite{barron2013rgbd} or reconstructed using multi-view stereo~\cite{wu-cvpr-11} or model-based fitting~\cite{valgaerts-tog-12} to estimate lighting. Moreno et al.~\shortcite{moreno-cng-10} assume that the lighting consists of a known set of discrete point lights. These techniques recover illumination from isolated objects~\cite{lombardi2016reflectance}, for which a low-dimensional lighting model is sufficient and the geometry and reflectance can be well modeled. However, we are interested in recovering illumination from indoor scenes that exhibit complex cluttered geometry, spatially-varying non-Lambertian appearance, and a wide variety of lighting conditions that cannot be be parameterized accurately with low-dimensional models. Therefore, we use a more general environment map representation and directly regress it from the image without relying on potentially inaccurate appearance models.

Lalonde et al.~\shortcite{lalonde-ijcv-10} focus on recovering outdoor illumination that is well approximated by analytical sun-sky models. Hold-Geoffroy et al.~\shortcite{holdgeoffroy-cvpr-17} propose a deep learning-based method to predict outdoor illumination from a single input image. Like us, they train their network with image/illumination pairs created from a panorama database. However, outdoor illumination is distant and well approximated by low-dimensional analytical models. As a result, they are able to fit the $3$-parameter Ho\v{s}ek-Wilkie model~\shortcite{Hosek2012} to LDR panoramas to recover HDR illumination, and train a deep neural network to predict these $3$ parameters. In contrast, lighting in indoor scenes is complex and cannot be approximated well with simple low-dimensional models. Instead, we use a non-parametric HDR IBL representation and this makes learning indoor illumination challenging in terms of both data generation and network training. In this work, we address these challenges with a number of novel contributions. We propose a two-stage training process that initially learns to predict light locations from LDR data, and is later fine-tuned to predict light intensities on an HDR dataset. To generate data for the light location prediction task, we develop a state-of-the-art light detector that we use to annotate a large LDR panorama dataset. We also propose a panorama warping step that accounts for spatially-localized indoor lighting (vis-a-vis distant outdoor illumination). We also present a new multi-head neural network to recover indoor illumination and train it with a novel rendering loss that progressively improves its light prediction estimates. 

Khan et al.~\shortcite{khan-siggraph-06} propose flipping an HDR input image to approximate the out-of-view illumination. Similar ideas have also been used in other 2D compositing techniques~\cite{bitouk-sig-08,lalonde-sig-07}. While these approximations might work in some cases, they can be completely incorrect in many others (for example, when the dominant light illuminating the scene is behind the camera). In contrast, our light predictions are significantly more accurate and only require an LDR image as input.

Karsch et al.~\shortcite{karsch-sig-11} estimate scene illumination from single images, but rely on user input to annotate geometry and initial lighting, which is then refined using a rendering-based optimization. 
Zhang et al.~\shortcite{zhang-siga-16} use a similar scheme to recover illumination from a complete RGBD scan of a scene and user-annotated light positions. Karsch et al.~\shortcite{karsch-tog-14} propose an automatic scene inference technique. They detect light sources visible in the image, and leverage the SUN360 panorama database to predict out-of-view lighting. This is done by finding panoramas that are similar in appearance to the input image and using pre-classified light sources in these panoramas as the light sources for the input image. This transforms the illumination estimation problem into one of image matching with the right metric; however, this is a coarse approximation, and in many cases, the matched panorama may have lighting that is arbitrarily different from the actual illumination in the image. In contrast, we propose directly learning the mapping between image appearance and scene illumination, and demonstrate that this leads to better illumination estimates.


More recently, deep neural network-based techniques have been proposed for estimating reflectance maps---the convolution of a surface BRDF with the incident illumination---from a single image~\cite{rematas-cvpr-16}. These reflectance maps can then be separated into reflectance and illumination estimates~\cite{georgoulis-arxiv-16a}. However, these techniques focus on objects of a known class with approximately known shape and spatially constant reflectance. In contrast, the indoor scenes we focus on have significantly more complex shape and reflectance variations. CNN-based methods have also been proposed for estimating scene depth~\cite{eigen-iccv-15}, surface normals~\cite{eigen-iccv-15,bansal2016marr}, and intrinsic decompositions for indoor scenes~\cite{zhou2015intrinsic}. While challenging, these problems involve recovering properties that are directly observed in the image, unlike lighting which can lie out of the field-of-view and only indirectly effects scene appearance. By recovering illumination from a single image, our work can likely benefit these other scene inference tasks.

\section{Method overview}

Our goal, illustrated in fig.~\ref{f:overview}, is to predict the HDR lighting conditions from a single photograph. If we cast our goal as a learning problem, training data in the form of $\{\textrm{photo}, \textrm{HDR light probe}\}$ pairs would be ideal for the learning task. Given sufficient data, one could try to regress the HDR light probe directly from the photo. However, such data does not exist currently, and gathering it in sufficient quantity is prohibitively expensive. On the other hand, large datasets of LDR panoramas already exist~\cite{xiao-cvpr-12}, and since they capture the entire environment of the scene, they can potentially be used to learn illumination. To generate input data for training, we extract rectified crops from these panoramas at various orientations and focal lengths, and attempt to learn the relationship between the crops and the panoramas (see fig.~\ref{f:overview}). 


Unfortunately, we cannot directly learn indoor lighting from LDR panoramas since their low dynamic range does not capture lighting properly~\cite{debevec-siggraph-97}. In addition, indoor illumination tends to be localized, i.e., the light sources cannot be assumed to be directional (as is the case outdoors~\cite{holdgeoffroy-cvpr-17}). The center of projection of the panorama can be arbitrarily far away from a point in the cropped scene; as a result, the true illumination incident at this point is, at the very least, a \emph{warped} version of the panorama. 

Given these limitations, we propose using the LDR data to train a network to identify the \emph{location} of light sources in the scene. We do so using two novel, practical solutions to deal with the aforementioned issues. First, we introduce in sec.~\ref{sec:lightdetection} a robust method to detect light sources in LDR panoramic images. Second, we introduce a method to warp the panorama such that it better reflects the lighting conditions at the cropped scene (sec.~\ref{sec:warping}). 

These solutions allow us to use a large dataset of LDR panoramas~\cite{xiao-cvpr-12} to learn to predict indoor lighting from photographs. To this end, we introduce in sec.~\ref{sec:learning} an end-to-end convolutional neural network which produces a binary light mask indicating the light positions, as well a low-resolution RGB approximation of the entire panorama. Since using a standard loss function to learn a binary light mask heavily penalizes even small shifts of a light source position, we introduce an in-network, differentiable cosine filter which enables efficient learning. 

In addition to estimating the \emph{positions} of light sources, we need to estimate the \emph{intensities} of these lights to recover complete scene illumination. To achieve this, we captured a new dataset of $2,100$ high-quality HDR environment maps spanning a wide range of scenes and illumination conditions. While this dataset is too small to train a network from scratch, we show that it is sufficient to fine-tune our pre-trained light position prediction network to infer light intensities (sec.~\ref{sec:hdr}). The fine-tuned network produces a light intensity map and RGB panorama that can be combined to create a final HDR environment map (sec.~\ref{sec:experiments}), which in turn can be used to relight virtual objects into the input photo.


\section{LDR panorama light source detection}
\label{sec:lightdetection}

In order to use LDR panoramas for training our CNN to detect light sources, we must first detect areas in the panoramas which correspond to bright light sources. To do so, we propose a novel light source detector, and show that it significantly outperforms the approach of Karsch et al.~\shortcite{karsch-tog-14}. 

\subsection{Light classification} 

After converting to grayscale, the panorama $P$ is rotated by $90^\circ$ about the pitch angle to yield $P_\text{rot}$, so that the zenith is aligned with the horizon line. This rotation is needed to account for the large distortions caused by the equirectangular projection, which severely stretches regions around the poles. Features are then computed over $P$ and $P_\text{rot}$ separately on square patches at five different scales\footnote{We use $30\times30$ squares at the lowest scale, multiplying their size by 1.5 at each scale.}. In particular, we use HOG~\cite{dalal-cvpr-05}, the mean patch elevation, as well as its mean, standard deviation, and 99th percentile intensity values. These features are used to train two logistic regression classifiers for small (e.g. spotlights and lamps) and large (e.g. windows, reflections) light sources. We found that training classifiers for these two types of classes separately yielded better performance, as these types of light sources significantly differ from one another. 

The resulting logistic regression classifiers are then applied in a sliding-window fashion over $P$ and $P_\text{rot}$ to yield a score at each pixel. 
Scores from both classifiers are added, then merged on a per-pixel manner according to their elevation angles $S_\text{merged} = S\cos(\theta) + S_\text{rot}^*\sin(\theta)$, where $S$ indicates the regression scores, $\theta$ the pixel elevation, and $S_\text{rot}^*$ is $S_\text{rot}$ rotated back to the original orientation. The resulting scores are then thresholded to obtain a binary mask, refined with a dense CRF~\cite{krahenbuhl-nips-12}, and adjusted with opening and closing morphological operations. The optimal threshold is obtained by maximizing the intersection-over-union (IoU) score between the resulting binary mask and the ground truth labels on the training set. 

\begin{figure}
    \centering
    \includegraphics[width=0.64\columnwidth]{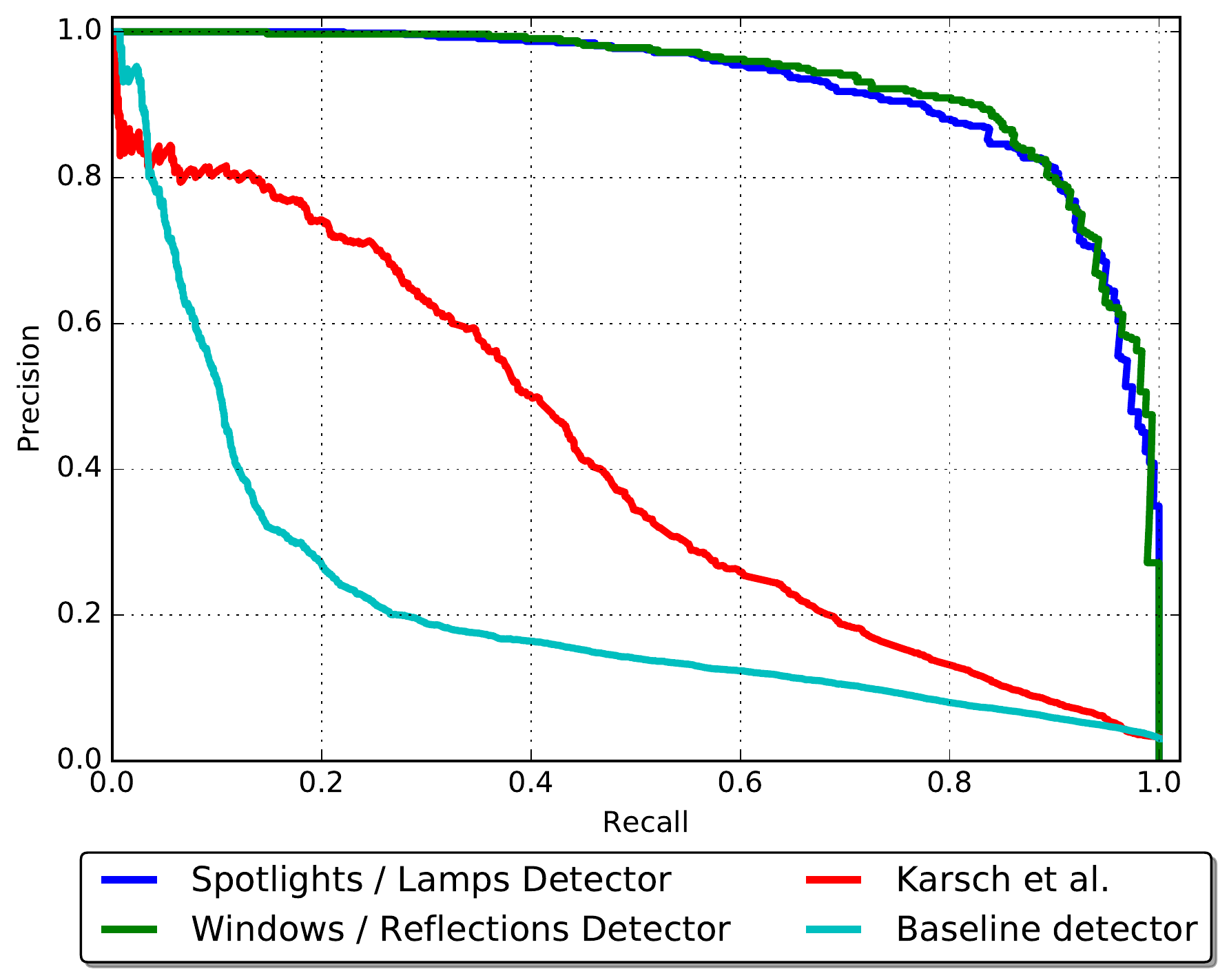}
    \caption[]{Precision-recall curves for the light detector on the test set for our detectors and the one of Karsch et al.~\shortcite{karsch-tog-14}. In blue, the curve for the spotlights and lamps detection, and in green, the curve for the windows and light reflections. In red, the result for \cite{karsch-tog-14}. In cyan, the curve for a baseline detector relying solely on the intensity of a pixel. Note that because of the inherent uncertainty of the importance of a light (including reflections) relative to the others (even for a human annotator), a perfect match between human and algorithm predictions is highly unlikely.}
    \label{fig:prcurves}
\end{figure}

\begin{figure}[t]
\centering
\setlength{\tabcolsep}{1pt}
\begin{tabular}{cc}
\includegraphics[width=0.48\linewidth]{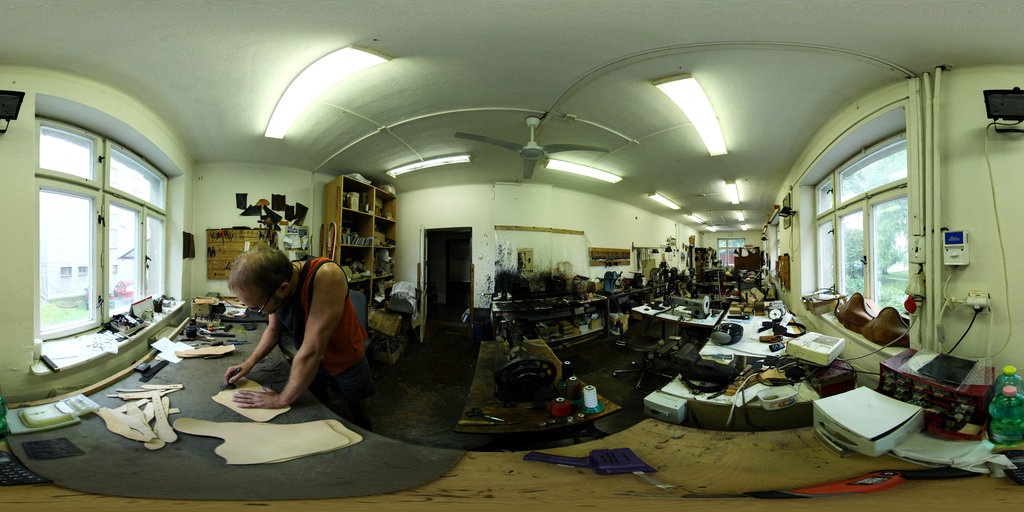} & 
\includegraphics[width=0.48\linewidth]{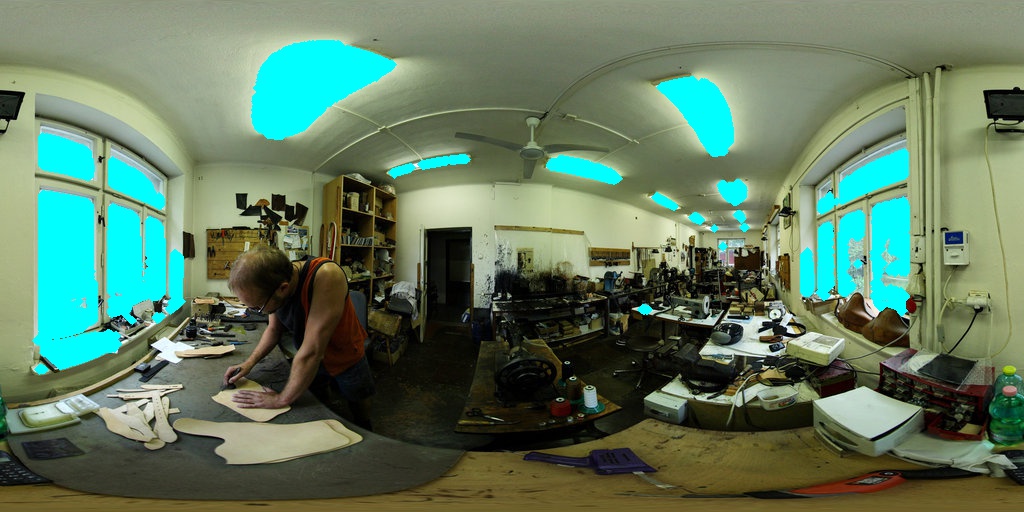} \\
\includegraphics[width=0.48\linewidth]{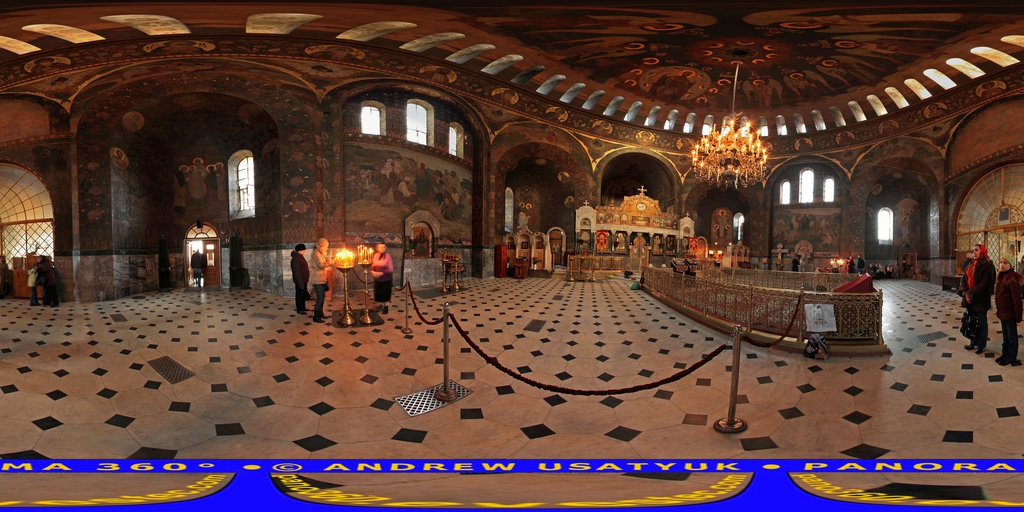} & 
\includegraphics[width=0.48\linewidth]{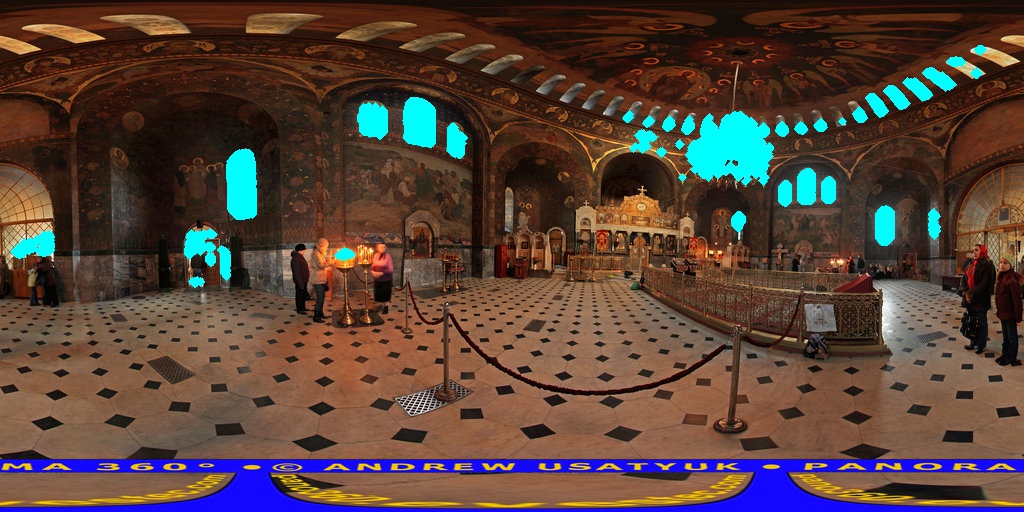} \\
\includegraphics[width=0.48\linewidth]{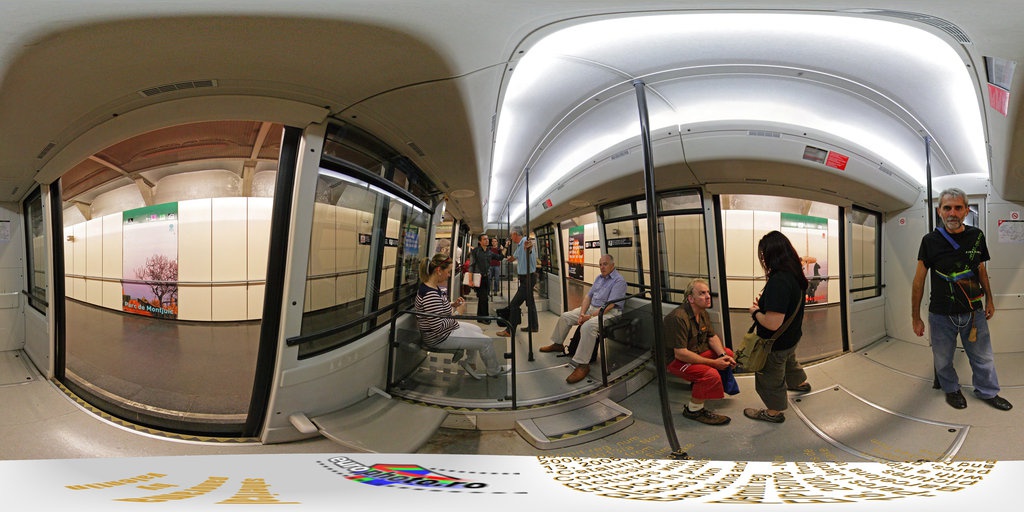} & 
\includegraphics[width=0.48\linewidth]{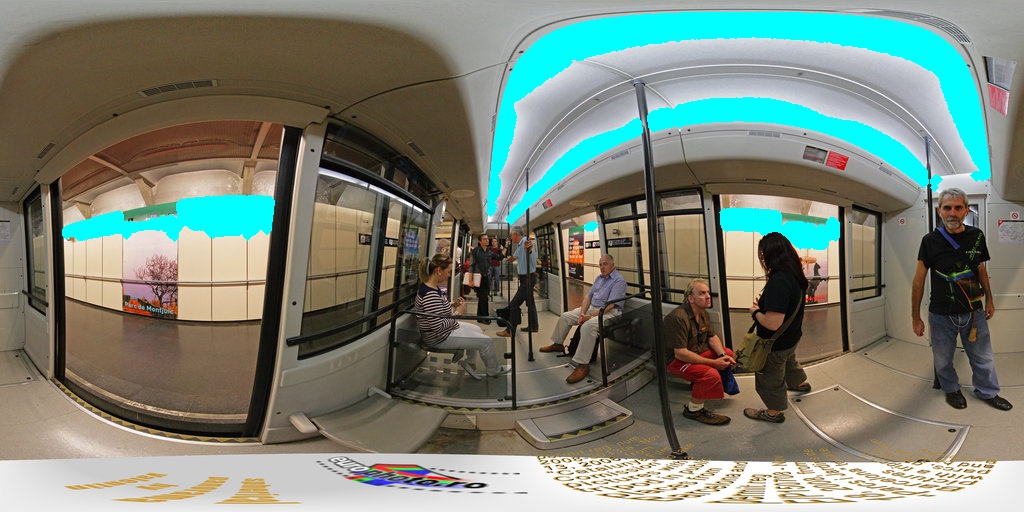} \\
\includegraphics[width=0.48\linewidth]{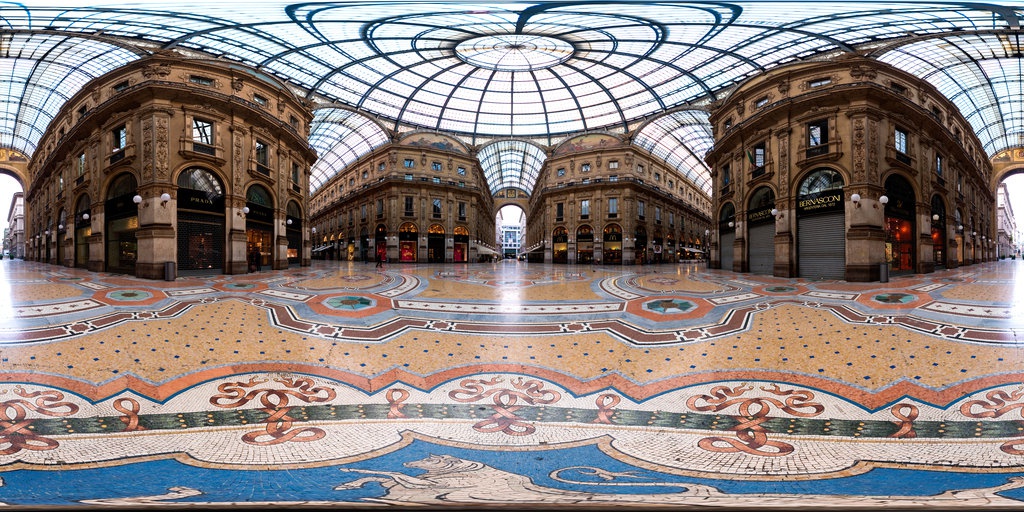} & 
\includegraphics[width=0.48\linewidth]{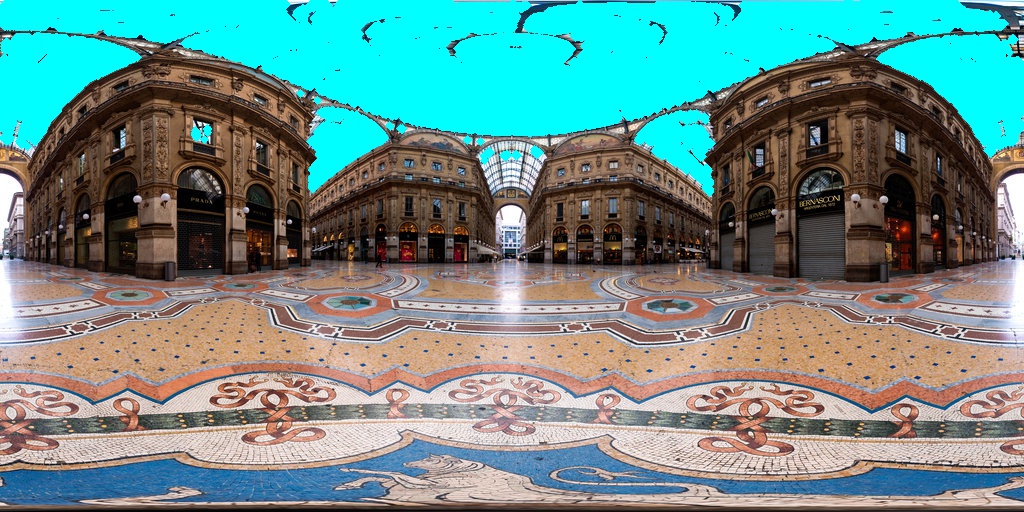} \\
\includegraphics[width=0.48\linewidth]{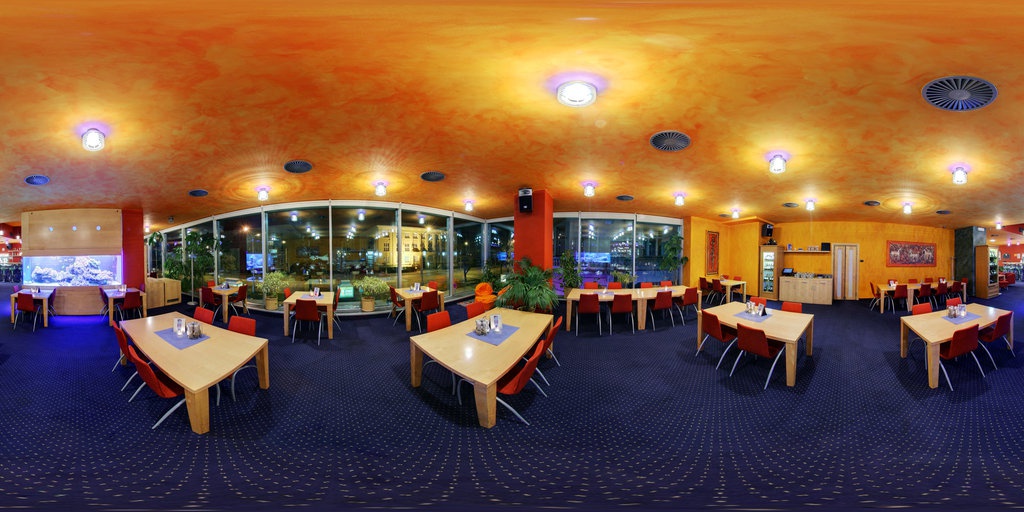} & 
\includegraphics[width=0.48\linewidth]{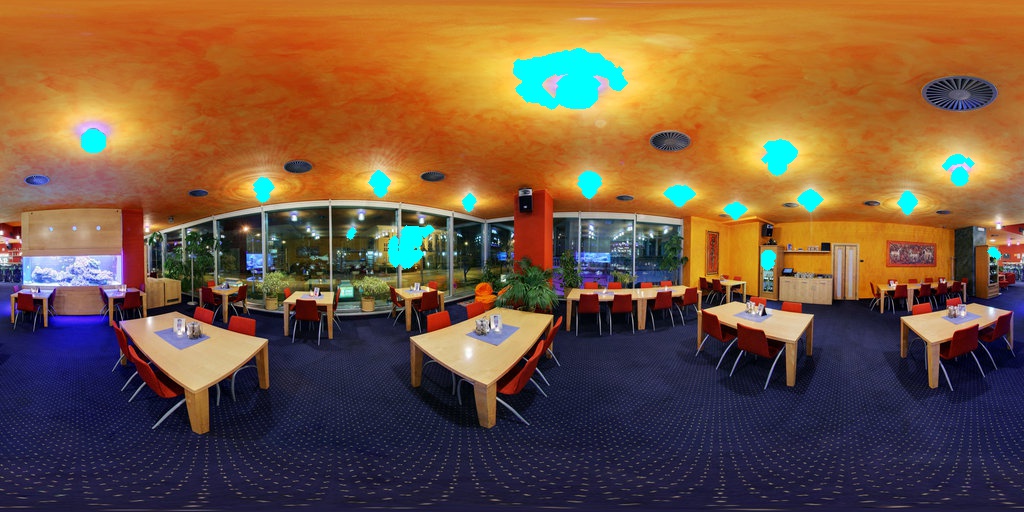} \\
\end{tabular}
\caption{Light detection results on SUN360 panoramas. (left) the input LDR panoramas; (right) light detection results, shown in cyan and overlaid on the original panorama for reference. The detector is able to handle a wide range of lighting arrangements, including large light patches and spotlights.}
\label{f:lightdetection-results}
\end{figure}

\subsection{Training details and evaluation} 

To train the classifiers, we manually annotate a set of 400 panoramas from the SUN360 database. Four types of light sources are labelled: spotlights, lamps, windows, and (bounce) reflections. We use 80\% of the panoramas for training, and 20\% for testing. The classifier is first trained using labeled lights as positive samples and random negative samples. Subsequently, hard negative mining~\cite{felzenszwalb-pami-10} is used over the entire training set. We discard the bottom 15\% of the panoramas because this region often contain watermarks and light sources are seldom located below the camera. 

Fig.~\ref{fig:prcurves} reports a comparison of precision-recall curves for our two detectors, a baseline method which directly maps the intensity of a pixel to its probability of belonging to a light source, and the approach of Karsch et al.~\shortcite{karsch-tog-14}. As expected, the baseline performs poorly on LDR data like SUN360. The detector from \cite{karsch-tog-14} offers better performance, but our performs significantly better at any level of recall. Fig.~\ref{f:lightdetection-results} shows light detection results on example panoramas from the SUN360 dataset.

\section{Panorama recentering warp}
\label{sec:warping}

\begin{figure}[!t]
\centering
\includegraphics[width=0.99\linewidth]{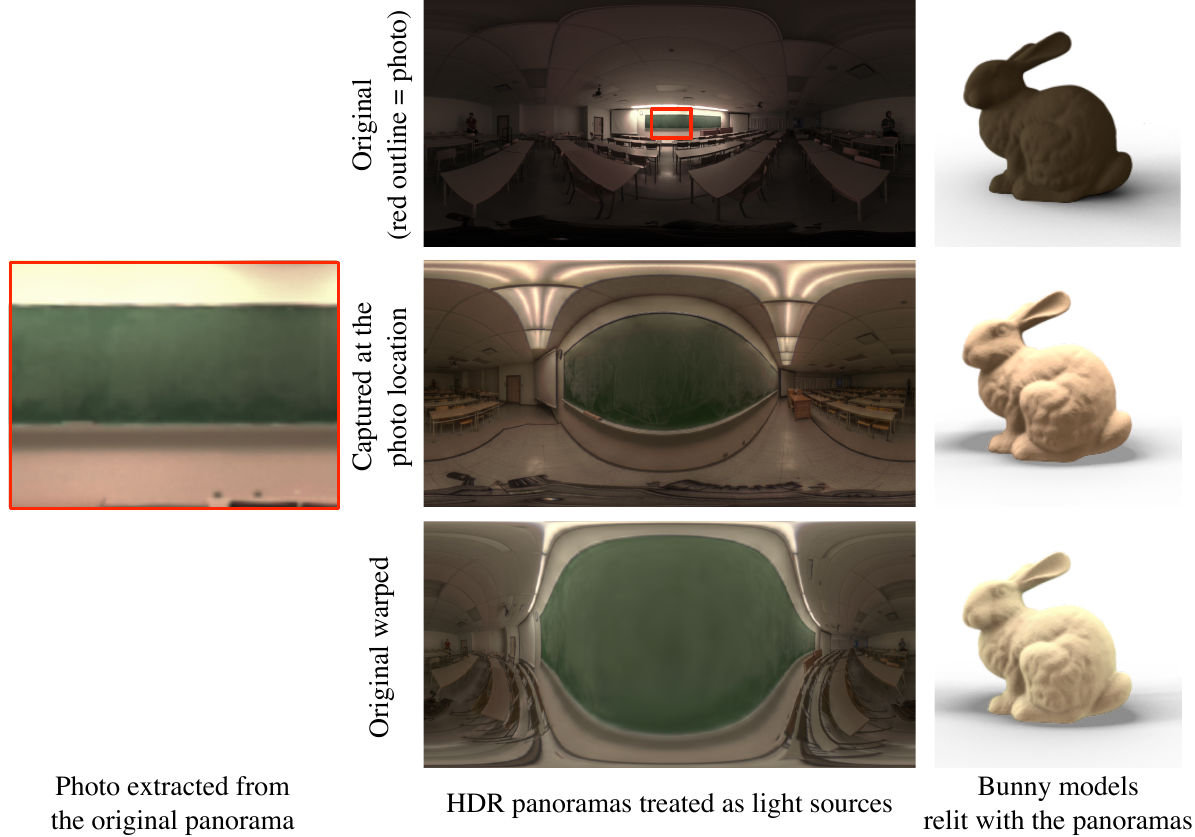}
\caption{The importance of light locality for indoor scenes. Left, a photo for which we want to estimate the lighting conditions. The photo was cropped from the ``original'' panorama (top row, middle). Treating this panorama as the light source for the photo is wrong; its center of projection is in front of the scene in the photo, and relighting a virtual bunny (top row, right) makes it appear to be backlit. The correct HDR panorama, captured with a light probe at the position of the cropped photo, is shown in the middle row, and captures the location of the lights on top of the scene. We introduce a warping operator that can be estimated with no scene information, and distorts the original panorama to approximate the location of the light sources on the top (bottom row). Relighting an object with the warped panorama yields results that are much closer to the ground truth.}
\label{f:warping-problem}
\end{figure}

Detecting the light sources in LDR panoramas is not sufficient for training the CNN to learn lighting from a single photo. The fundamental problem is that the panorama does not represent the lighting conditions in the \emph{cropped scene}, since the panorama center of projection can be arbitrarily far away from the location of the scene points in the cropped photo. Fig.~\ref{f:warping-problem} illustrates this issue. The photo shown on the left was cropped from the ``original'' panorama in the middle column. Treating this original panorama as a light source is incorrect, and results in a backlit bunny. We captured the \emph{actual} lighting conditions by placing a light probe at the scene (middle column of fig.~\ref{f:warping-problem}). Notice how the lighting conditions at the scene differ from those in the original panorama. To allow the use of the SUN360 database (from which we can crop photos but do not have access to the scenes to capture ground truth lighting) for training, we present a novel method that warps the original panorama to approximate the lighting in the cropped photo (bottom row). 

\subsection{Warping operator}

The aim of the warping operator is to generate the panorama that would be captured by a virtual camera placed at a point in the cropped photo. This is a challenging problem that is made especially harder by the fact that we do not know the scene geometry, and we make two assumptions to make this task feasible. First, we assume that the scene lies on a sphere, i.e., all scene points are equidistant from the original center of projection. Second, we assume that an image warping suffices to model the effect of moving the camera, i.e., occlusions are not an important factor. These assumptions may not hold for all scene points; however, note that our goal is to model light sources, which are typically located at scene extremities (ceiling, walls, etc.) and are better approximated by these assumptions.

Let us assume that the panorama is placed on the unit sphere, i.e. $x^2 + y^2 + z^2 = 1$, with the camera that captured this panorama at the origin of this sphere. The outgoing rays emanating from a virtual camera placed at $(x_0,y_0,z_0)$, can be parameterized as:
\begin{equation}
x(t) = v_x t + x_0  \quad
y(t) = v_y t + y_0  \quad
z(t) = v_z t + z_0  \,.
\label{eq:warp2}
\end{equation}
Intersecting these rays with the panorama sphere yields:
\begin{equation}
    (v_x t + x_0)^2 + (v_y t + y_0)^2 + (v_z t + z_0)^2 = 1 \,.
    \label{eq:warp3}
\end{equation}

\begin{figure}[!t]
    \centering
    \includegraphics[width=0.375\linewidth]{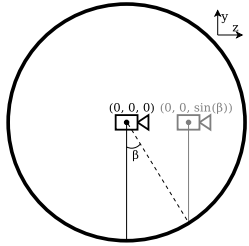}
    \caption{Overview of the warping problem, illustrated in 2D for simplicity. The circle represents a slice of the spherical panorama along the $y$--$z$ plane, with the center of projection (illustrated by a camera) at its center. The aim of the warp operator is create a virtual center of projection with a nadir at an angular distance of $\beta$ with respect to the original nadir. The angle $\beta$ corresponds to the point in the panorama where the photo is extracted.}
    \label{f:warp-basics}
\end{figure}

As illustrated in fig.~\ref{f:warp-basics}, we want to model the effect of using a virtual camera whose nadir is at $\beta$. The angle $\beta$ corresponds to the point in the panorama where the photo is extracted, and we will discuss how this is computed shortly. For the case of translating along the $z$-axis, this results in a new camera center, $\{x_0, y_0, z_0\}$ = $\{0, 0, \sin \beta\}$. Warping in arbitrary directions can trivially be achieved by rotating the environment map before and after the warp. Substituting this in eq.~\ref{eq:warp3}, results in the following second degree equation:
\begin{equation}
(v_x^2 + v_y^2 + v_z^2)t^2 + 2v_z t \sin\beta  + \sin^2\beta - 1 = 0 \,.
\label{eq:warp4}
\end{equation}
Solving (\ref{eq:warp4}) for $t$ (keeping only positive solutions, as negative roots represent the intersection on the other side of the sphere), maps the coordinates from the original environment map to the ones in the warped camera coordinate system. 

\begin{figure}[!t]
\centering
\footnotesize
\setlength{\tabcolsep}{1pt}
\begin{tabular}{cc}
\includegraphics[width=0.493\linewidth]{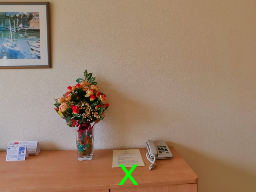} &
\includegraphics[width=0.493\linewidth]{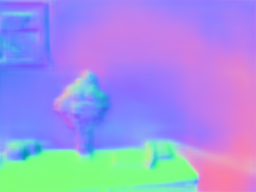} \\
(a) Input image & (b) Normals \\
\includegraphics[width=0.493\linewidth]{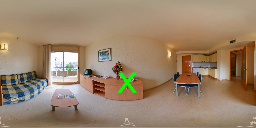} &
\includegraphics[width=0.493\linewidth]{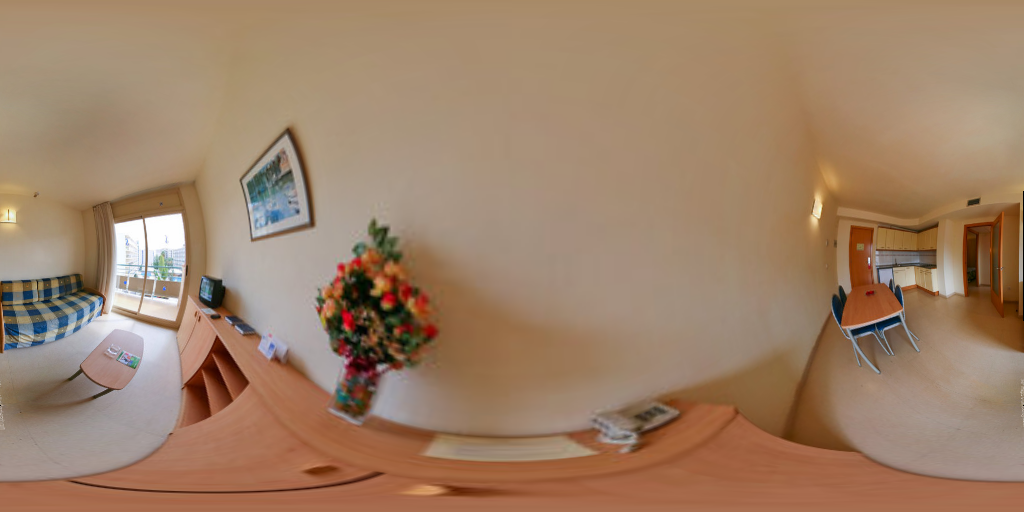} \\
(c) Original panorama & (d) Warped panorama \\
\end{tabular}
\caption{$\beta$ selection procedure. From a given crop picture (a), we extract the normals using the method of  Bansal et al.~\shortcite{bansal2016marr} (b). We pick the insertion point by looking at the lowest pixel with a horizontal surface (green X in (a)) and backproject it on to the panorama (c). This gives us the point where we would like the nadir to be, from which $\beta$ can be trivially recovered. We then warp the panorama using this $\beta$ (d).}
\label{f:warp-beta-pick}
\end{figure}

The value of $\beta$ in eq.~(\ref{eq:warp4}) represents the point in the panorama where the photo is extracted. We expect that users will want to insert objects on to flat horizontal surfaces in the photo, and we reflect this in the choice of $\beta$ as follows (see fig.~\ref{f:warp-beta-pick}): we first use the approach of Bansal et al.~\shortcite{bansal2016marr} to detect surface normals in the cropped image, and find flat surfaces by thresholding based on the angular distance between surface normal and the up vector. We back-project the $y$-coordinate of the lowest point of the largest flat area (i.e., the lowest point on the flattest horizontal surface) on to the panorama to obtain $\beta$. In cases where no horizontal surfaces are found (e.g., a flat vertical wall), no warp is applied as the panorama is assumed to be sufficiently close to scene. Note that we always assume the insertion point to be x-centered ---that is, we do not ask the network to estimate the light at far-left or far-right of the image.


\begin{figure}[!t]
\centering
\footnotesize
\setlength{\tabcolsep}{1pt}
\begin{tabular}{ccc}
\includegraphics[width=0.325\linewidth]{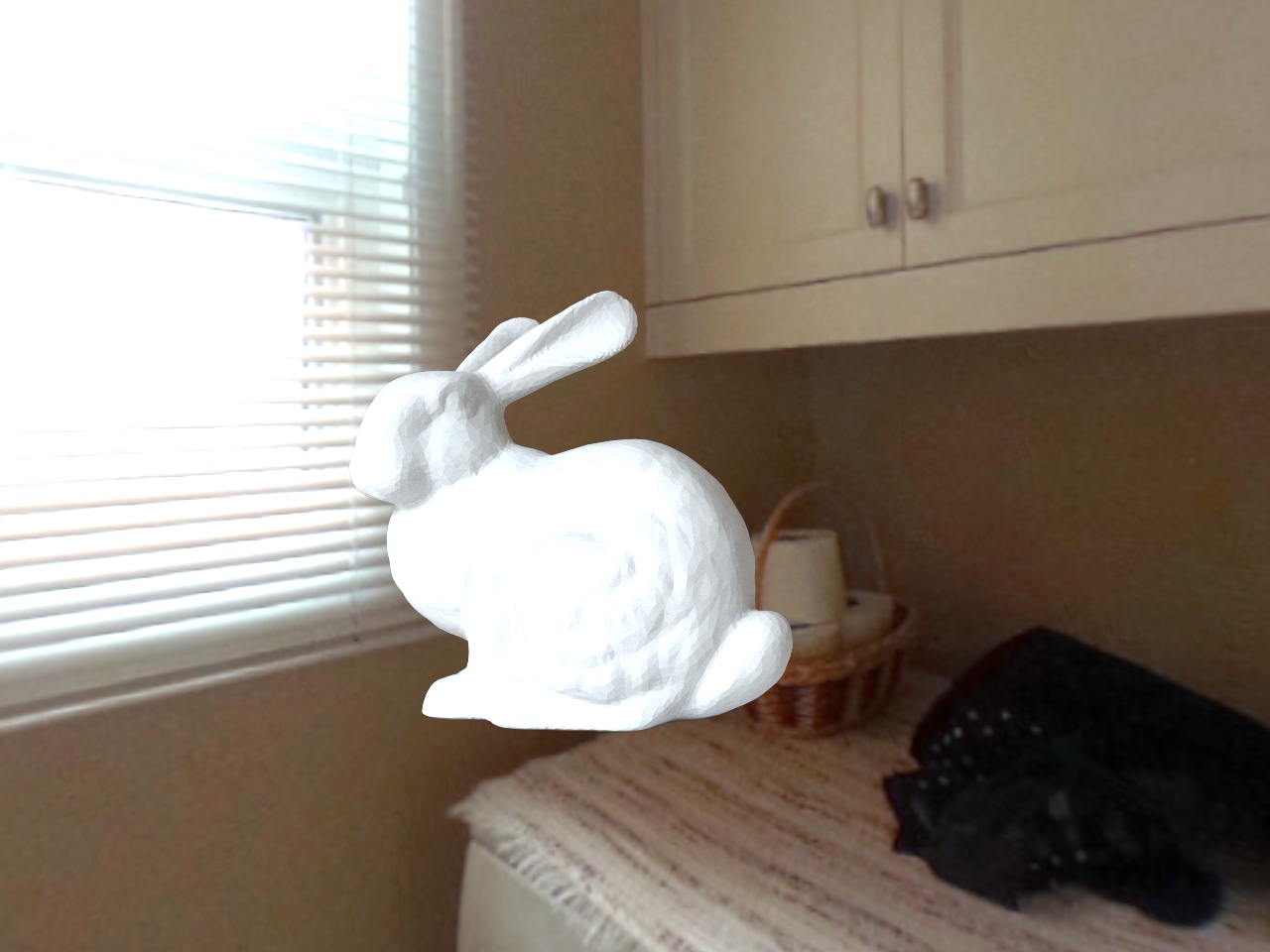} &
\includegraphics[width=0.325\linewidth]{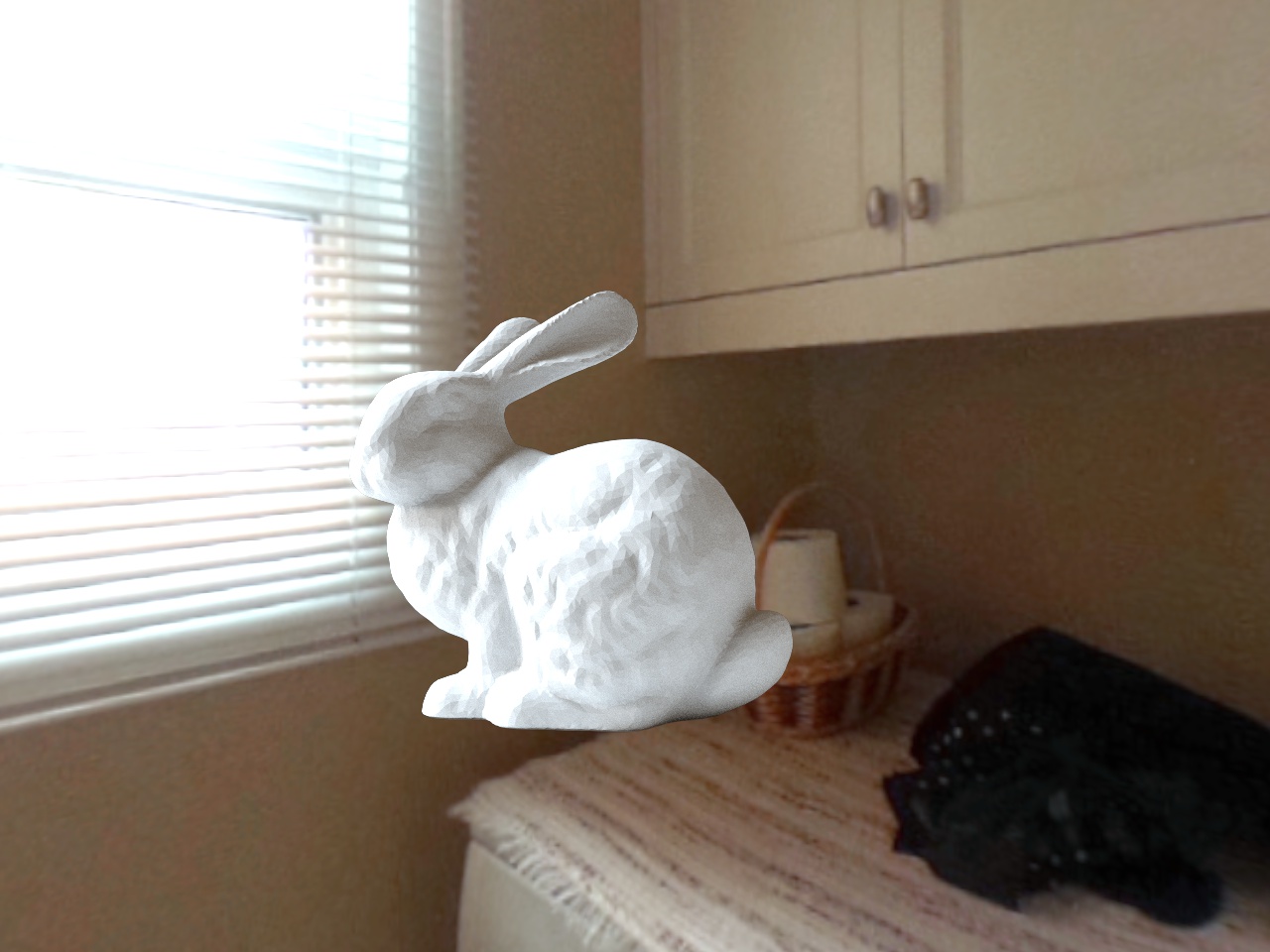} & 
\includegraphics[width=0.325\linewidth]{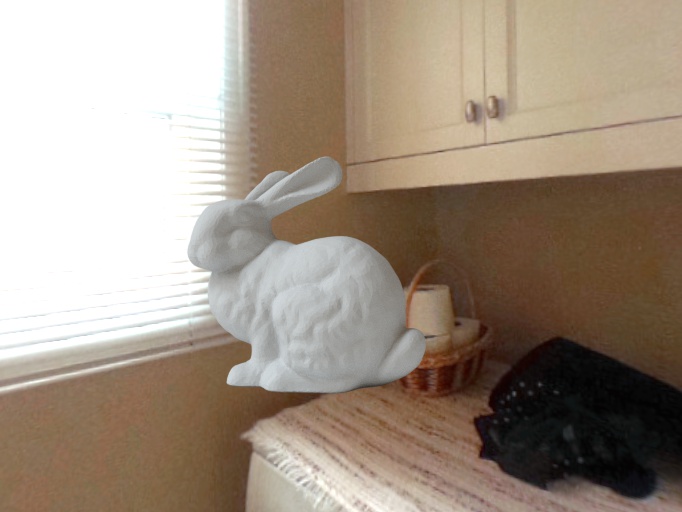} \\
\includegraphics[width=0.325\linewidth]{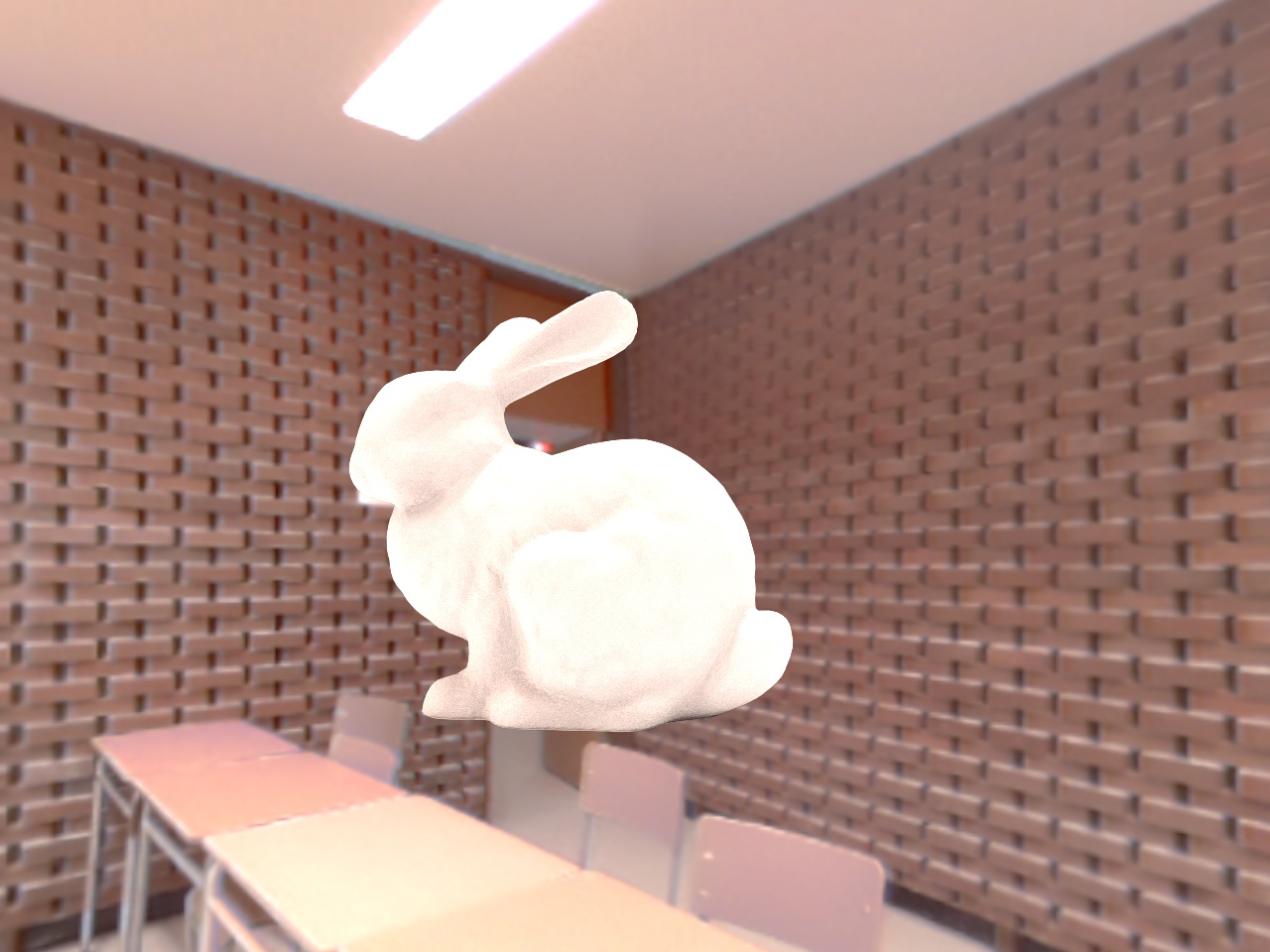} &
\includegraphics[width=0.325\linewidth]{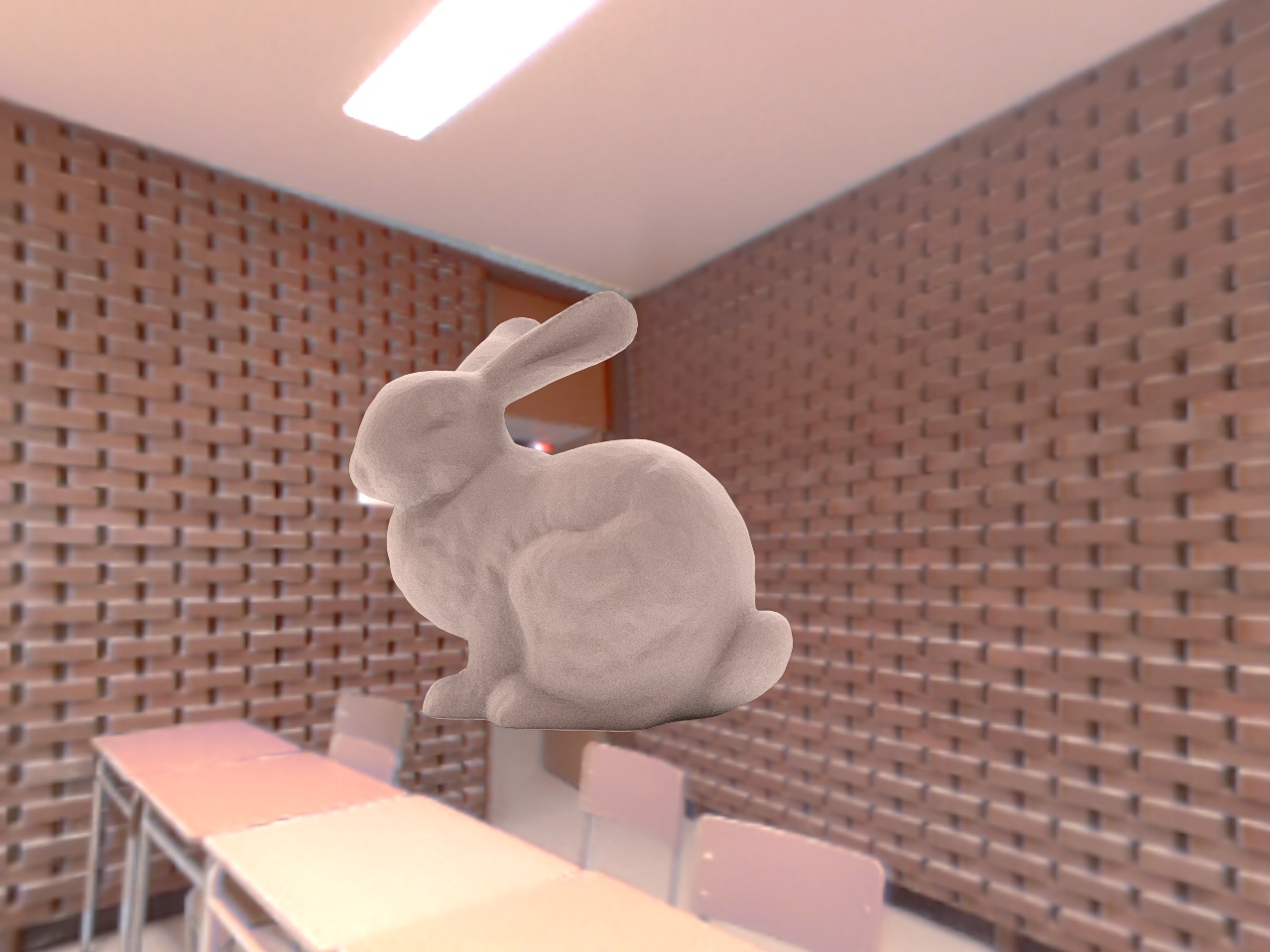} &
\includegraphics[width=0.325\linewidth]{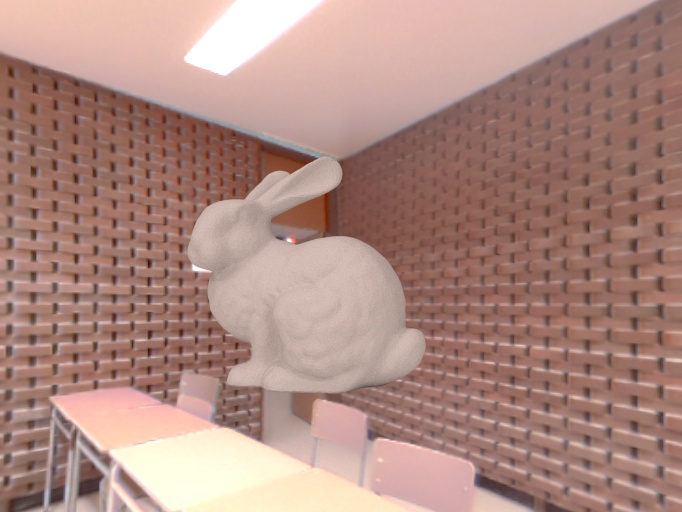} \\
\includegraphics[width=0.325\linewidth]{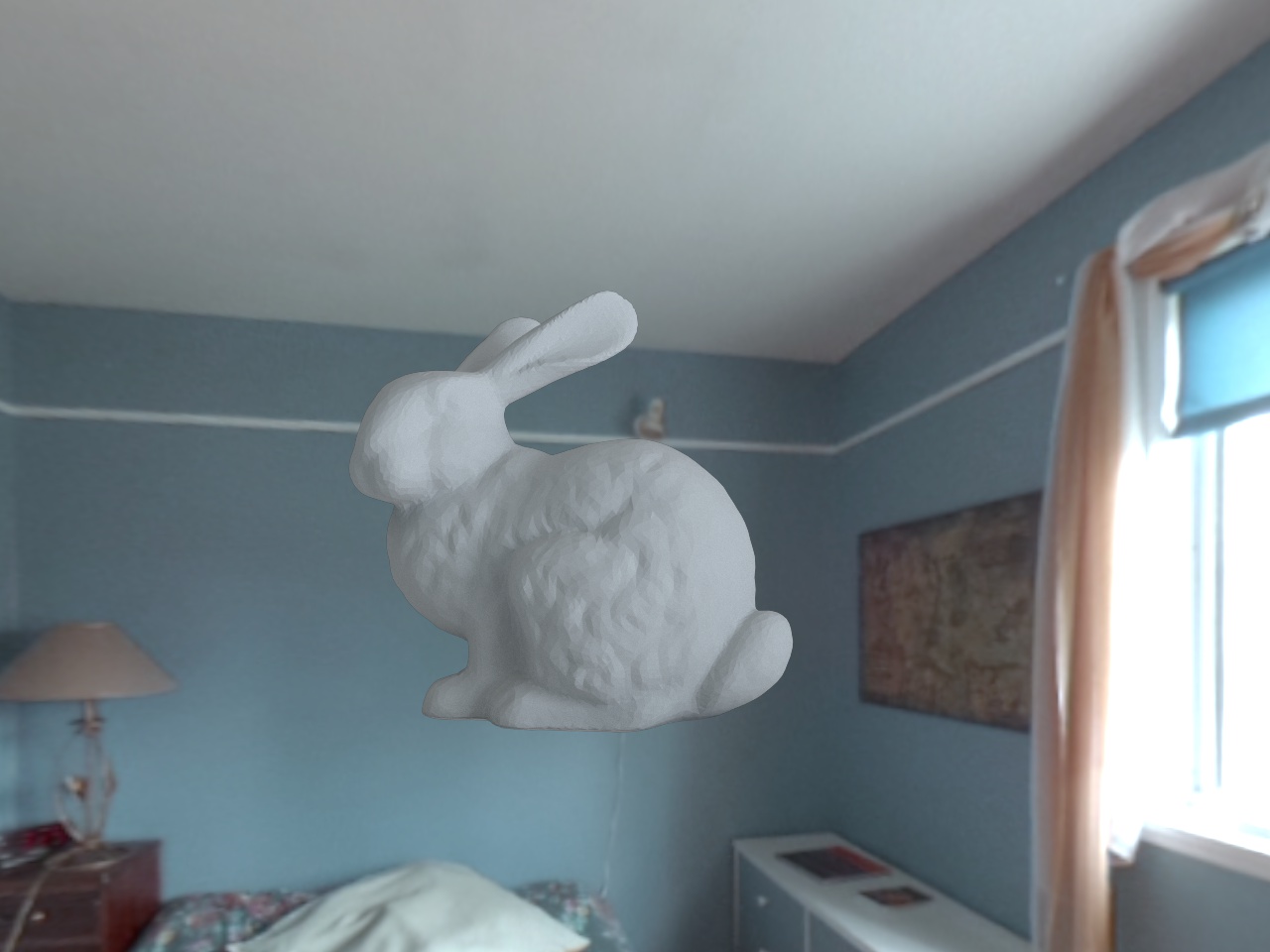} &
\includegraphics[width=0.325\linewidth]{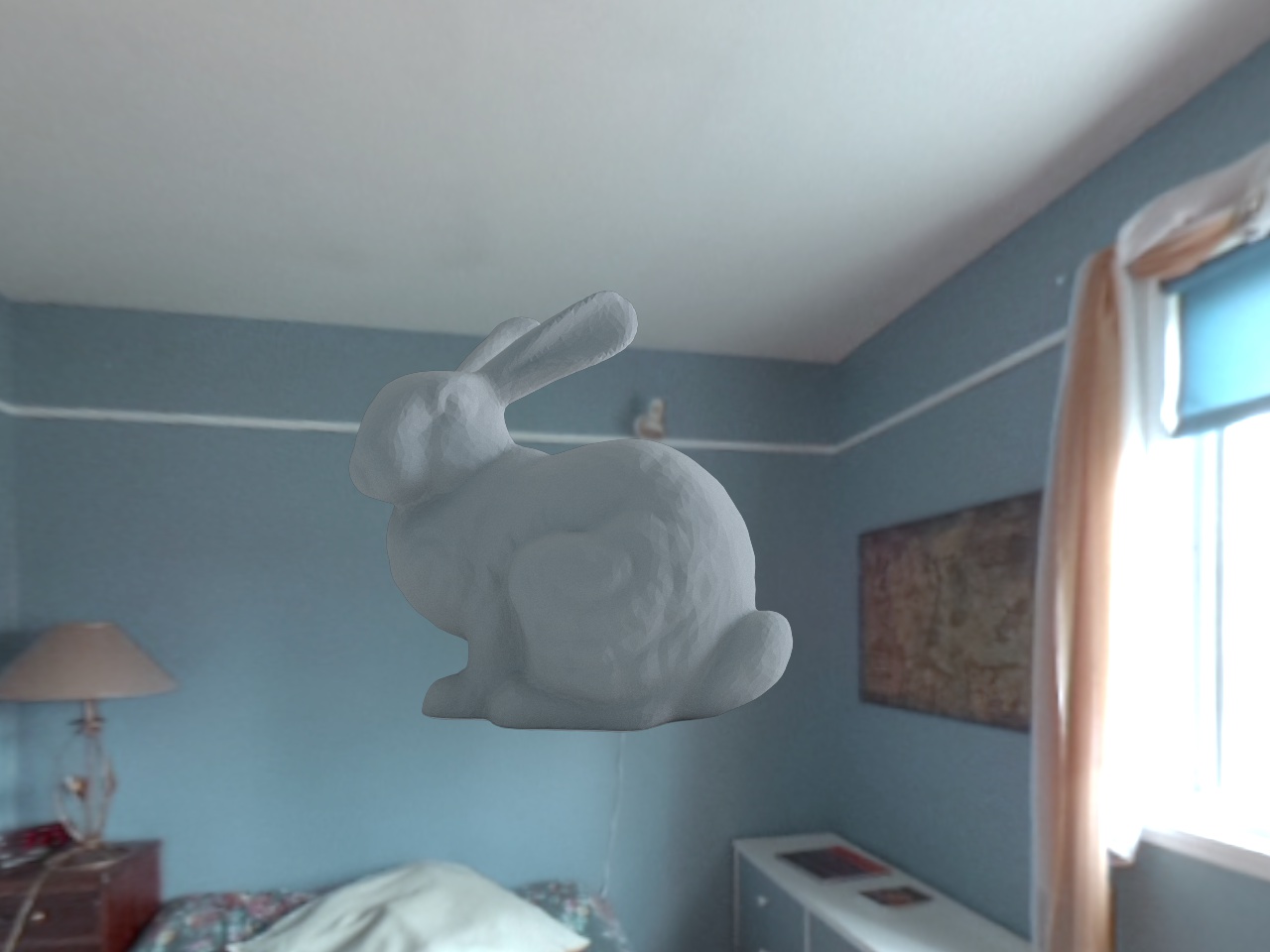} & 
\includegraphics[width=0.325\linewidth]{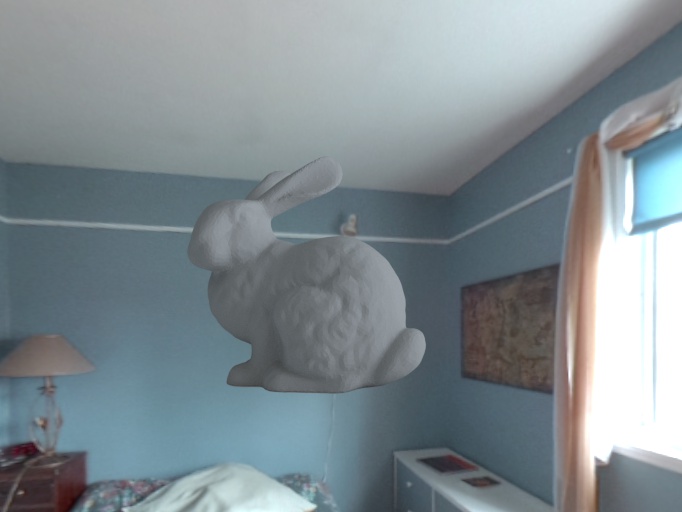} \\
\includegraphics[width=0.325\linewidth]{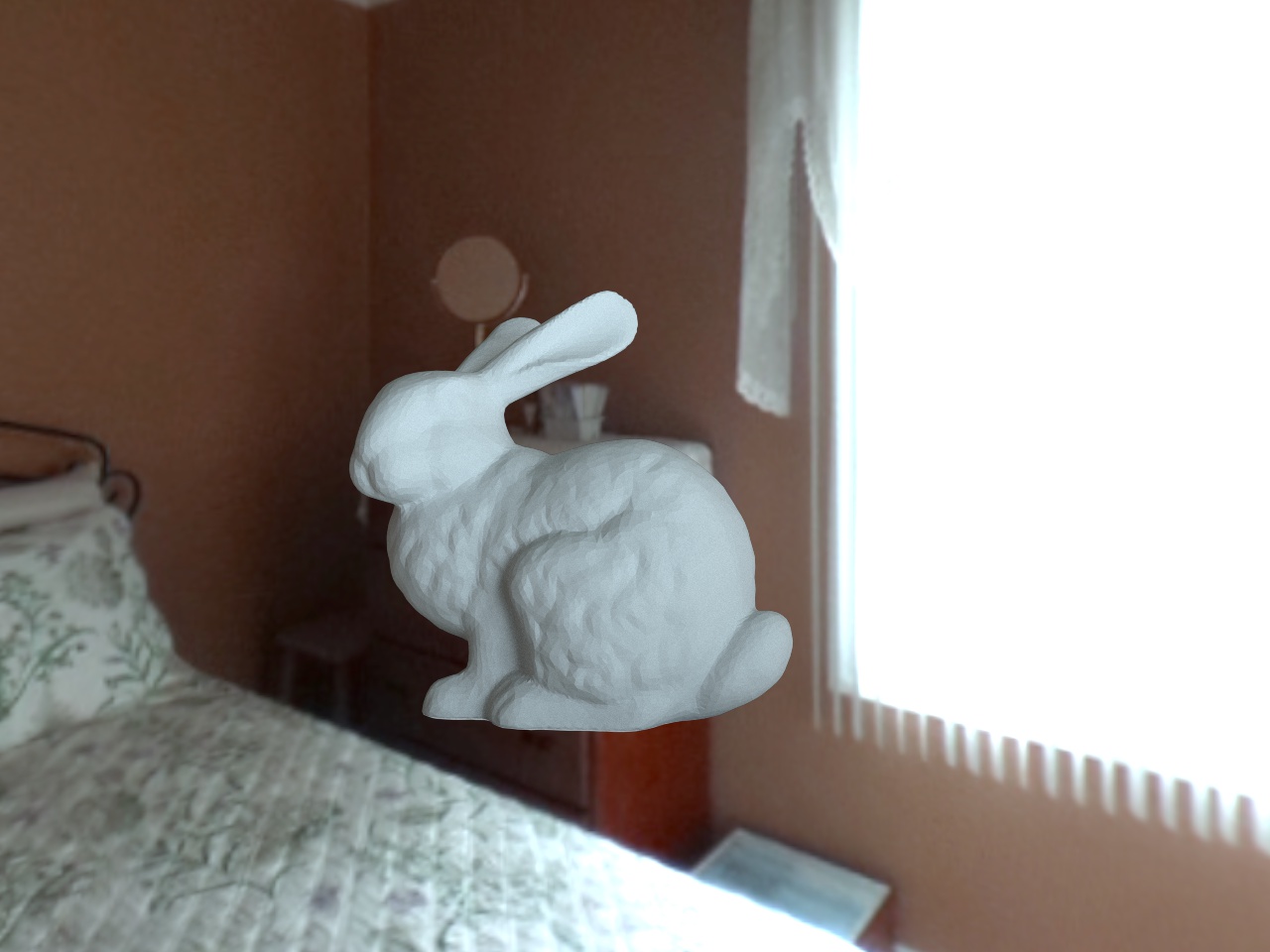} &
\includegraphics[width=0.325\linewidth]{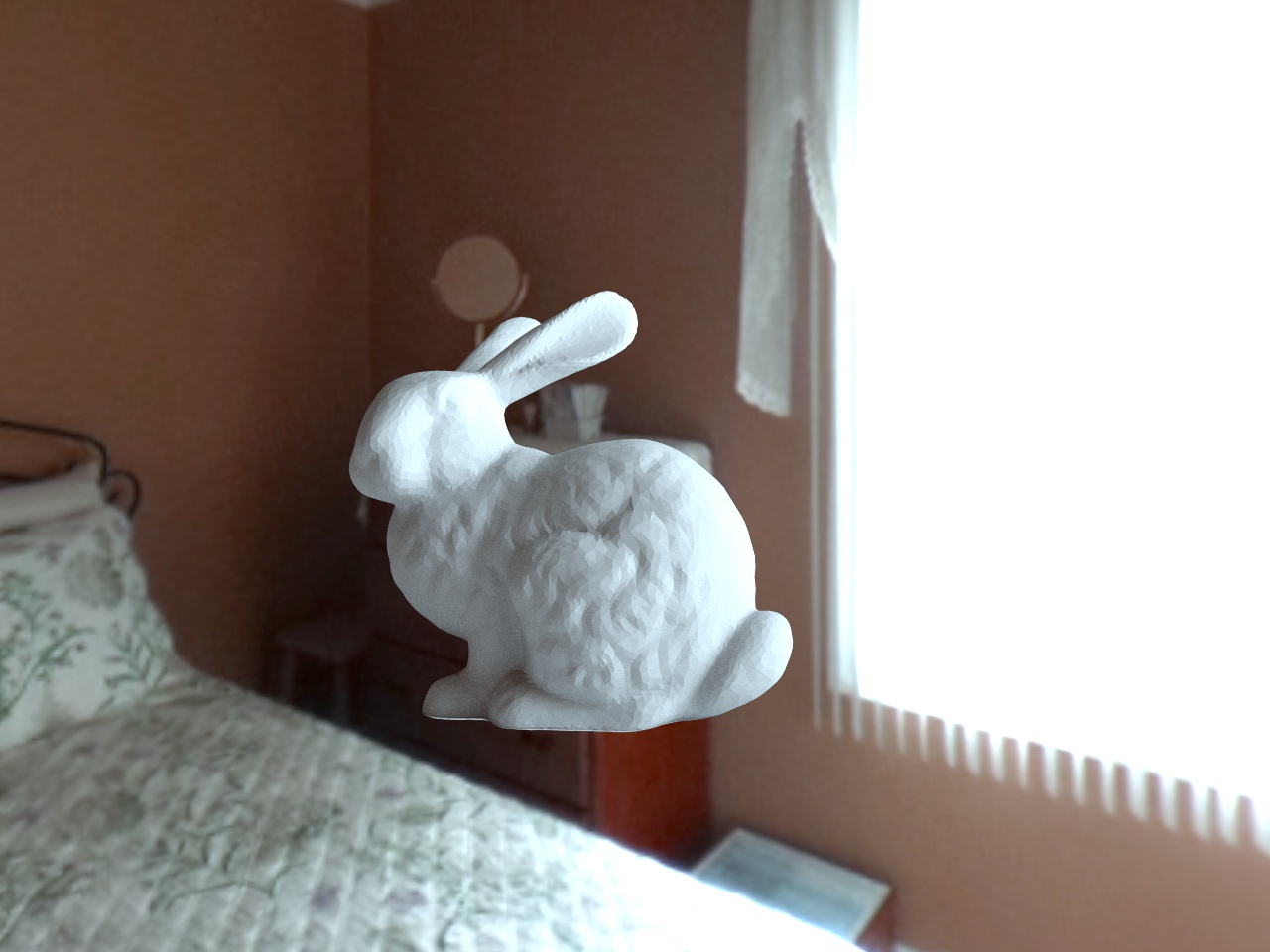} &
\includegraphics[width=0.325\linewidth]{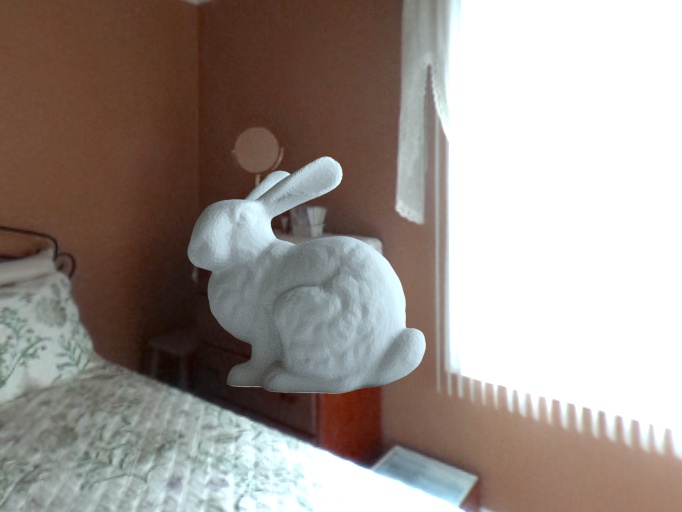} \\
(a) Original panorama & (b) Our warp & (c) \cite{banterle-cgf-13}
\end{tabular}
\caption{Comparison of objects relit with (a) the original panoramas, (b) our warped panoramas, and (c) panoramas warped using EnvyDepth~\cite{banterle-cgf-13}. The objects relit by our panoramas closely approximate those obtained with EnvyDepth, without the lengthy manual annotation required.}
\label{f:warp-results}
\end{figure}

\subsection{Impact on lighting estimation}

Fig.~\ref{f:warping-problem} compares our warped panorama with a ground truth panorama captured in-place for one scene. We also compare our spherical warp with a geometry-based warp obtained with EnvyDepth~\cite{banterle-cgf-13}, a system that extracts spatially-varying lighting from environment maps by projecting them onto proxy geometry estimated from manual annotations. Comparative relighting results using the original, spherical warp, and geometry-based warp panoramas are presented in fig.~\ref{f:warp-results}. While our operator makes several simplifying scene assumptions, these results illustrate that relighting with our approach provides a close approximation to more expensive techniques, while being completely automatic and without requiring access to the scene. In contrast, the manual labeling process required for the geometric warp takes around 10 minutes per panorama. 

The main limitation of our warping operator is that it fails to appropriately model occlusions. Since we treat the panorama as a projection on a sphere, lights that illuminate a scene point, but are not visible from the original camera are not handled by this approach. However, these situations are rare, and as we show in our results, our network filters them out as outliers, and learns a robust scene appearance to illumination mapping.




\section{Learning from LDR panoramas}
\label{sec:learning}

Now that we have the tools to extract accurate lighting information from LDR panoramas, we detail our approach for learning the relationship between a single photo and its lighting conditions.

\begin{table}
\caption[]{The proposed CNN architecture. After a series of 7 convolutional layers (conv), some with residual connections (res), a fully-connected layer (FC) segues to two heads. The heads aim at reconstructing the light mask $\mathbf{y}_\text{mask}$ (left) and the RGB panorama $\mathbf{y}_\text{RGB}$ (right) through a series of deconvolutional layers (deconv). The ELU activation function~\cite{clevert-iclr-16} and batch normalization are used on all layers except the outputs, which are sigmoids for light mask and tangent hyperbolic for panorama prediction. The stride at each layer is indicated between parentheses. The ``res'' identifiers indicate residual layers~\cite{he-cvpr-16}. }
\centering
\begin{tabular}{c}
\toprule
\textbf{Layer (stride)} \\
\midrule
Input \\
\midrule
conv9-64 (2) \\
conv4-96 (2) \\
res3-96  (1) \\
res4-128 (2) \\
res4-192 (2) \\
res4-256 (2) \\
\midrule
FC-1024 \\
\end{tabular}
\\
\begin{tabular}{cc}
\midrule
FC-8192 & FC-6144 \\
deconv4-256 (2) & deconv4-192 (2) \\
deconv4-128 (2) & deconv4-128 (2) \\
deconv4-96 (2) & deconv4-64 (2) \\
deconv4-64 (2) & deconv4-32 (2) \\
deconv4-32 (2) & deconv4-24 (2) \\
conv5-1 (1) & conv5-3 (1) \\
Sigmoid & Tanh \\*[-.5em]
\noindent\rule{3.2cm}{.8pt} &
\noindent\rule{3.2cm}{.8pt} \\
Output: light mask $\mathbf{y}_\text{mask}$ &
Output: RGB panorama $\mathbf{y}_\text{RGB}$ \\
\end{tabular}
\label{t:learning-architecture}
\end{table}

\subsection{Training data preparation}
\label{sec:ldr-data-prep}

For each SUN360 indoor panorama, we compute the light mask to represent ground truth during the learning process (sec.~\ref{sec:lightdetection}). We then take 8 crops from each panorama at random elevation between $-30^\circ$ and $30^\circ$ and make a projection of them as rectilinear photos. Using our recentering warp (sec.~\ref{sec:warping}), we generate a corresponding warped panorama (and light mask) for each rectilinear photo. We also rotate the warped panorama and corresponding light mask so that the crop region always sits at center azimuth (fig.~\ref{f:overview}).  At the end of this process, we have 96,000 input-output pairs, where the input is a photo, and the output is a pair of a warped panorama and its corresponding light mask.

\subsection{Network architecture} 

As shown in table~\ref{t:learning-architecture}, we use a convolutional neural network that takes the photo as input, produces a low-dimensional encoding of the input through a series of convolutions downstream and splits into two upstream expansions, with two distinct tasks: (1) intensity estimation / binary light mask prediction, and (2) RGB panorama prediction. The encoder is split into two standard convolution layers, followed by four residual layers~\cite{he-cvpr-16}. The two individual decoders are exclusively composed of deconvolution layers. The input photo is of size $256\times192$, whereas the panorama and light mask outputs are of size $256\times128$. Each time a stride of 2 is encountered with a convolution (deconvolution) layer, the resolution of its output feature map is divided (multiplied) by two. The output light mask $\mathbf{x}_\text{mask}$ represents the probability of light for each pixel in the environment map. The RGB panorama $\mathbf{x}_\text{mask}$ serves as a high level colored texture in the final environment map. Please see sec.~\ref{sec:experiments} for several examples of estimated light masks and RGB panoramas.

\begin{figure}
\centering
\footnotesize
\setlength{\tabcolsep}{1pt}
\begin{tabular}{ccc}
\includegraphics[width=0.32\linewidth]{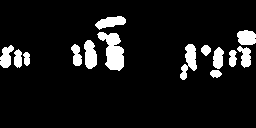} &
\includegraphics[width=0.32\linewidth]{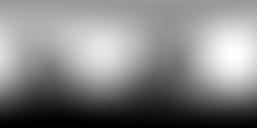} &
\includegraphics[width=0.32\linewidth]{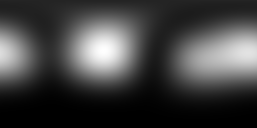} \\
(a) Original & (b) $\alpha e=1$ & (c) $\alpha e=5$ \\
\includegraphics[width=0.32\linewidth]{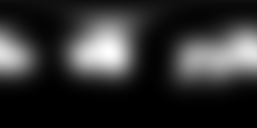} &
\includegraphics[width=0.32\linewidth]{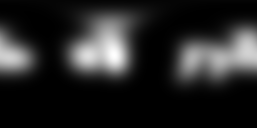} &
\includegraphics[width=0.32\linewidth]{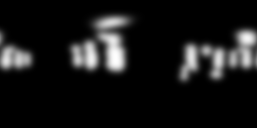} \\
(d) $\alpha e=10$ & (e) $\alpha e=20$ & (f) $\alpha e=80$ \\
\end{tabular}
\caption{Effect of the cosine blurring from eq.~(\ref{e:filter}) on the light mask at various blurring levels. Note how this simple, differentiable scheme allows a smooth progression towards higher frequency content over time, but without the ringing artifacts of spherical harmonics.}
\label{f:learning-filter}
\end{figure}

\subsection{Loss function} 

For the RGB panorama prediction task, we use an L2 distance on the pixel output:
\begin{equation}
    \mathcal{L}_\text{L2}(\mathbf{y}, \mathbf{t}) = \frac{1}{N}\sum_{i=1}^{N} \mathbf{s}_i (\mathbf{y}_i - \mathbf{t}_i)^2 \,,
\label{e:rgbloss}
\end{equation}
where $N=\mathtt{width}\times\mathtt{height}\times 3$ is the total number of elements in the image, $\mathbf{y}$ is the network prediction, $\mathbf{t}$ the ground truth panorama and $\mathbf{s}_i$ the solid angle for pixel $i$.

Designing the loss function for the light mask $\mathbf{y}_\text{mask}$ is not as straightforward. Take, for example, the standard L2 or binary cross entropy losses computed on the light mask directly. If a small bright spotlight is estimated to be located slightly off its ground truth location, a huge penalty will incur. Since pinpointing the exact location of all the light sources from a single photo is not necessary, we instead blur the target light mask with a filter and compute the L2 loss on the blurred version. The filter starts with a coarse, low-frequency representation of the target light mask and progressively sharpens it over training time. To this end, we design a filter based on the cosine distance, followed by an L2 loss for the light mask: 
\begin{equation}
    \mathcal{L}_\text{cos}(\mathbf{y}, \mathbf{t}, e) = \frac{1}{N}\sum_{i=1}^{N} (\mathcal{F}(\mathbf{y}, i, e) - \mathcal{F}(\mathbf{t}, i, e))^2 \,,
    \label{e:maskloss}
\end{equation}
where $e$ is a real value corresponding to the current epoch (formally, $e=\textrm{\#epochs}+\textrm{\#current-mini-batch}/\textrm{\#total-mini-batches}$.). The filter $\mathcal{F}$ is defined as:
\begin{equation}
    \mathcal{F}(\mathbf{p}, i, e) = \frac{1}{K_i} \sum_{\omega \in \Omega_i} \mathbf{p}(\omega) s(\omega) \left( \omega \cdot n_i \right)^{\alpha e},
    \label{e:filter}
\end{equation}
where $\Omega_i$ is the hemisphere centered at pixel $i$ on the panorama $\mathbf{p}$,  $n_i$ the unit normal at pixel $i$, and $K_i$ the sum of solid angles on $\Omega_i$. $\omega$ is a unit vector in a specific direction on $\Omega_i$ and $s(\omega)$ the solid angle for the pixel in the direction $\omega$. As seen before, we define $e\in\mathbb{R}$ as the real valued number of training samples collectively seen, normalized by the total number of training samples. 

\begin{figure}
\centering
\footnotesize
\setlength{\tabcolsep}{1pt}
\begin{tabular}{cc}
\includegraphics[width=0.493\linewidth]{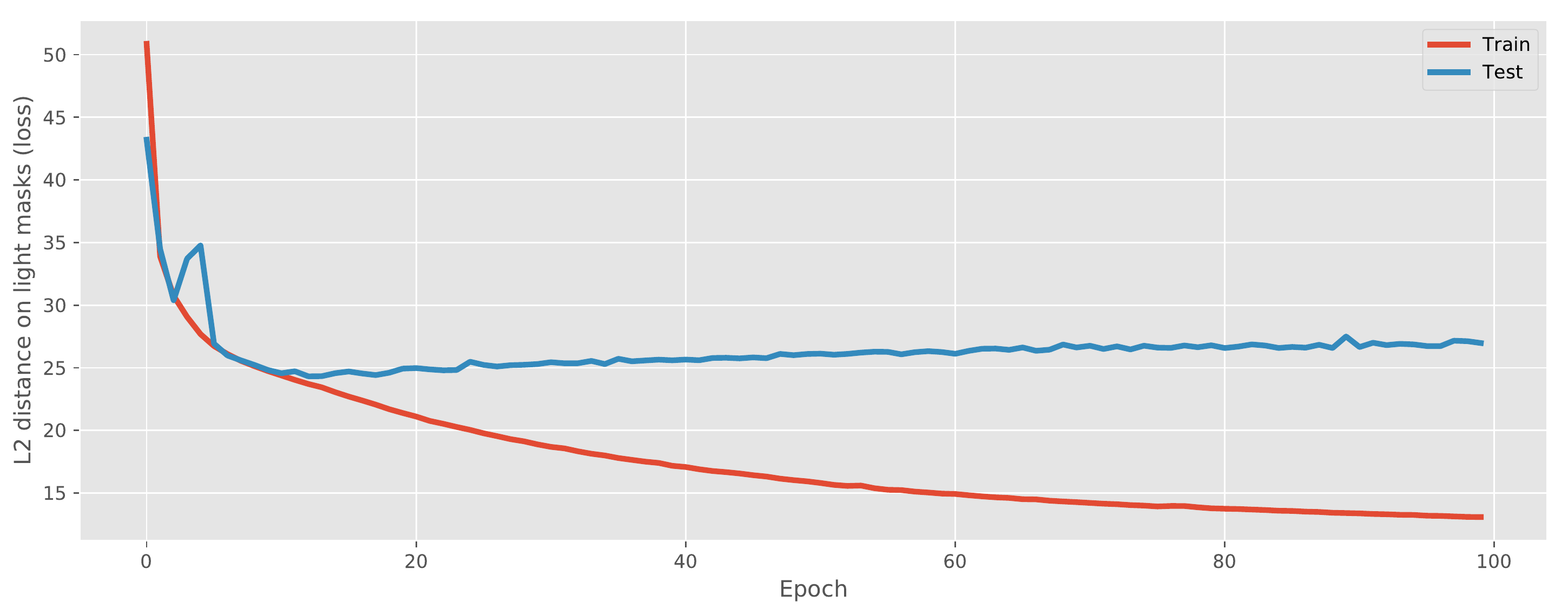} &
\includegraphics[width=0.493\linewidth]{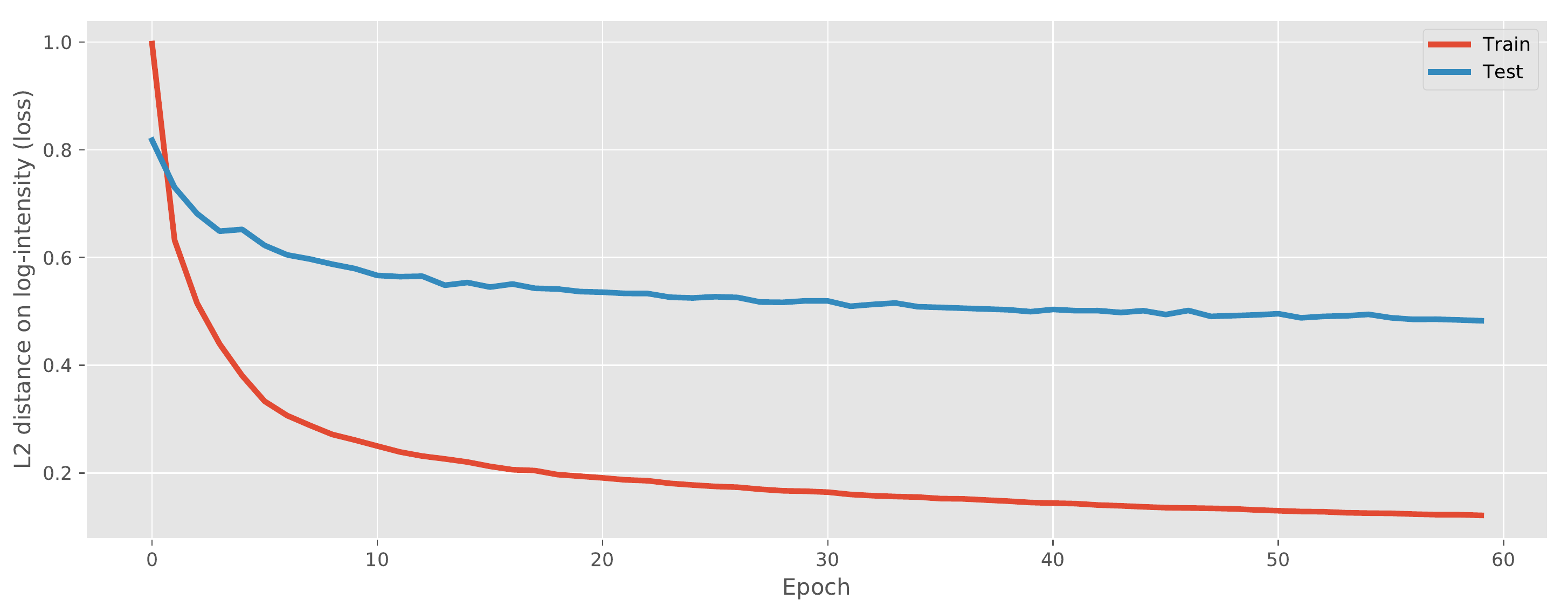} \\
(a) LDR network & (b) HDR network
\end{tabular}
\caption{Evolution of training and test loss over the number of epochs for the (a) LDR and (b) HDR training.}
\label{f:learning-curves}
\end{figure}

Since eq.~\ref{e:filter} is differentiable, back-propagation can be used to efficiently train our CNN. Fig.~\ref{f:learning-filter} shows a visual example of the effect of the cosine distance filter on a binary light mask. Note how the target light mask becomes progressively sharper over time. 

\begin{figure*}[!t]
\centering
\footnotesize
\setlength{\tabcolsep}{1pt}
\begin{tabular}{cccc}
\includegraphics[height=2.5cm]{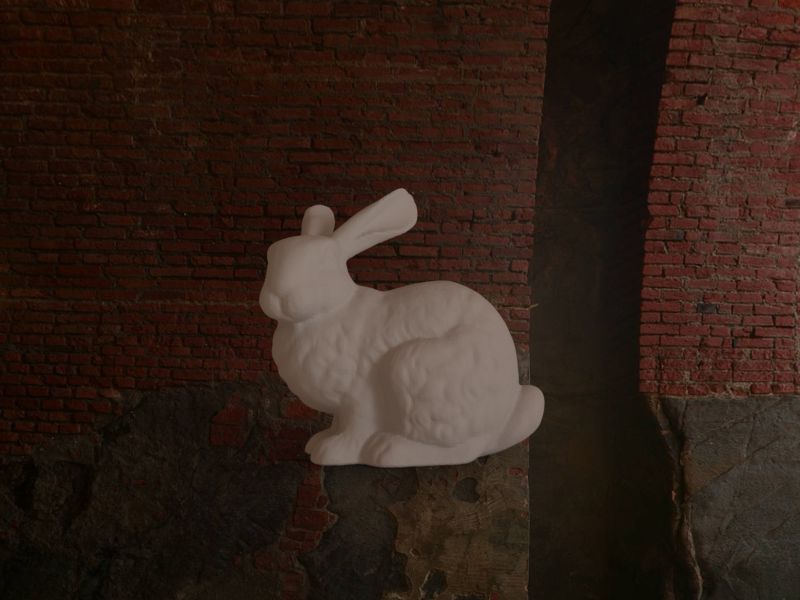} & 
\includegraphics[height=2.5cm]{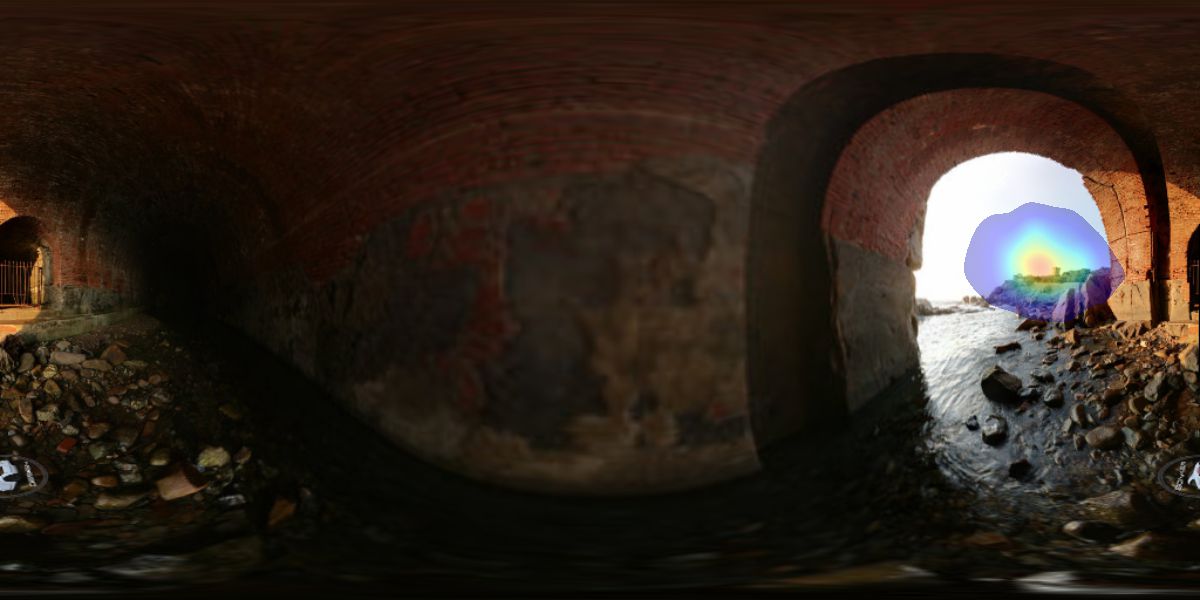} & 
\hspace{.5em}
\includegraphics[height=2.5cm]{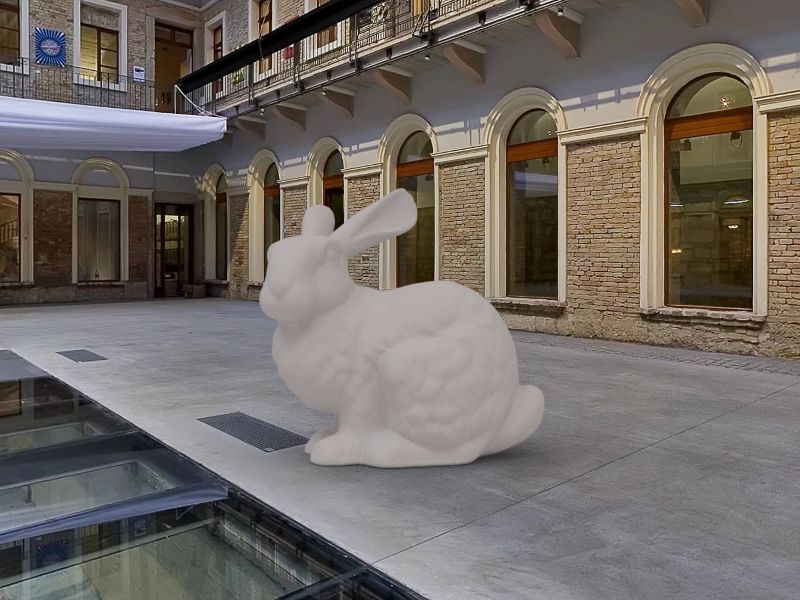} & 
\includegraphics[height=2.5cm]{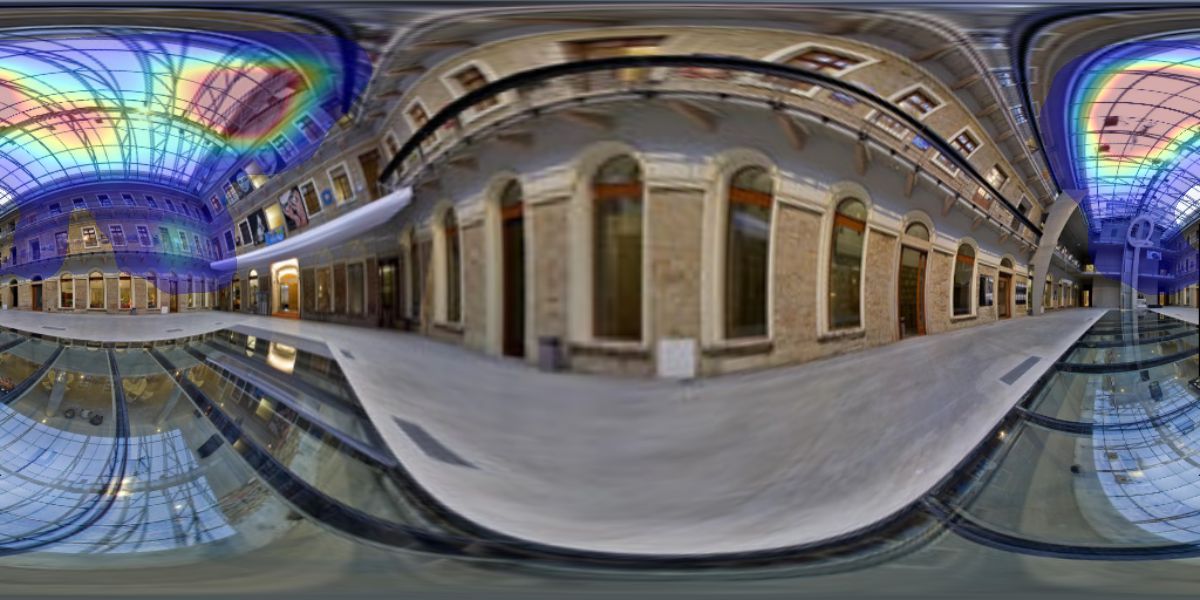} \\
\includegraphics[height=2.5cm]{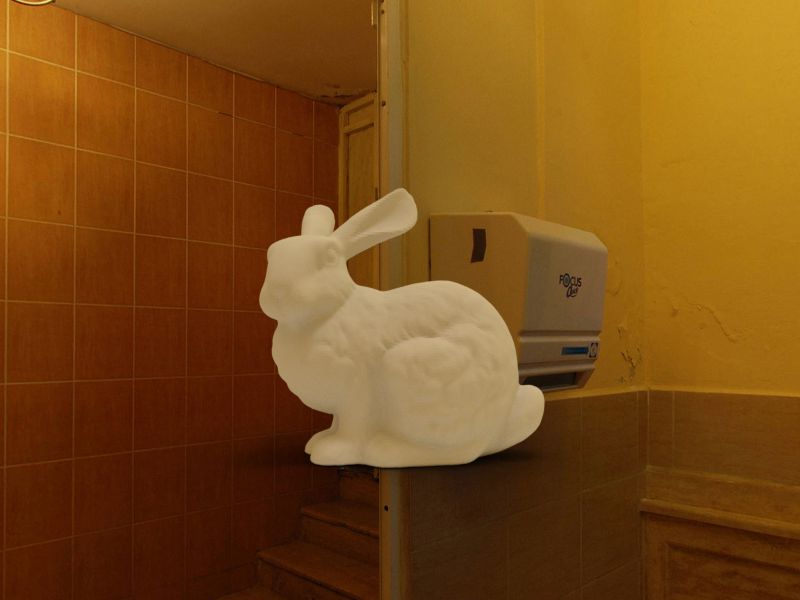} & 
\includegraphics[height=2.5cm]{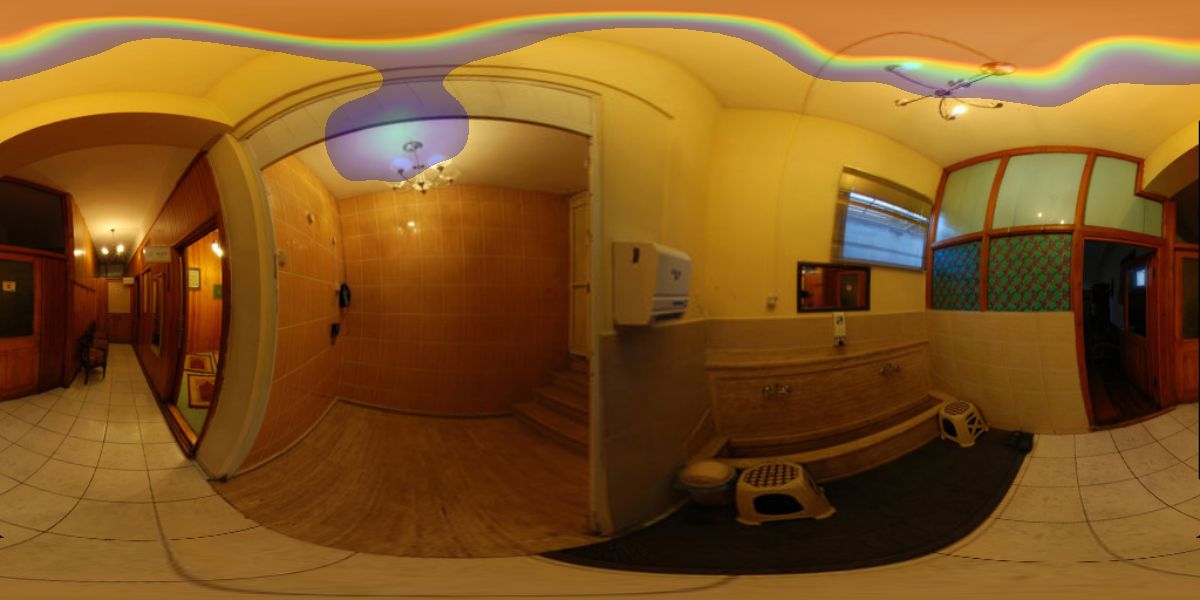} & 
\hspace{.5em}
\includegraphics[height=2.5cm]{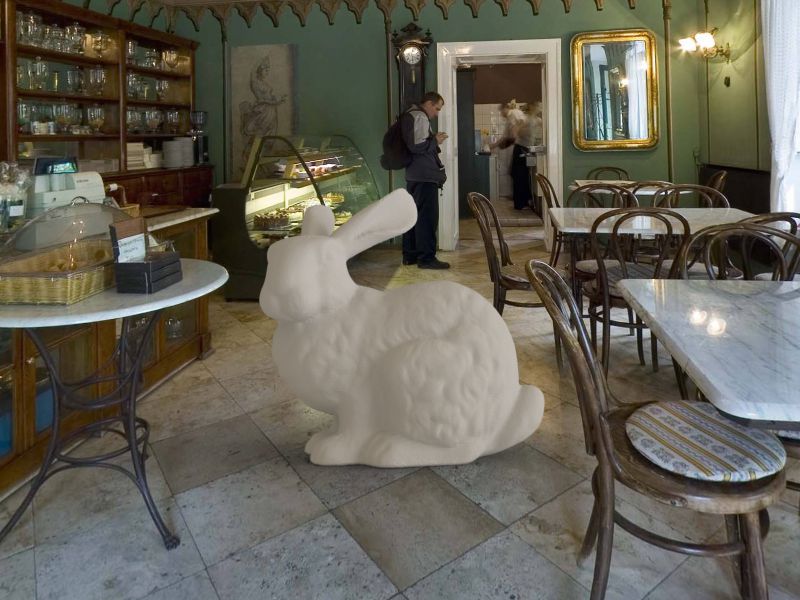} & 
\includegraphics[height=2.5cm]{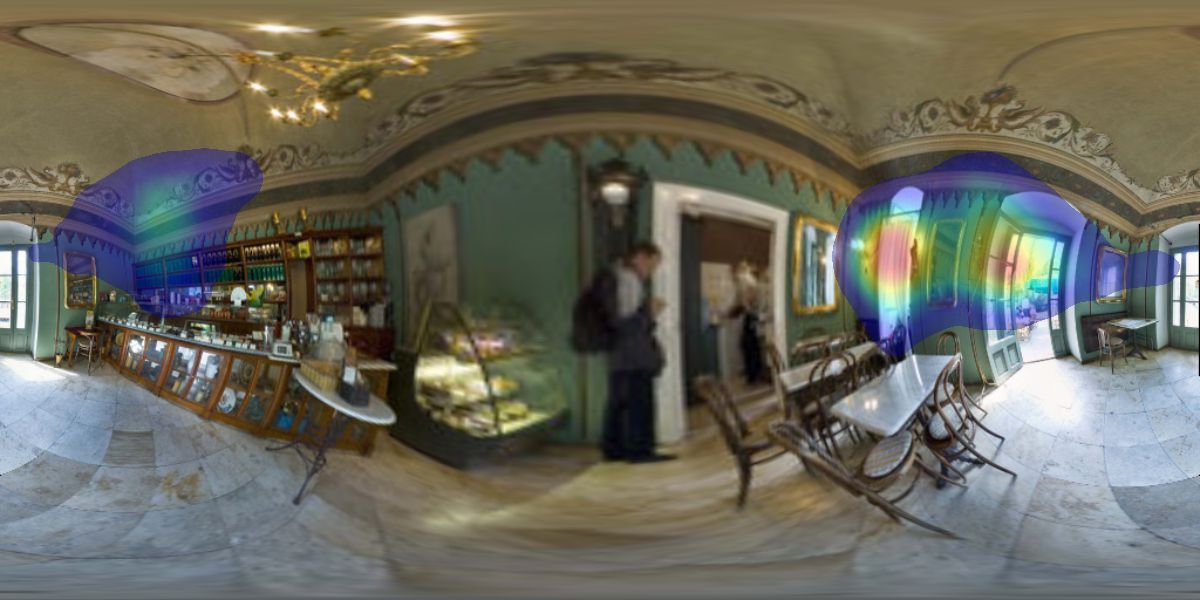} \\
\includegraphics[height=2.5cm]{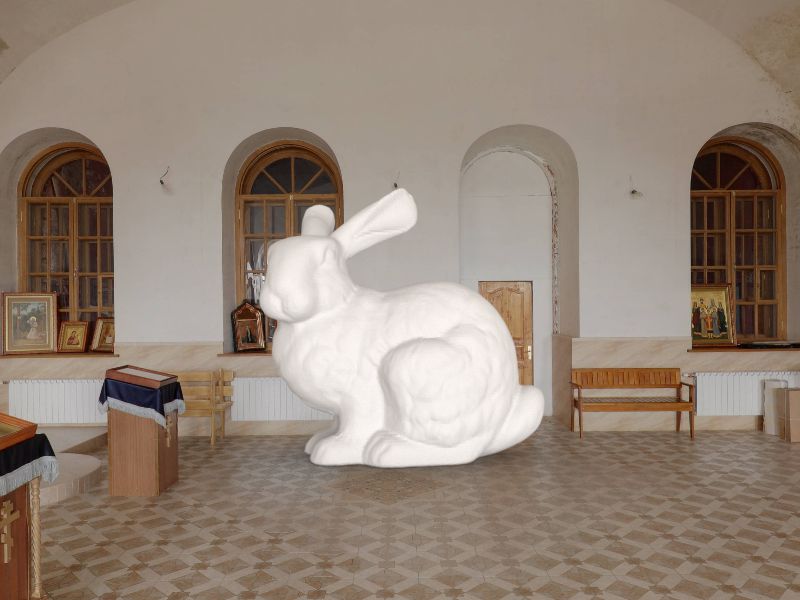} & 
\includegraphics[height=2.5cm]{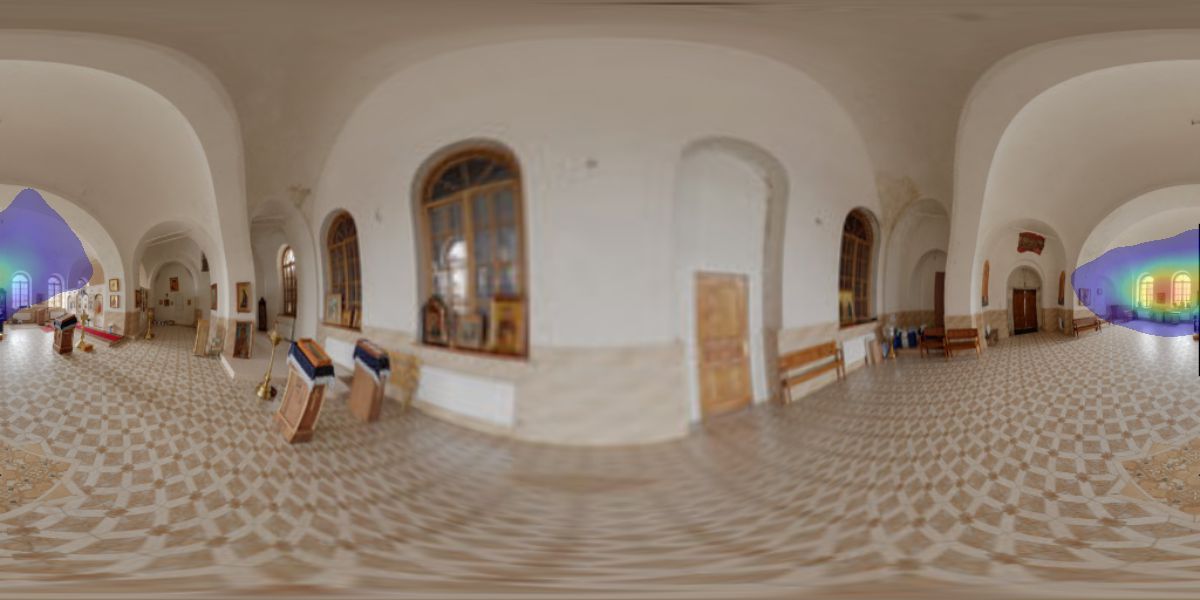} & 
\hspace{.5em}
\includegraphics[height=2.5cm]{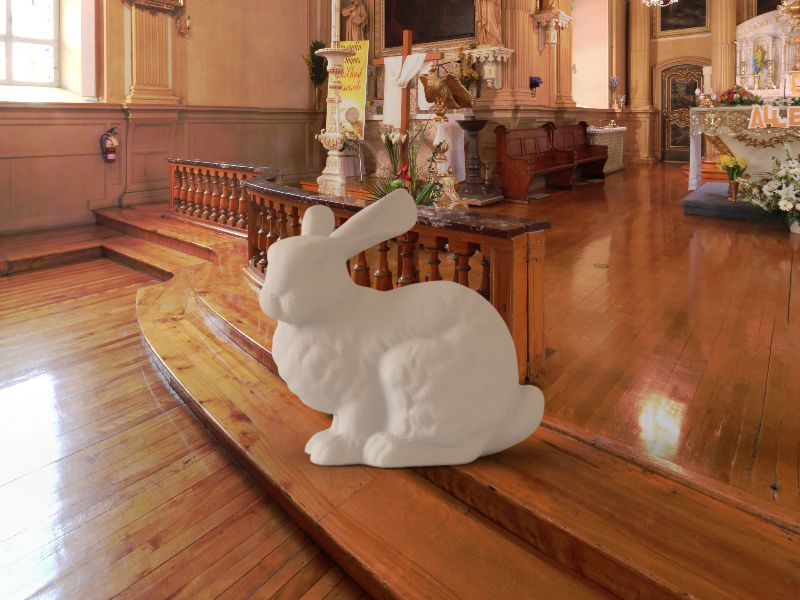} & 
\includegraphics[height=2.5cm]{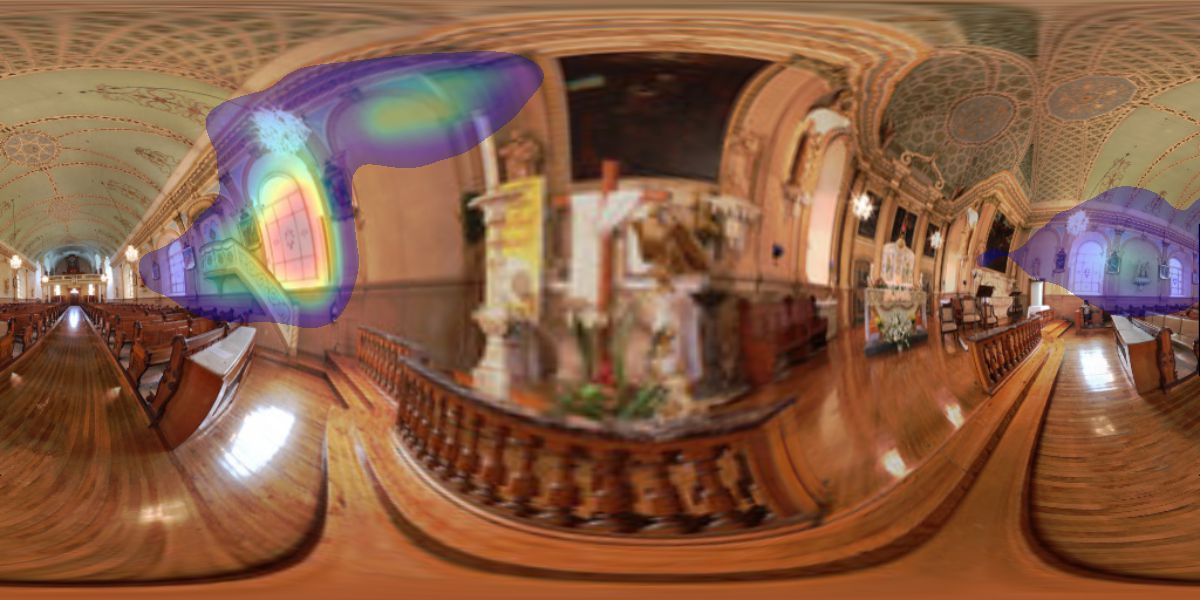} \\
\includegraphics[height=2.5cm]{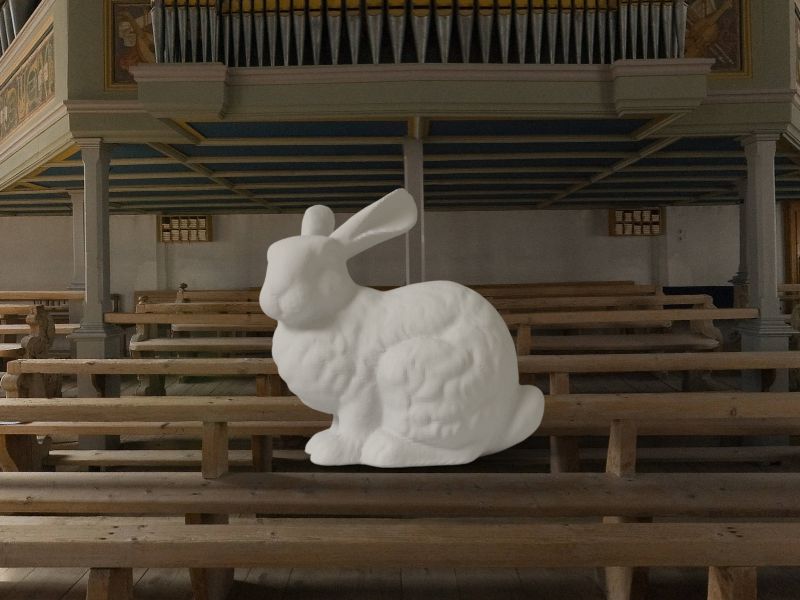} & 
\includegraphics[height=2.5cm]{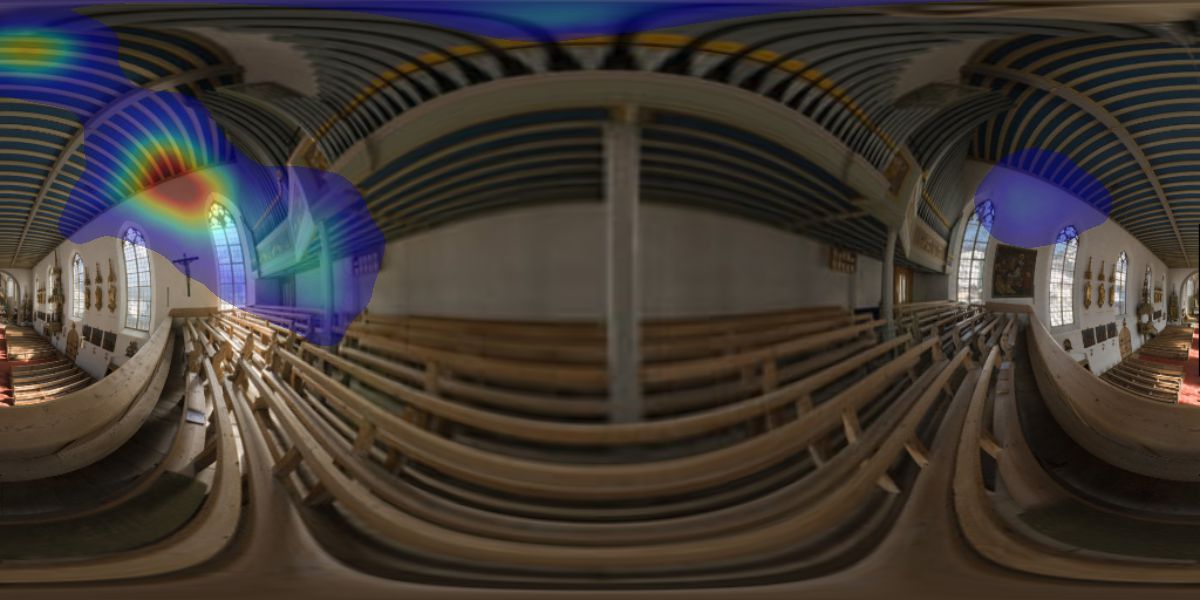} & 
\hspace{.5em}
\includegraphics[height=2.5cm]{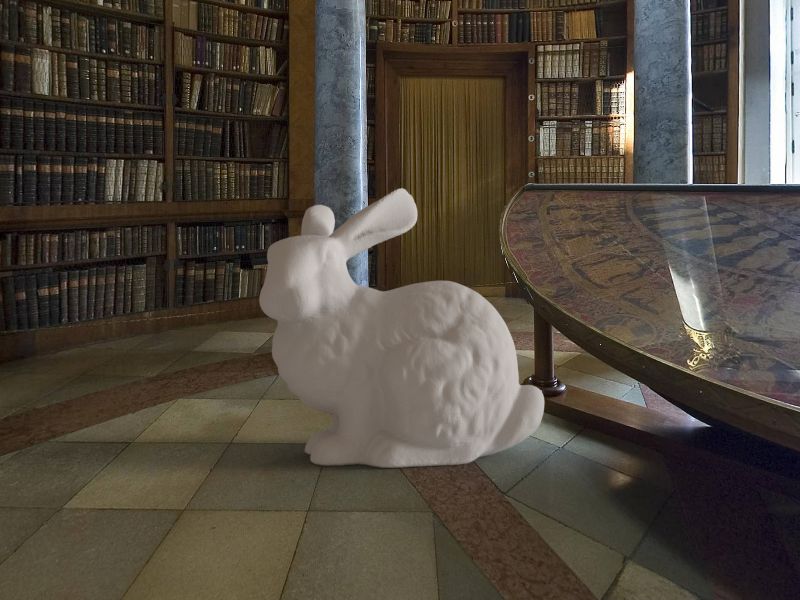} & 
\includegraphics[height=2.5cm]{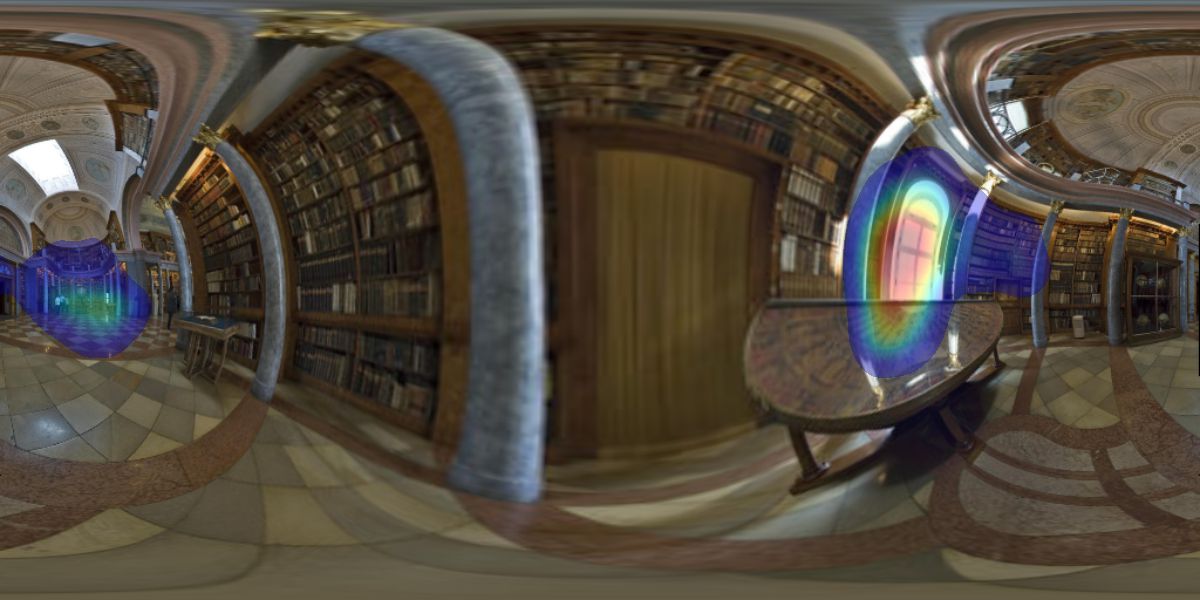} \\
(a) Relit with estimate & (b) Predicted light probability & 
\hspace{.5em}
(c) Relit with estimate & (d) Predicted light probability
\end{tabular}
\caption{Evaluation of the LDR network at predicting light source positions. For each example, we show a virtual bunny model inserted in a background image and relit with the LDR network estimate for that image ((a) and (c)), and the predicted lighting probabilities overlaid on the panorama ((b) and (d)). As can been seen, our method generalizes to a wide range of indoor scenes and illumination conditions. Many more examples are available in the supplementary material.}
\label{f:relighting-bunnies}
\end{figure*}

The global loss function is then defined as:
\begin{equation}
    \mathcal{L}(\mathbf{y}, \mathbf{t}, e) = w_1 \mathcal{L}_\text{L2}(\mathbf{y}_\text{RGB}, \mathbf{t}_\text{RGB})  
    + w_2 \mathcal{L}_\text{cos}(\mathbf{y}_\text{mask}, \mathbf{t}_\text{mask}, e) \,.
\label{e:globloss}
\end{equation}
In our experiments, we use $w_1=100$, $w_2=1$, and $\alpha=3$.

Our filtering scheme also has a rendering-based interpretation. It is well known that surface reflection for Lambertian objects can be modeled as low-pass filtering~\cite{ramamoorthi-sig-01}, while specular objects preserve more high frequencies of the illumination. In this sense, our loss function can be thought of evaluating the inferred illumination in terms of the resulting \emph{appearance} of spheres with increasingly glossy surface reflectance. In this vein, we experimented with directly representing the illumination with spherical harmonics (gradually increasing the number of coefficients to represent higher frequencies of illumination), but found that the network had a tendency to overfit to the ringing artifacts caused by high frequencies in the binary light mask.

\subsection{Training details} 
\label{sec:training-details}

We use $85\%$ of the panoramas as training data, and $15\%$ as test data. Note that we generate the train-test split such that no crop of the test panoramas exist in the training set. Hence, all tests are performed for scenes and lighting conditions that have not been seen by the network before. We use the ADAM optimizer~\cite{kingma2014adam} with a minibatch size of $64$, learning rate of $0.005$, and momentum parameters of $\beta_1=0.9$, $\beta_2=0.999$. Fig.~\ref{f:learning-curves}-(a) shows the loss (from eq.~(\ref{e:globloss})) curves on the training and test set during training. Training takes roughly 40 hours on an Nvidia Titan X Pascal GPU. At test time, lighting inference (both mask and RGB) from a photo takes approximately 10ms. The batch size was selected so it fills the 12GB memory of the GPU.

\section{Learning high dynamic range illumination}
\label{sec:hdr}

Up to this point we have trained a network that can predict the \emph{position} of the light sources quite accurately (see sec.~\ref{sec:experiments}), but, since it was trained on LDR data, it does not know about the \emph{intensities} of the light sources. In this section, we further train the network on a novel dataset of high dynamic range panoramas which enables it to jointly reason about light source direction and intensity. 

\subsection{A new dataset of HDR indoor panoramas}

We have captured a novel dataset of 2,100 high-resolution ($7768\times3884$), high dynamic range indoor panoramas. To do so, a Canon 5D Mark III camera with a Sigma 8mm fisheye lens was mounted on a tripod equipped with a robotic panoramic tripod head, and programmed to shoot 7 bracketed exposures at $60^\circ$ increments. The photos were shot in RAW mode, and automatically stitched into a 22 f-stop HDR $360^\circ$ panorama using the PTGui Pro commercial software. The dynamic range is sufficient to correctly expose all pixels in the scenes, including the light sources. Panoramas were captured in a wide variety of indoor environments, such as schools, houses, apartments, museums, laboratories, factories, sports facilities, etc. A visual overview of panoramas in our novel HDR dataset is shown in the supplementary material. The size and variety of this dataset is significantly larger than other similar datasets in the literature (which consist of tens of panoramas), making it extremely useful for training and testing a wide range of problems from scene inference, high dynamic range image processing, and rendering\footnote{This dataset is publicly available at \url{http://www.jflalonde.ca/projects/deepIndoorLight}.}. 

\subsection{Adapting the network to HDR data}

Since the light sources are not saturated in the HDR data, the network can be adjusted to directly learn the light source \emph{intensities} $\mathbf{y}_\text{int}$ instead of the binary light mask $\mathbf{y}_\text{mask}$. To do so, the network undergoes the following four simple changes. First, the weights of the last layer of the light mask predictor (``conv5-1'' in table~\ref{t:learning-architecture}) are initialized to random values. Second, training is performed to update only the weights of the decoders---that is, up to the FC-1024 layer in table~\ref{t:learning-architecture}. This is done to avoid overfitting on the encoder. Third, the target intensity $\mathbf{t}_\text{int}$ is defined as the log of the HDR intensity ($\log_{10}$ is used). Low intensities (below the median of the training dataset) are clamped to 0, since we only care about the light sources: in the unusual case where no pixels would be over this threshold, the ambient term given by the RGB recovery should be enough to light the scene. Finally, the loss is modified to:
\begin{align}
    \mathcal{L}_\text{HDR}(\mathbf{y}, \mathbf{t}, e) &= w_1 \mathcal{L}_\text{L2}(\mathbf{y}_\text{RGB}, \mathbf{t}_\text{RGB}) \nonumber \\ 
    &+ w_2 \mathcal{L}_\text{cos}(\mathbf{y}_\text{int}, \mathbf{t}_\text{int}, e)
    + w_3 \mathcal{L}_\text{L2}(\mathbf{y}_\text{int}, \mathbf{t}_\text{int}, e)  \,,
\label{e:hdrloss}
\end{align}
where $\mathcal{L}_\text{L2}$ and $\mathcal{L}_\text{cos}$ were defined in eq.~(\ref{e:rgbloss}) and (\ref{e:maskloss}) respectively, and $e$ is continued from training on the LDR data (so the HDR intensities are not overblurred). The L2 term on the intensity was added to reduce deconvolution artifacts. Here, $w_1 = 10$, $w_2 = 1$, and $w_3 = 0.1$. Training is otherwise performed with the same parameters as in sec.~\ref{sec:training-details}, and, just as with the LDR data, 85\% of the HDR data was used for training and 15\% for testing. Similar to the LDR data (sec.~\ref{sec:ldr-data-prep}), 8 crops were extracted from each panorama in the HDR dataset, yielding 14,000 input-output pairs. These are tone-mapped to ensure that the input to the network are LDR images. Finally, the panoramas are also warped using the same procedure as their LDR counterparts. Fig.~\ref{f:learning-curves}-(b) shows the loss (from eq.~(\ref{e:hdrloss})) curves on the training and test set during training.

\section{Experiments}
\label{sec:experiments}

In this section, we evaluate our approach in several different ways. First, we present light prediction results on the SUN360 dataset (LDR data), where the lights found by the detector of sec.~\ref{sec:lightdetection} are treated as ground truth. Then, we show the results of the HDR fine-tuning procedure, comparing our results with actual HDR ground truth. Finally, we compare our technique to previous work and present a user study comparing the performance of each method at relighting virtual objects. In the following analysis, the term ``LDR network'' refers to the network trained on the SUN360 dataset which recovers a binary light mask (sec.~\ref{sec:learning}), and ``HDR network'' refers to the network finetuned on the HDR dataset which recovers the light intensity (sec.~\ref{sec:hdr}). Note that when not otherwise specified, every lighting estimate shown in the paper and the supplementary material has been computed \emph{completely automatically}. The only manual intervention in these results has been, in some cases, to place virtual objects in the scene. Please refer to the supplementary material for more results for each step.


\subsection{Evaluation of the LDR network}

We provide a qualitative way of evaluating the LDR network ability to estimate the light direction from single images by rendering a virtual bunny model into the image. To do so, a coarse environment map is obtained by thresholding the light mask $\mathbf{x}_\text{mask}$ (at $t > 0.5$), detecting the connected components, and assigning a weight corresponding to the mean light probability from $\mathbf{x}_\text{mask}$ to each component. This modified $\mathbf{x}^*_\text{mask}$ is combined with the RGB panorama $\mathbf{x}_\text{RGB}$ into a single environment map by:
\begin{equation}
\mathbf{x}_\text{combined} = \lambda_\text{mask} \mathbf{x}^*_\text{mask} + \lambda_\text{RGB}(1-\mathbf{x}^*_\text{mask}) \mathbf{x}_\text{RGB} \,,
\label{e:ldr-envmap}
\end{equation}
where the parameters $\lambda_\text{mask}$ and $\lambda_\text{RGB}$ are set to 500 and 1, respectively (these choices are arbitrary and used only to visualize the LDR network capabilities at predicting the positions of light sources). As can be seen in fig.~\ref{f:relighting-bunnies}, our LDR network successfully localizes light sources even in images with small field-of-view and few visually obvious illumination cues. 

\begin{figure}[!t]
\centering
\footnotesize
\setlength{\tabcolsep}{1pt}
\begin{tabular}{cc}
\includegraphics[height=2.5cm]{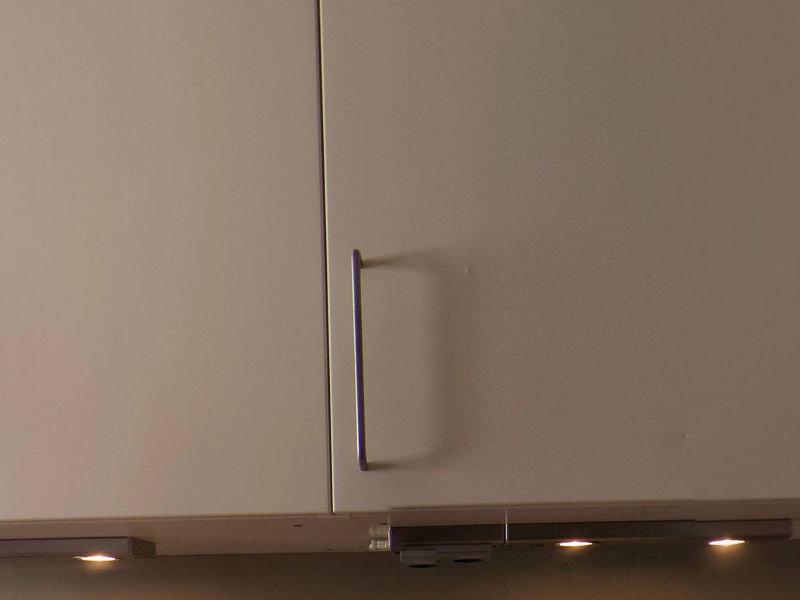} & 
\includegraphics[height=2.5cm]{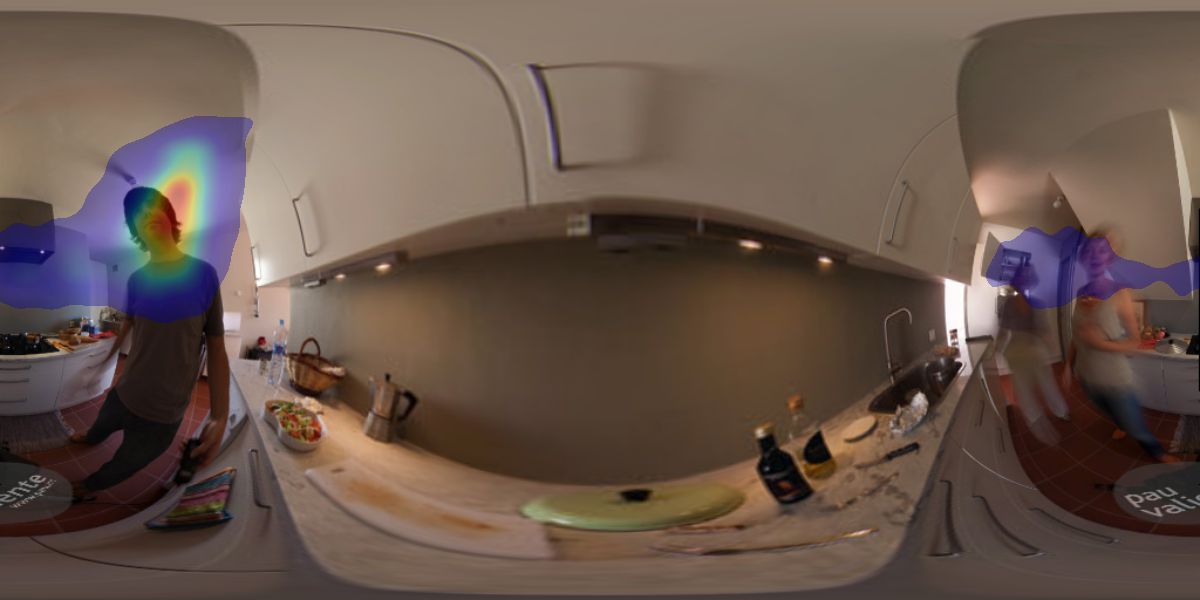} \\
\end{tabular}
\caption{Effect of occlusions on the LDR network predictions. While our warping operator does not handle light occlusions, our network is able to generalize beyond these rare instances. Here, even though the window causing the shadows on the handle in the image (left) is occluded in the panorama (right), our network places the highest probability of a light in this direction, thus producing results that are ``better'' than the ground truth.}
\label{f:occluded-lights}
\end{figure}

\paragraph{Handling occluded lights} In sec.~\ref{sec:warping}, we noted that our warping operator is an approximation that does not handle occluded light sources. However, this scenario is rare, and as a result our network is able to robustly generalize beyond these erroneous input data. This is demonstrated in Fig.~\ref{f:occluded-lights}, where our network predicts lights that are not in the ground truth annotation because of occlusion, but are consistent with the shading and shadows observed in the cropped photo.

\subsection{Evaluation of the HDR network}
\label{ss:hdreval}
In this section, a more thorough evaluation of the HDR network, including quantitative and qualitative results, is provided. 

\paragraph{Quantitative evaluation}

\begin{figure}[!t]
\centering 
\includegraphics[width=.72\linewidth]{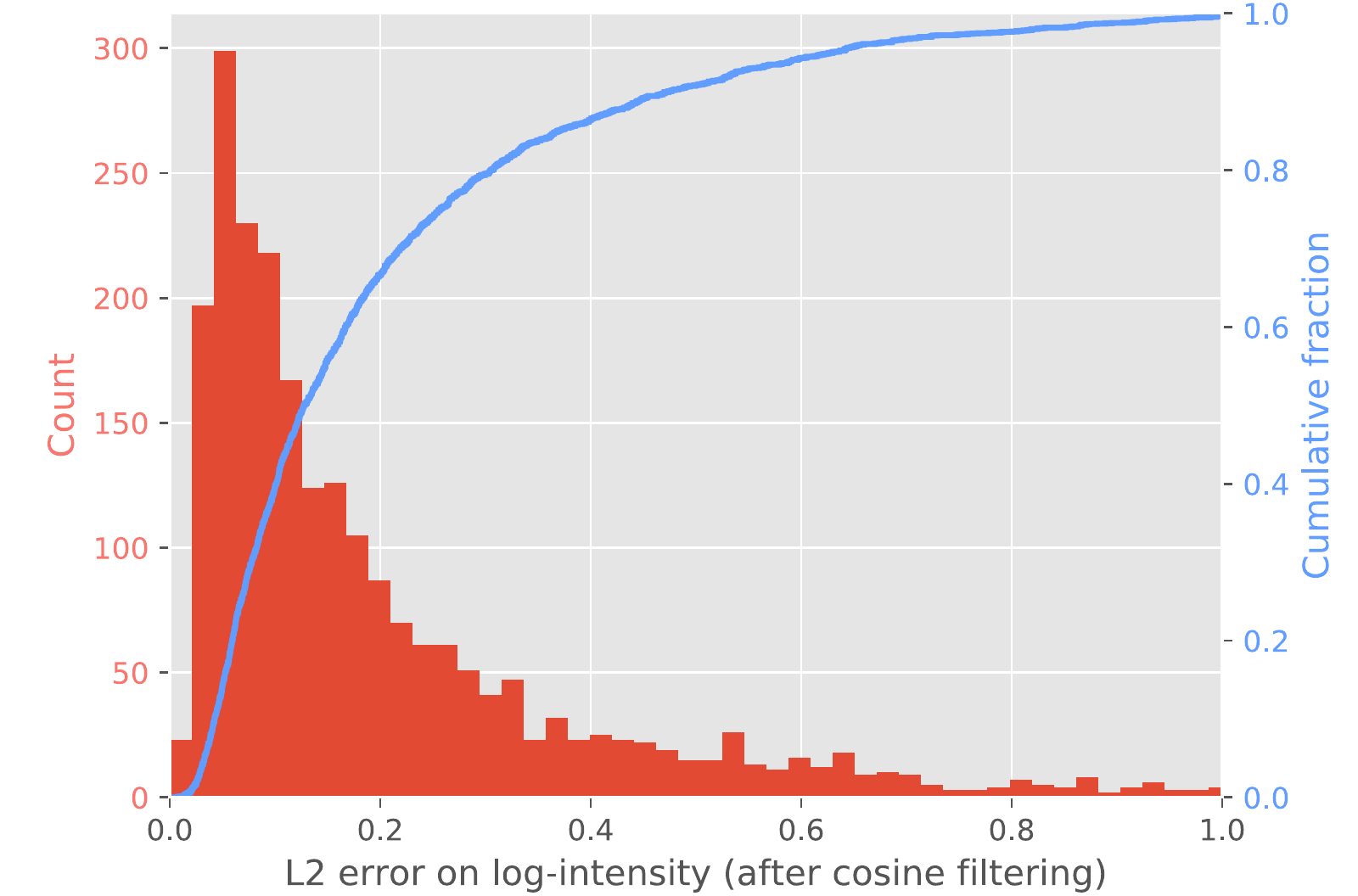}
\caption{Histogram (red) and cumulative histogram (blue) of the loss on the log-intensity (\ref{e:hdrloss}) over the HDR test set, after convergence ($e = 70$). See fig.~\ref{f:results-loss-qualitative} for qualitative examples corresponding to different loss percentiles.}
\label{f:results-loss-histogram}
\end{figure}

\begin{figure*}[!t]
\centering
\footnotesize
\setlength{\tabcolsep}{1pt}
\begin{tabular}{ccc}
\includegraphics[width=.33\linewidth]{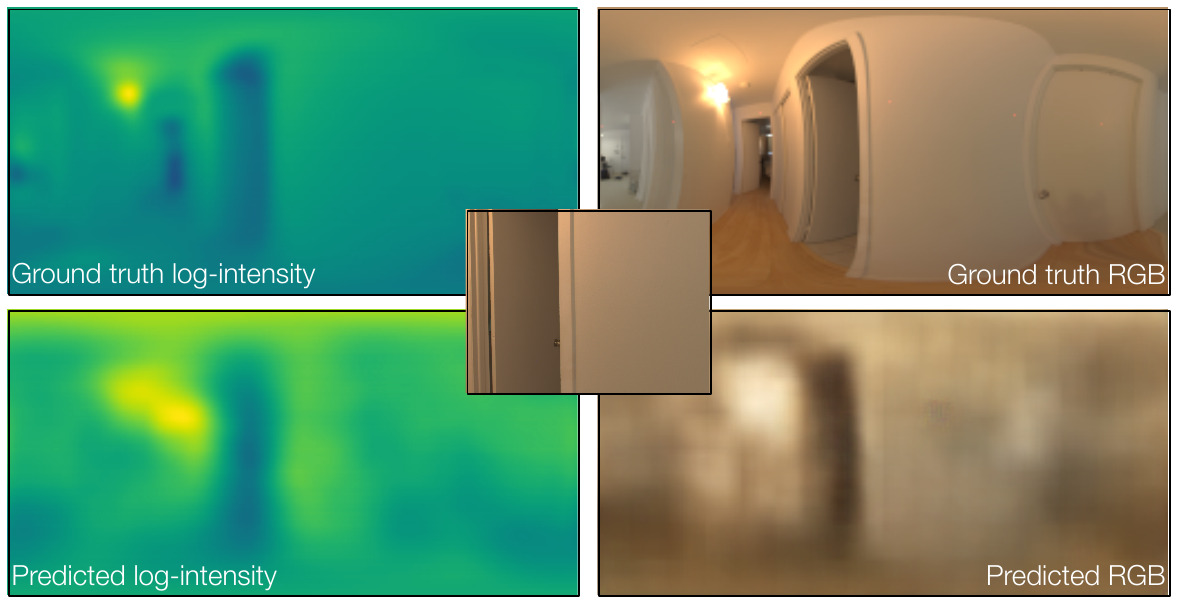} &
\includegraphics[width=.33\linewidth]{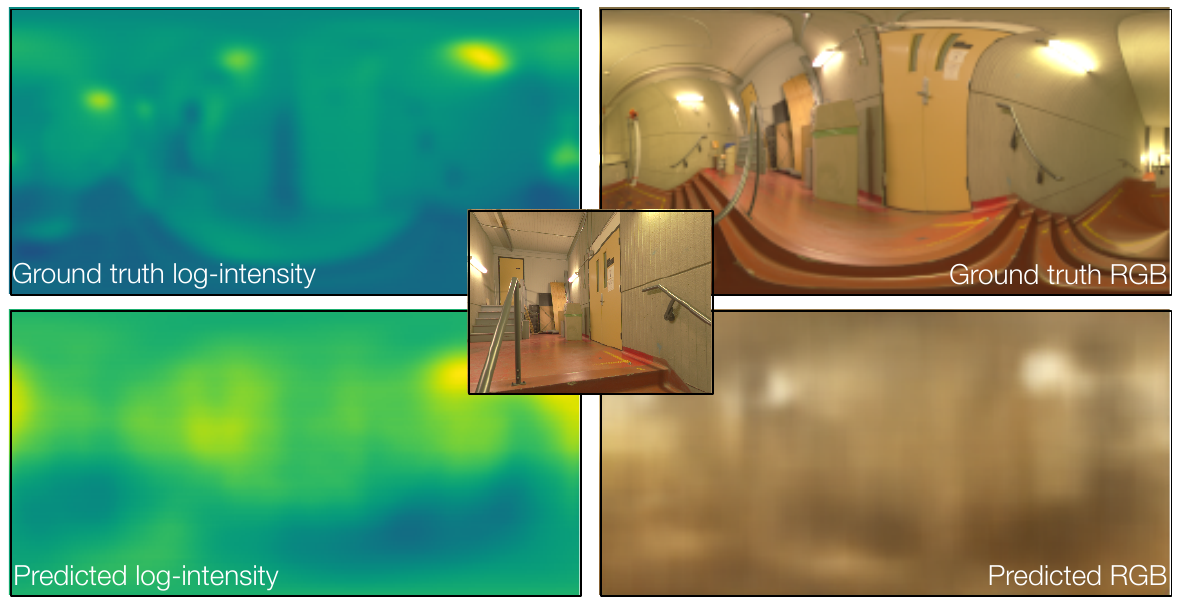} &
\includegraphics[width=.33\linewidth]{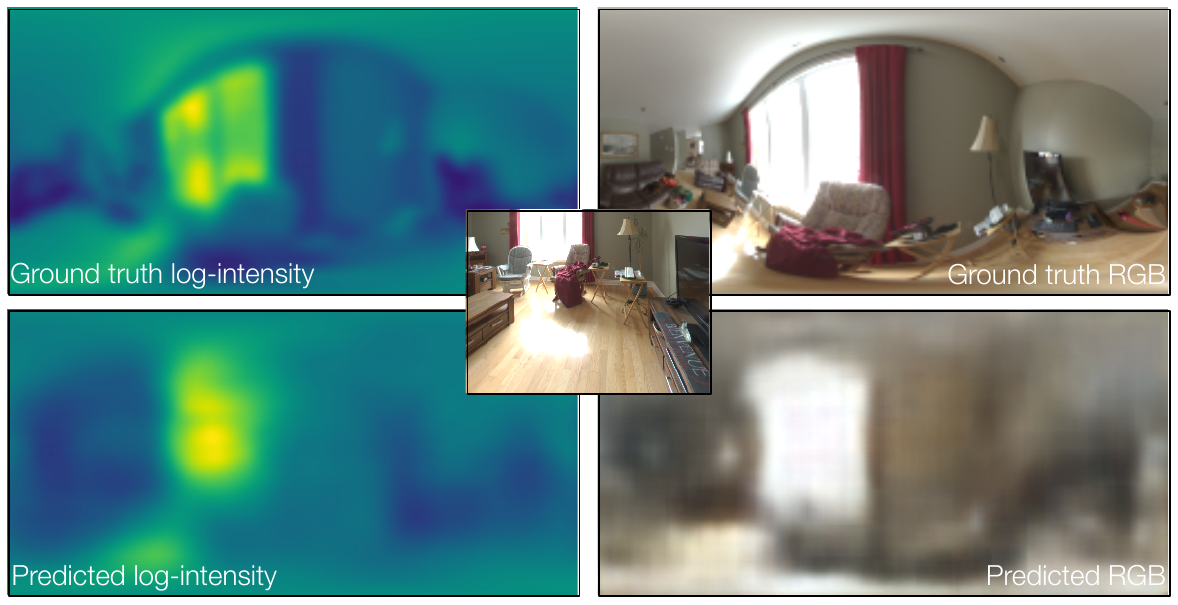} \\
(a) $p = 1, \mathcal{L} = 0.02$ & (b) $p = 10, \mathcal{L} = 0.04$ & (c) $p = 25, \mathcal{L} = 0.07$ \\*[.5em]
\includegraphics[width=.33\linewidth]{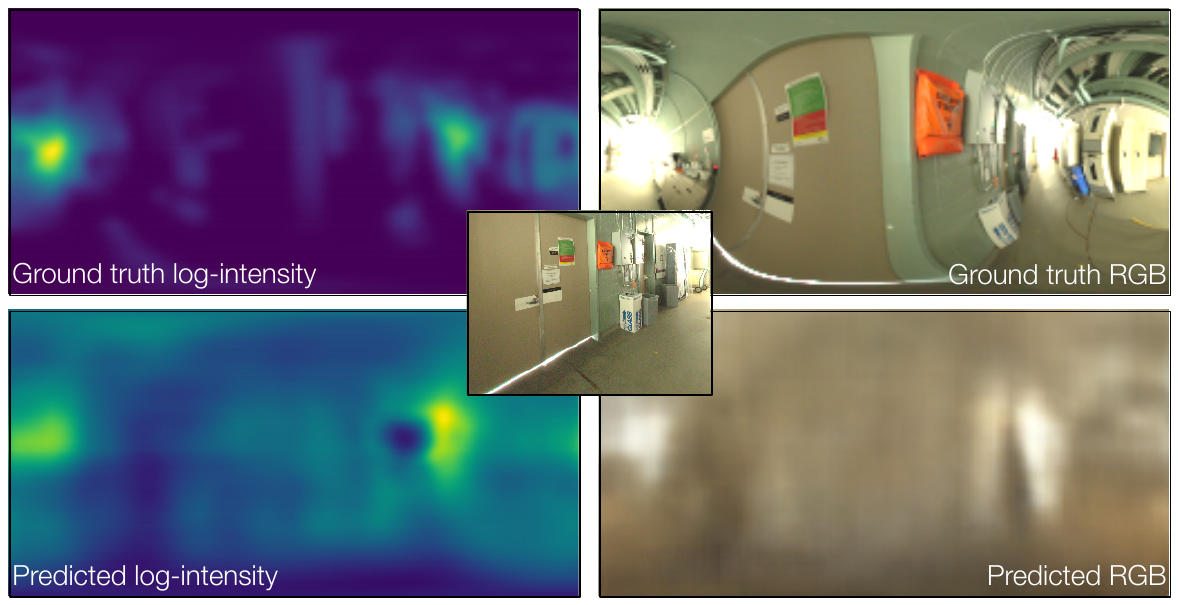} &
\includegraphics[width=.33\linewidth]{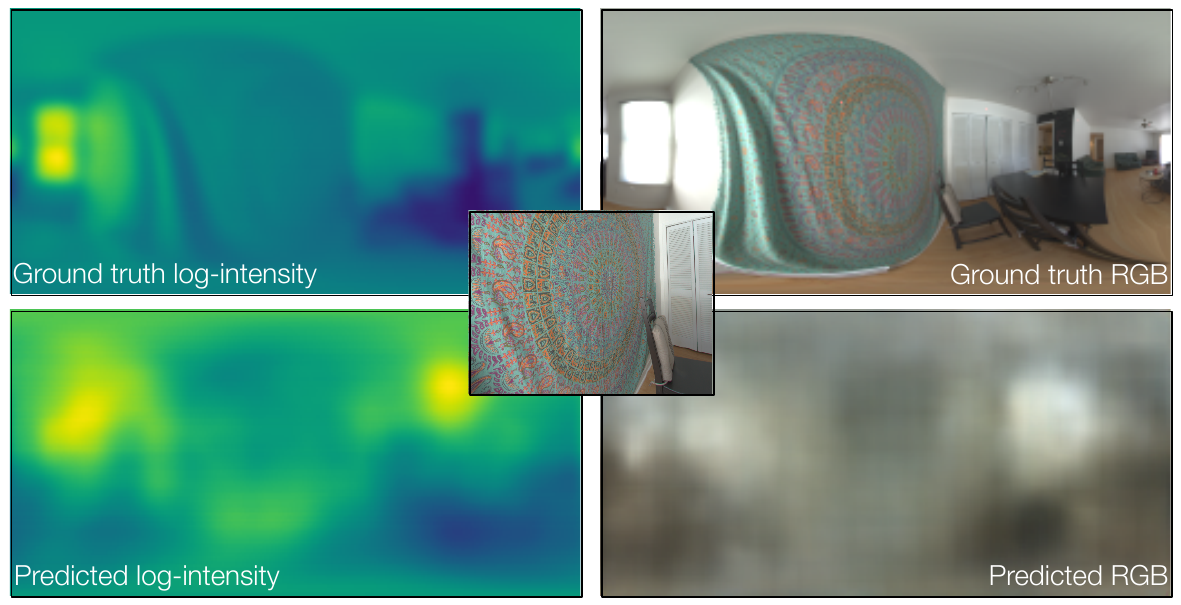} &
\includegraphics[width=.33\linewidth]{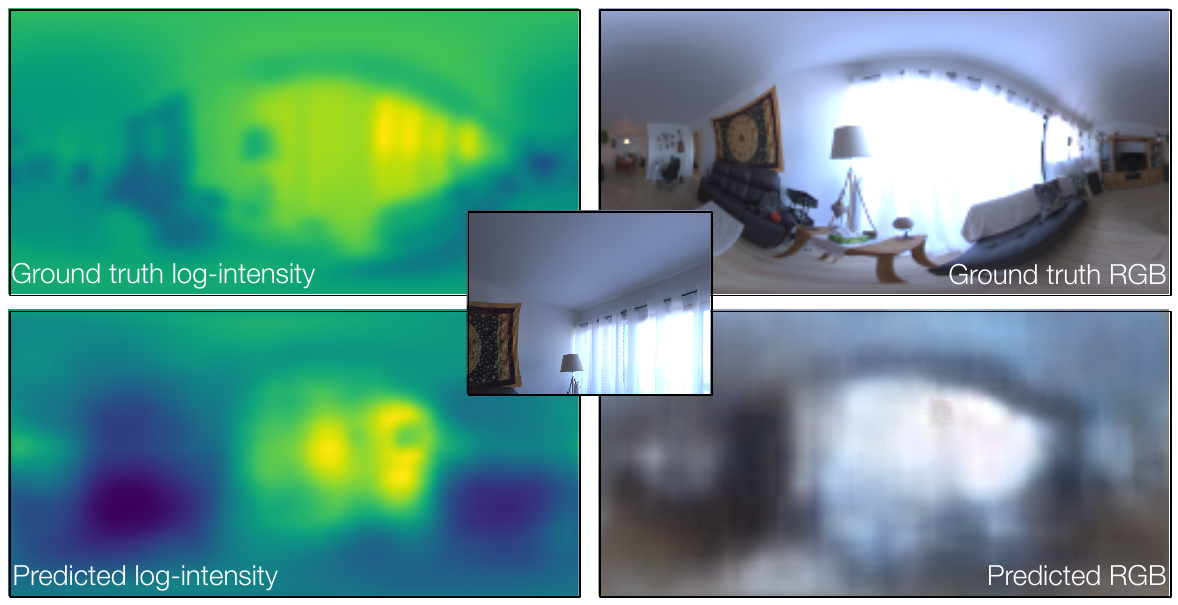} \\
(c) $p = 50, \mathcal{L} = 0.13$ & (d) $p = 75, \mathcal{L} = 0.27$ & (e) $p = 100, \mathcal{L} = 3.1$
\end{tabular}
\caption{Qualitative light intensity and RGB predictions on examples from the HDR test set. For each example, we show (middle) the input image, (top) the ground truth log-intensity $\mathbf{t}_\text{int}$ and RGB panoramas $\mathbf{t}_\text{RGB}$, and (bottom) the corresponding predictions from the HDR network $\mathbf{x}_\text{int}$ and $\mathbf{x}_\text{RGB}$. Light intensities are color-coded from yellow (high intensity) to blue (low intensity). The examples are sorted by error percentile $p$ on the loss (\ref{e:hdrloss}) from top-left to bottom-right in reading order. See fig.~\ref{f:results-loss-histogram} for the complete error distribution on the test set.} 
\label{f:results-loss-qualitative}
\end{figure*}

We begin by evaluating the performance of the HDR network on the HDR test set. Fig.~\ref{f:results-loss-histogram} shows the distribution of the HDR losses (eq.~\ref{e:hdrloss}) on our test set of 2,100 images (15\% of 14,000), and fig.~\ref{f:results-loss-qualitative} shows qualitative examples corresponding to various percentiles of this distribution. In this figure, note that the range of ground truth log-intensities is $[0.04, 3.01]$. Since we use a L2 loss (mean squared error), a value of $\mathcal{L} = 0.02$ (as in fig.~\ref{f:results-loss-qualitative}-(a)) thus indicates a global relative error of around $4.7\%$.

Even though the network has trouble in pinpointing the exact location of small, concentrated light sources (e.g. fig.~\ref{f:results-loss-qualitative}-(b)), it succeeds in finding the dominant light sources in the scene, even when they are located outside the field of view of the photo (as is the case in most of the examples of fig.~\ref{f:results-loss-qualitative}). Larger errors typically occur when the scene is lit by a very large area light sources, such as the window in fig.~\ref{f:results-loss-qualitative}-(e). 

\paragraph{Virtual object relighting} 

The HDR network output can directly be used to generate an HDR environment map suitable for relighting, by combining its two outputs like so: 
\begin{equation}
\mathbf{x}_\text{combined} = 10^{\mathbf{x}_\text{mask}} + \mathbf{x}_\text{RGB} \,,
\label{e:hdr-envmap}
\end{equation}
since the HDR network is trained to output \emph{log}-intensity. As a post-process we matched the mean RGB value of the RGB prediction and the color of the light source to the mean RGB value of the input image (following the assumption of Gray World surface reflectance, this value captures the color of the illumination).

\begin{figure*}
\centering
\footnotesize
\setlength{\tabcolsep}{1pt}
\begin{tabular}{ccccc}
\includegraphics[width=.195\linewidth]{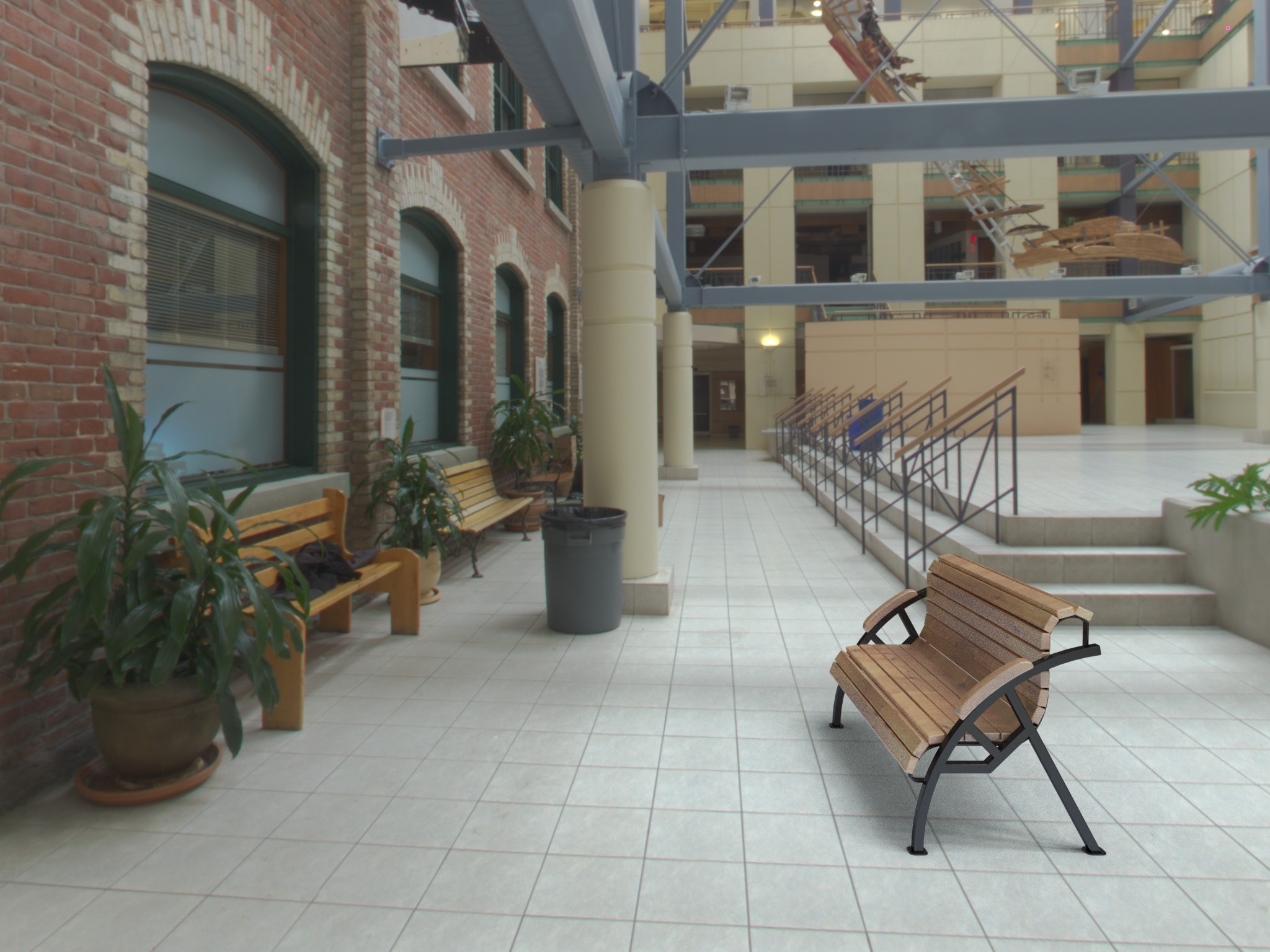} & 
\includegraphics[width=.195\linewidth]{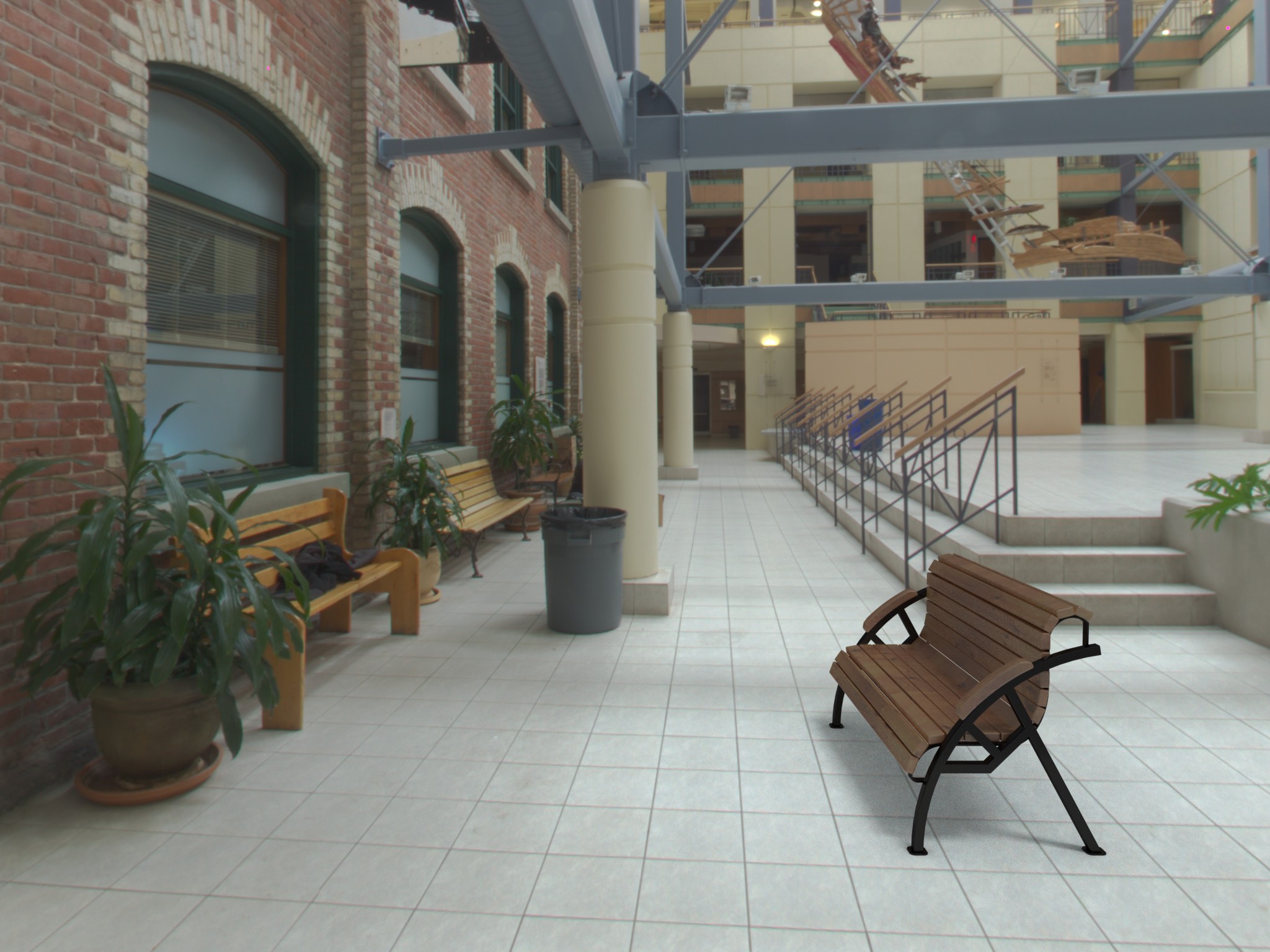} & 
\includegraphics[width=.195\linewidth]{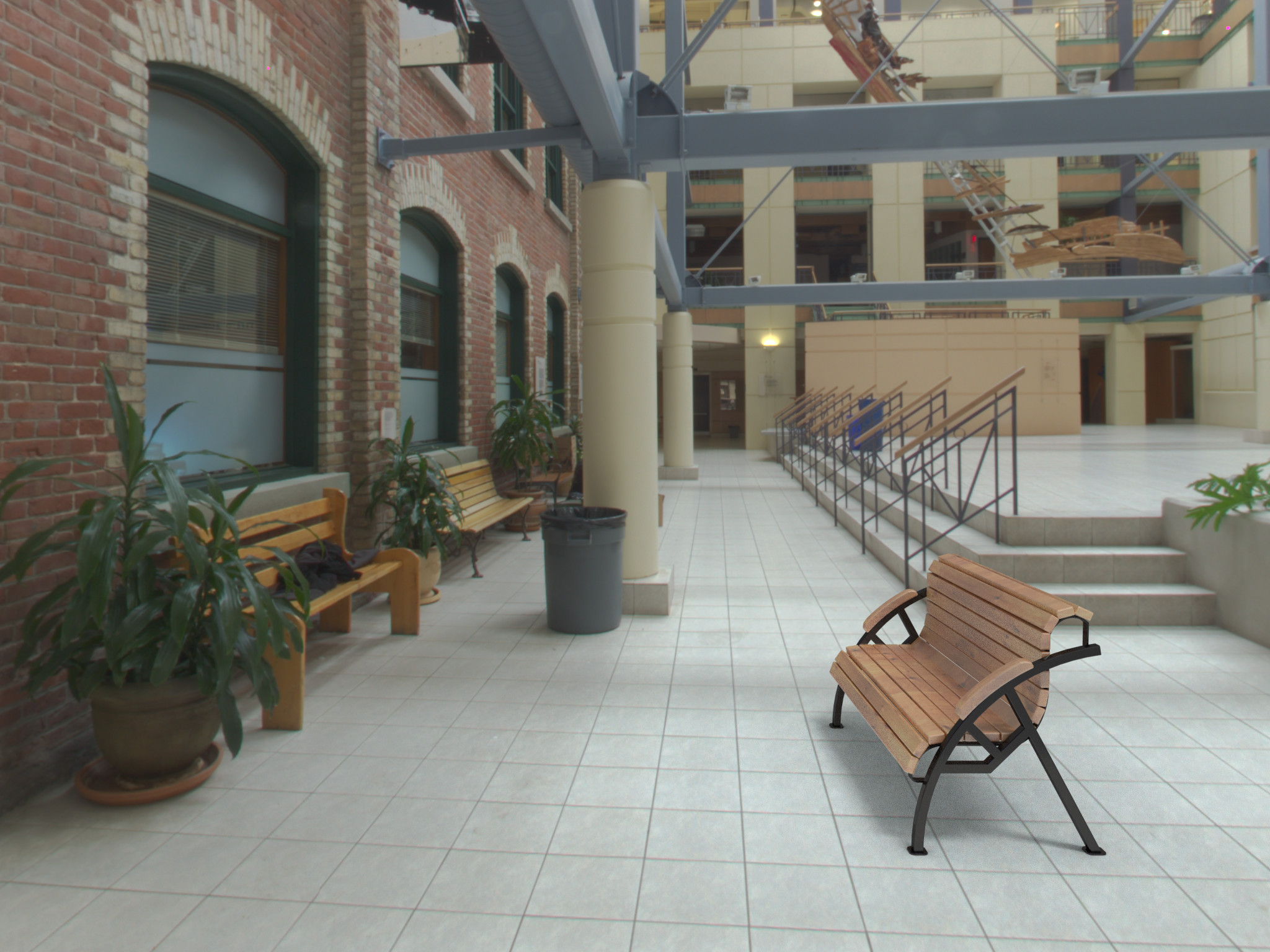} & 
\includegraphics[width=.195\linewidth]{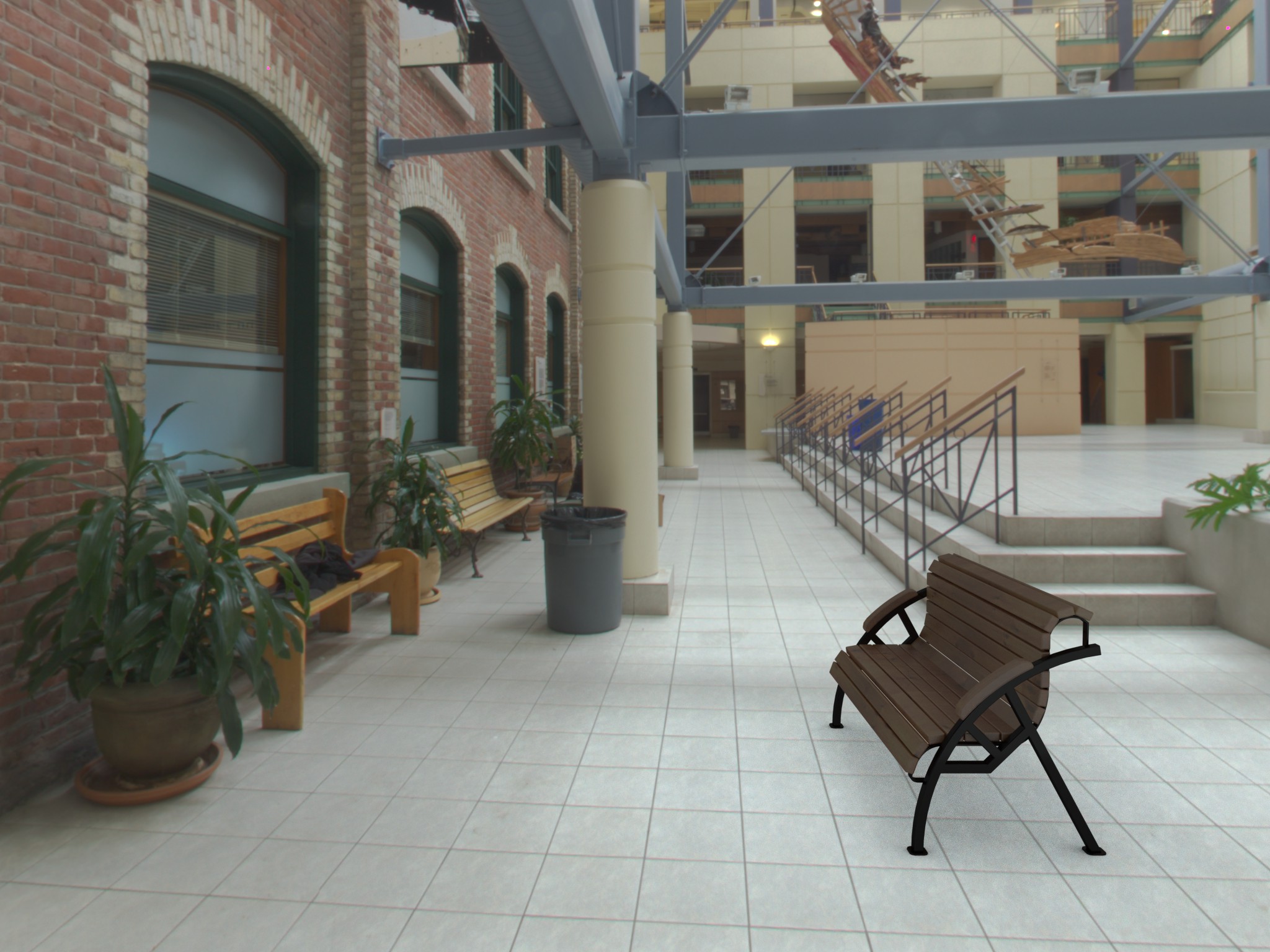} & 
\includegraphics[width=.195\linewidth]{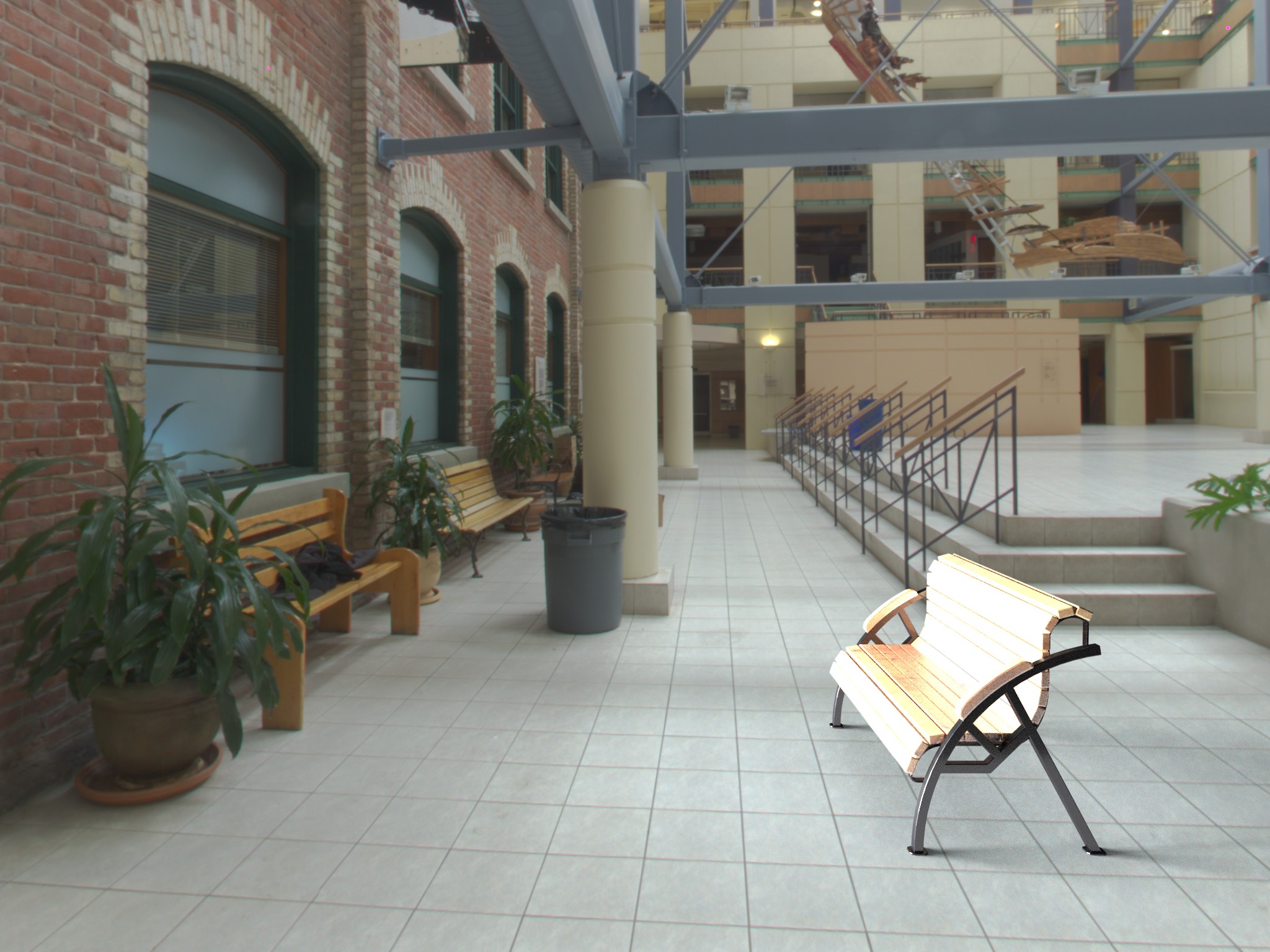} \\
\includegraphics[width=.195\linewidth]{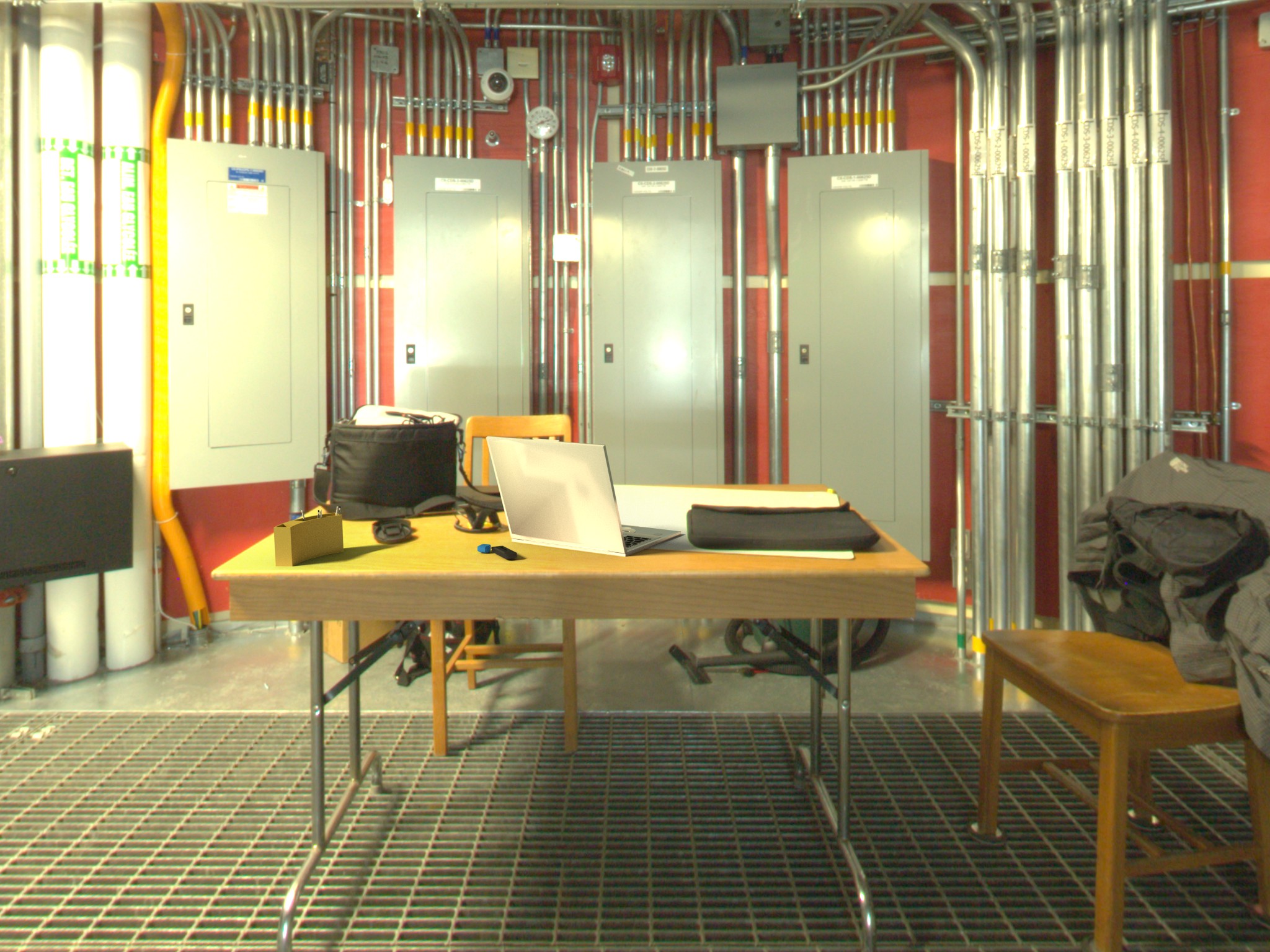} & 
\includegraphics[width=.195\linewidth]{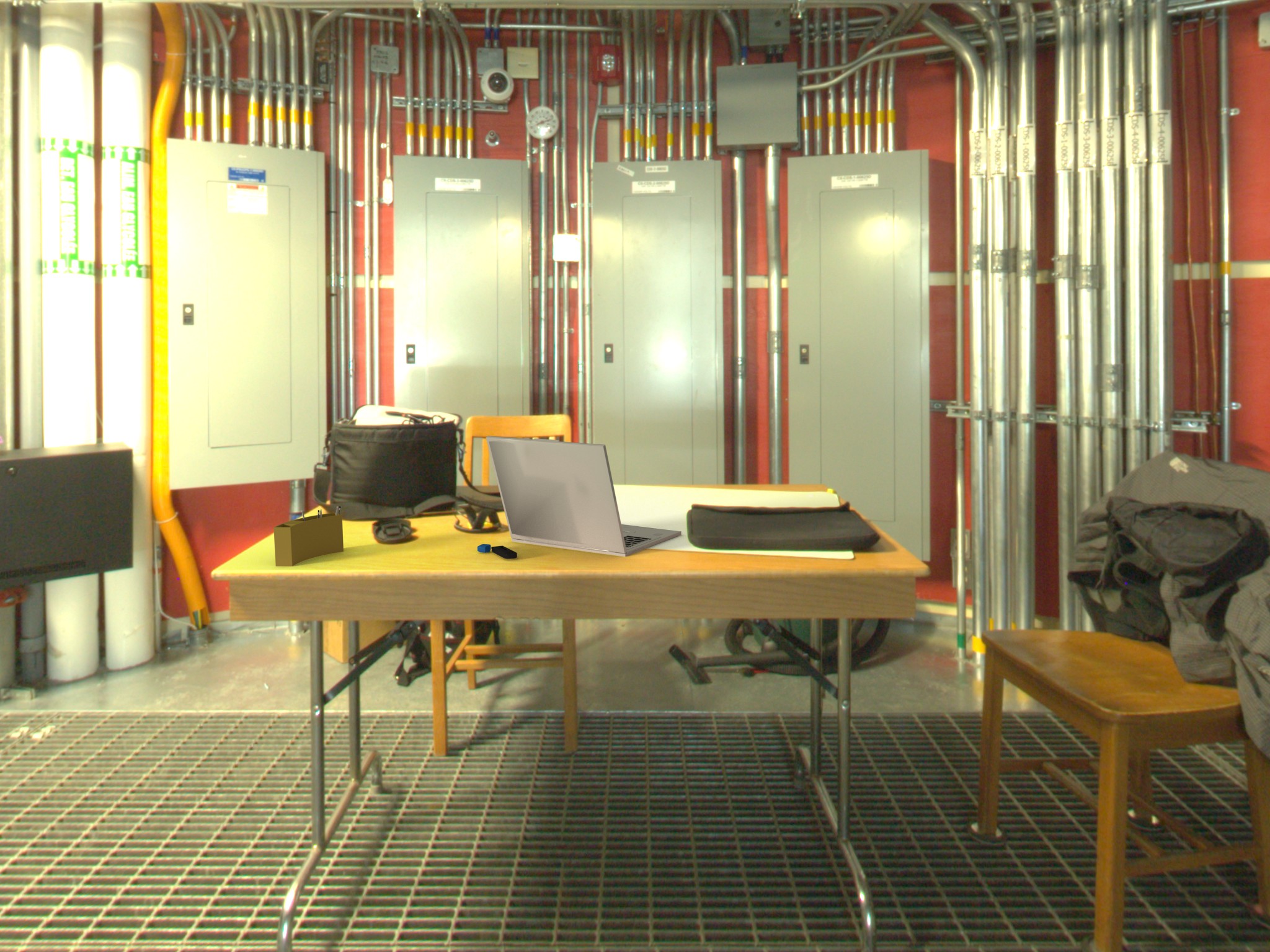} & 
\includegraphics[width=.195\linewidth]{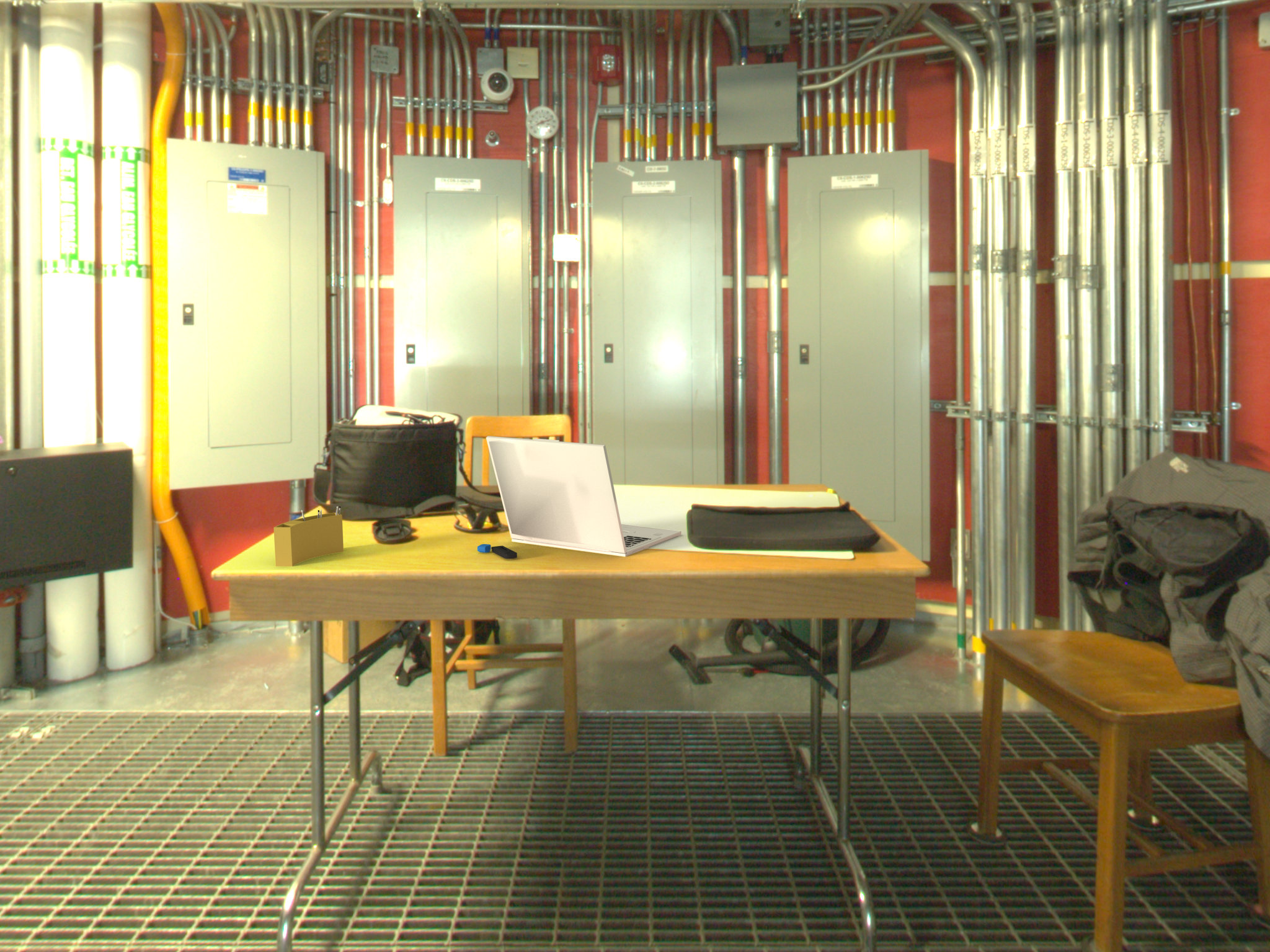} & 
\includegraphics[width=.195\linewidth]{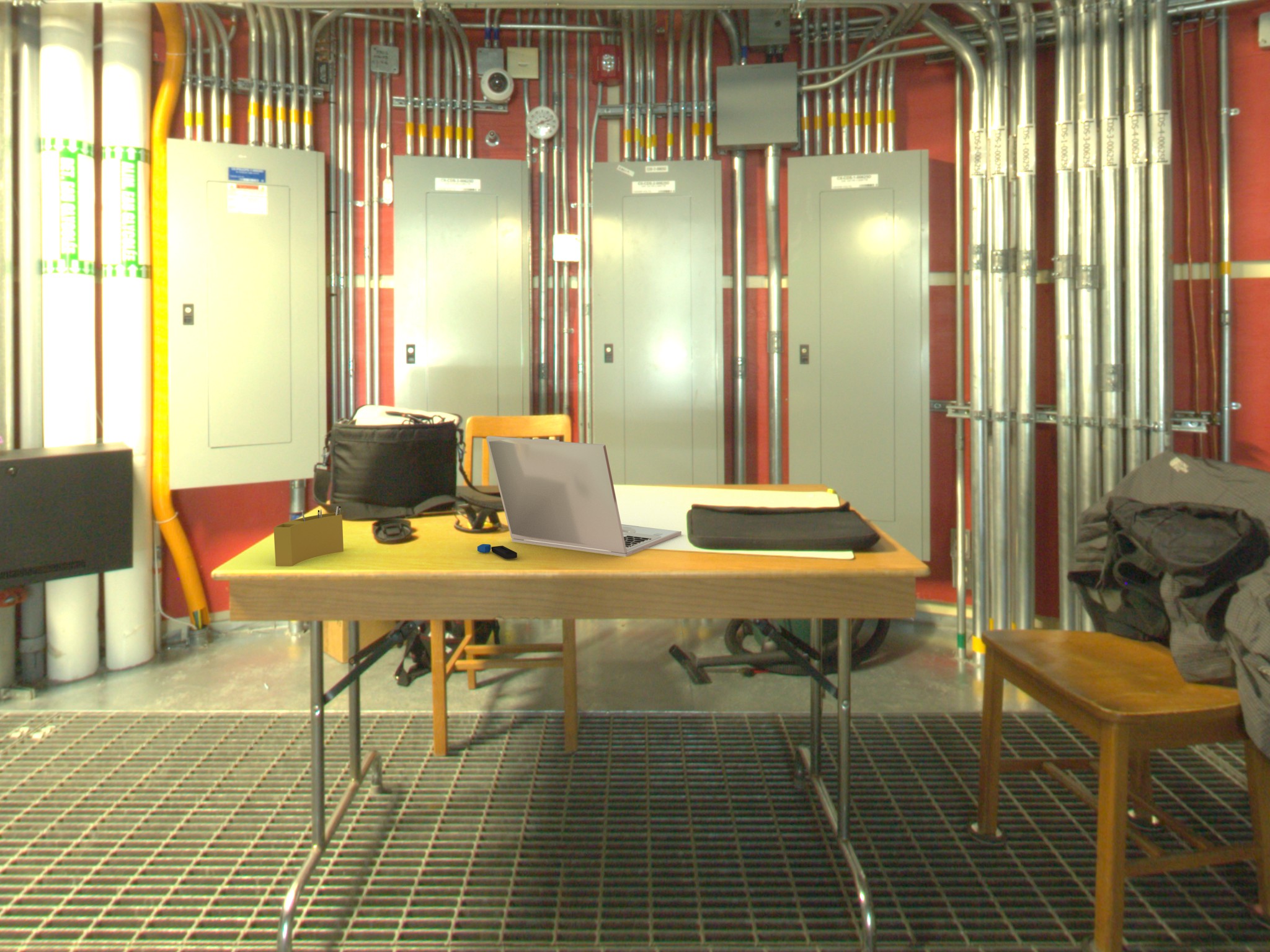} & 
\includegraphics[width=.195\linewidth]{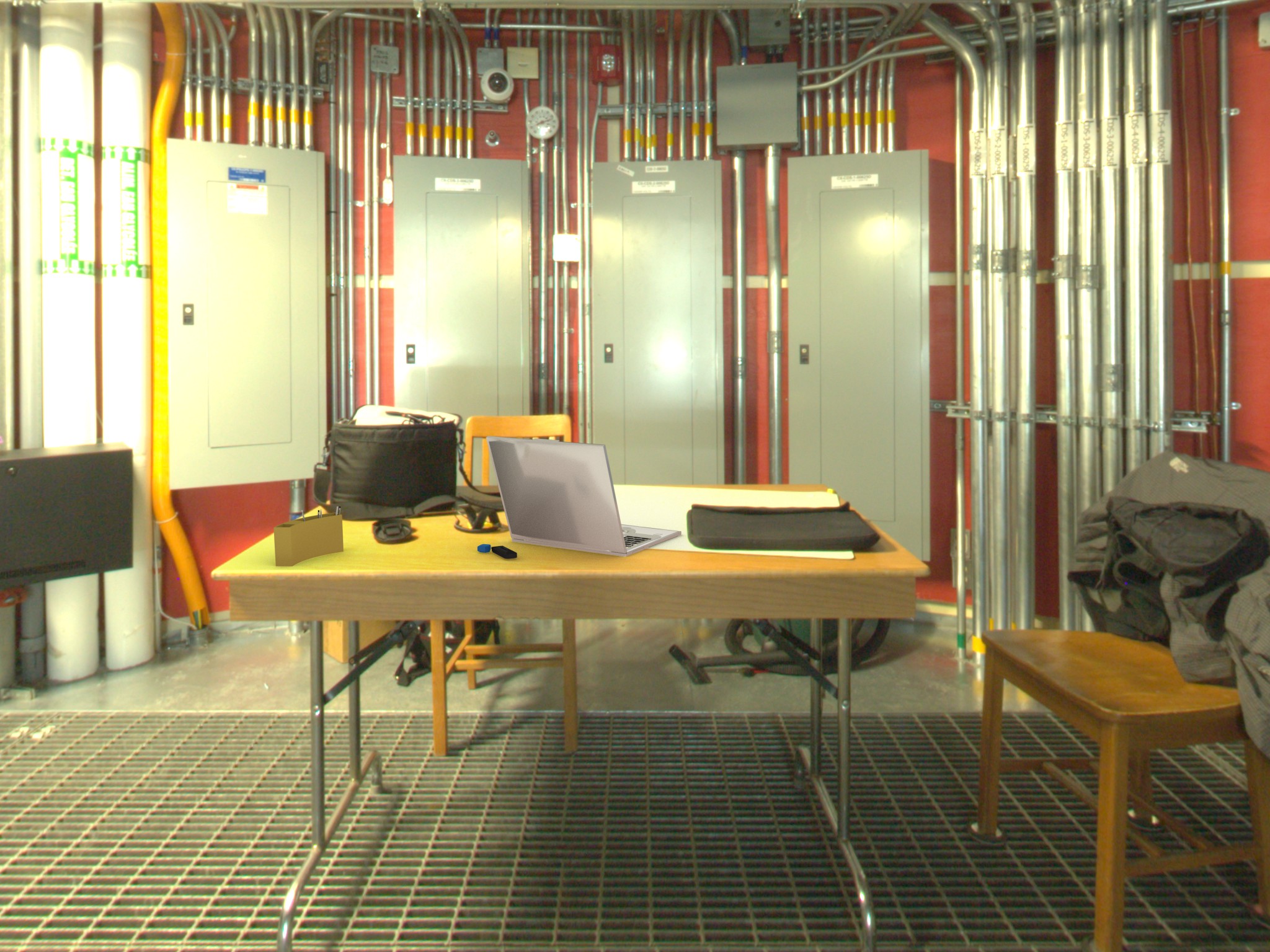} \\
\includegraphics[width=.195\linewidth]{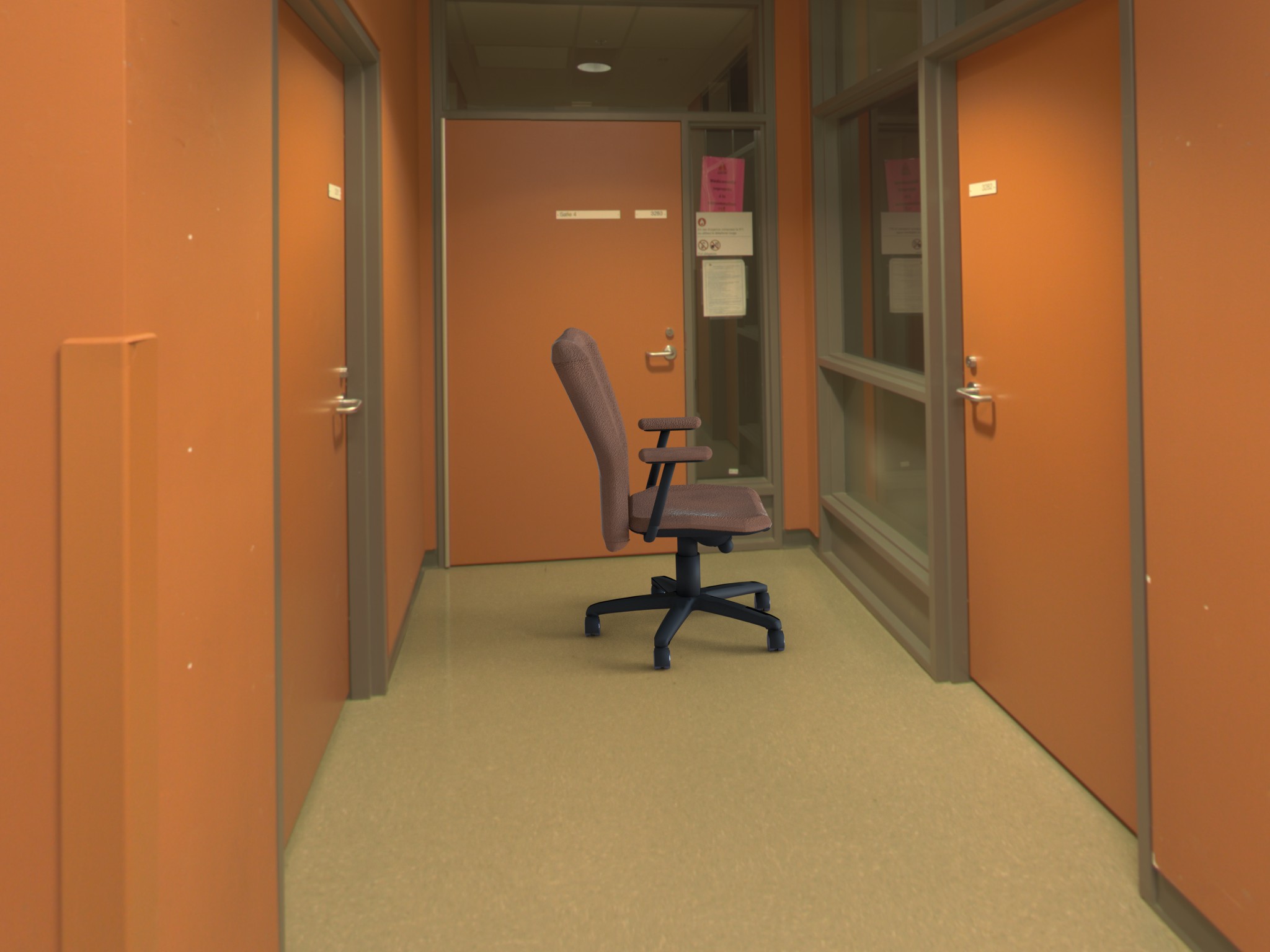} & 
\includegraphics[width=.195\linewidth]{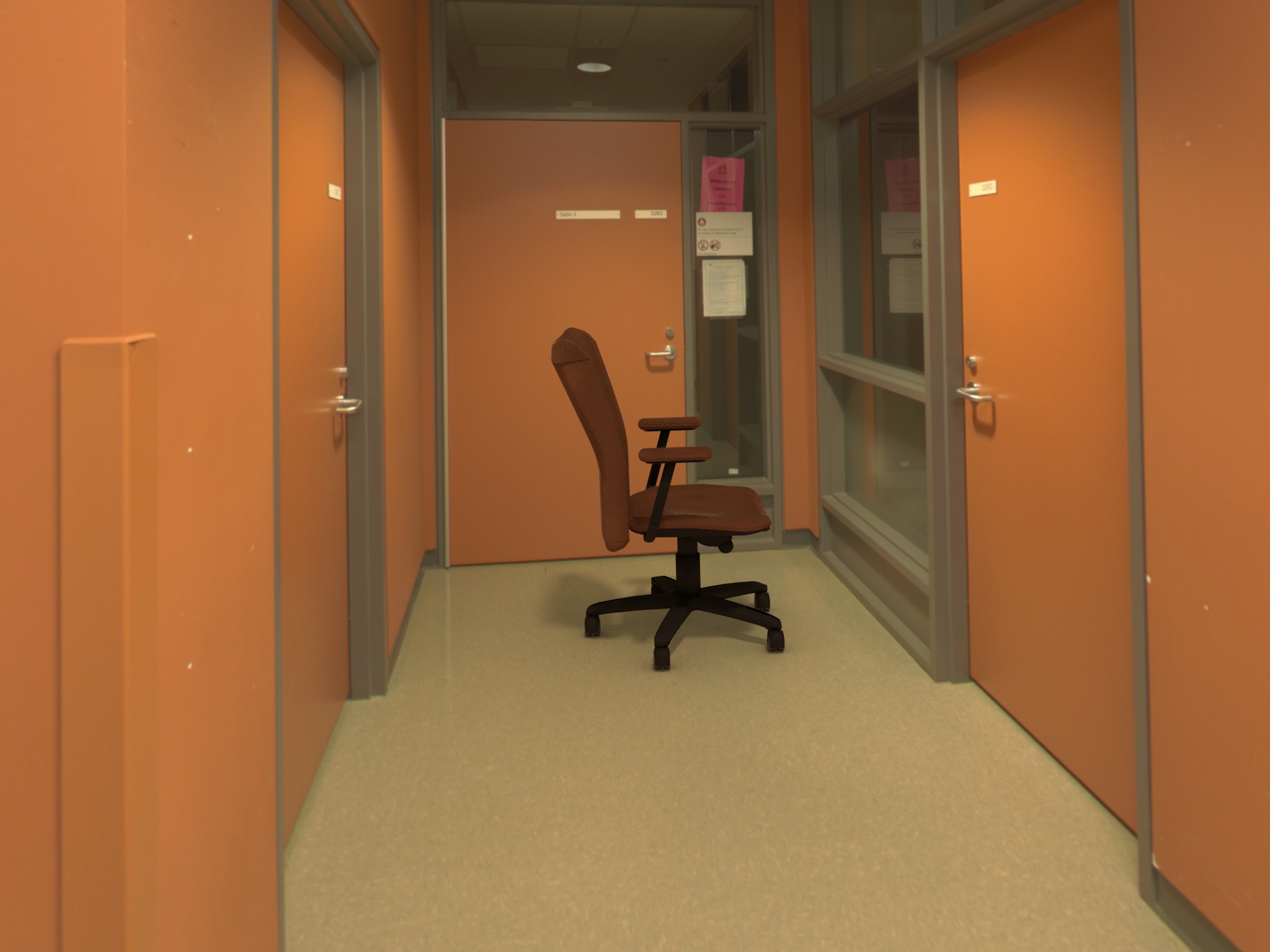} & 
\includegraphics[width=.195\linewidth]{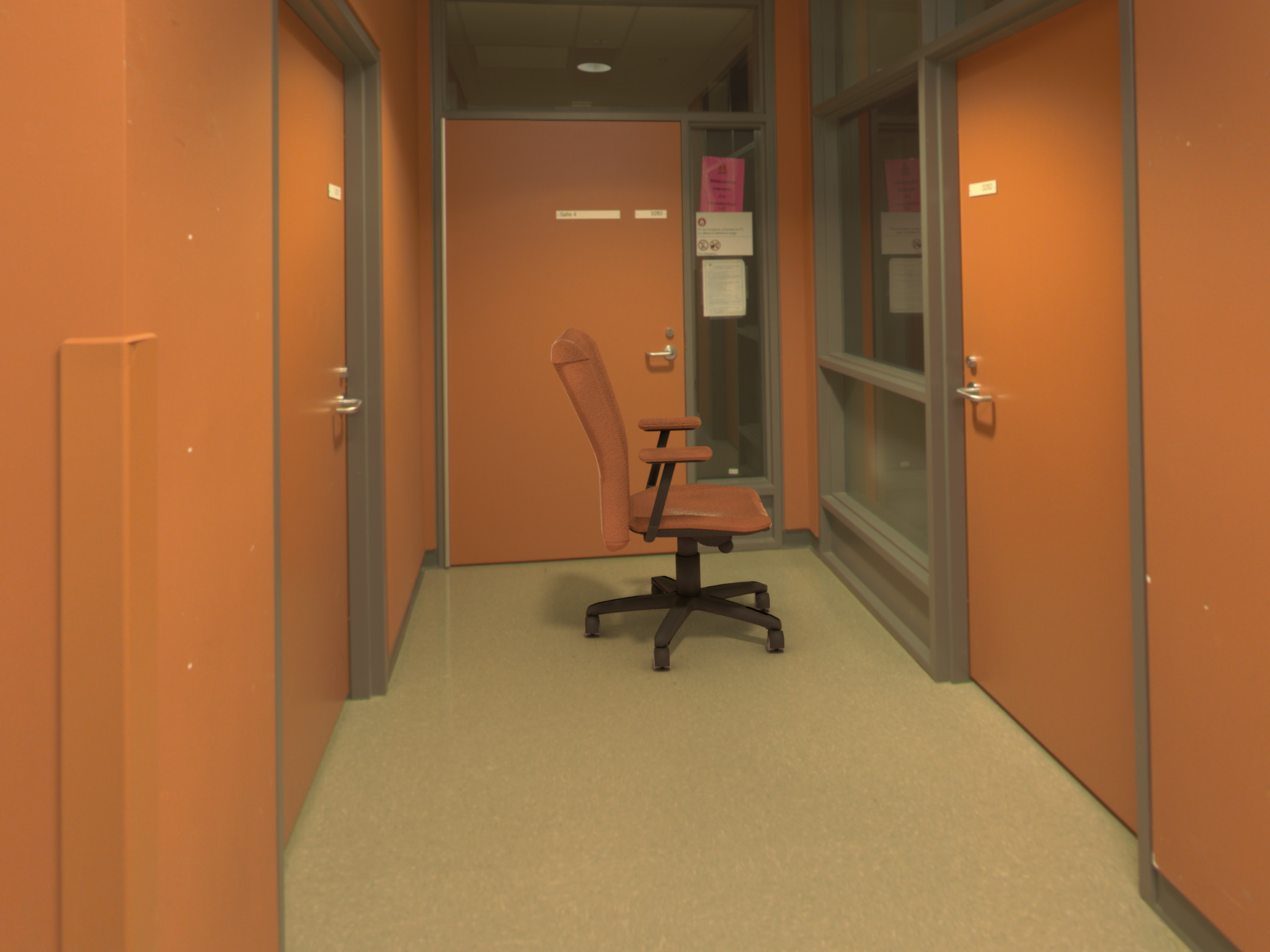} & 
\includegraphics[width=.195\linewidth]{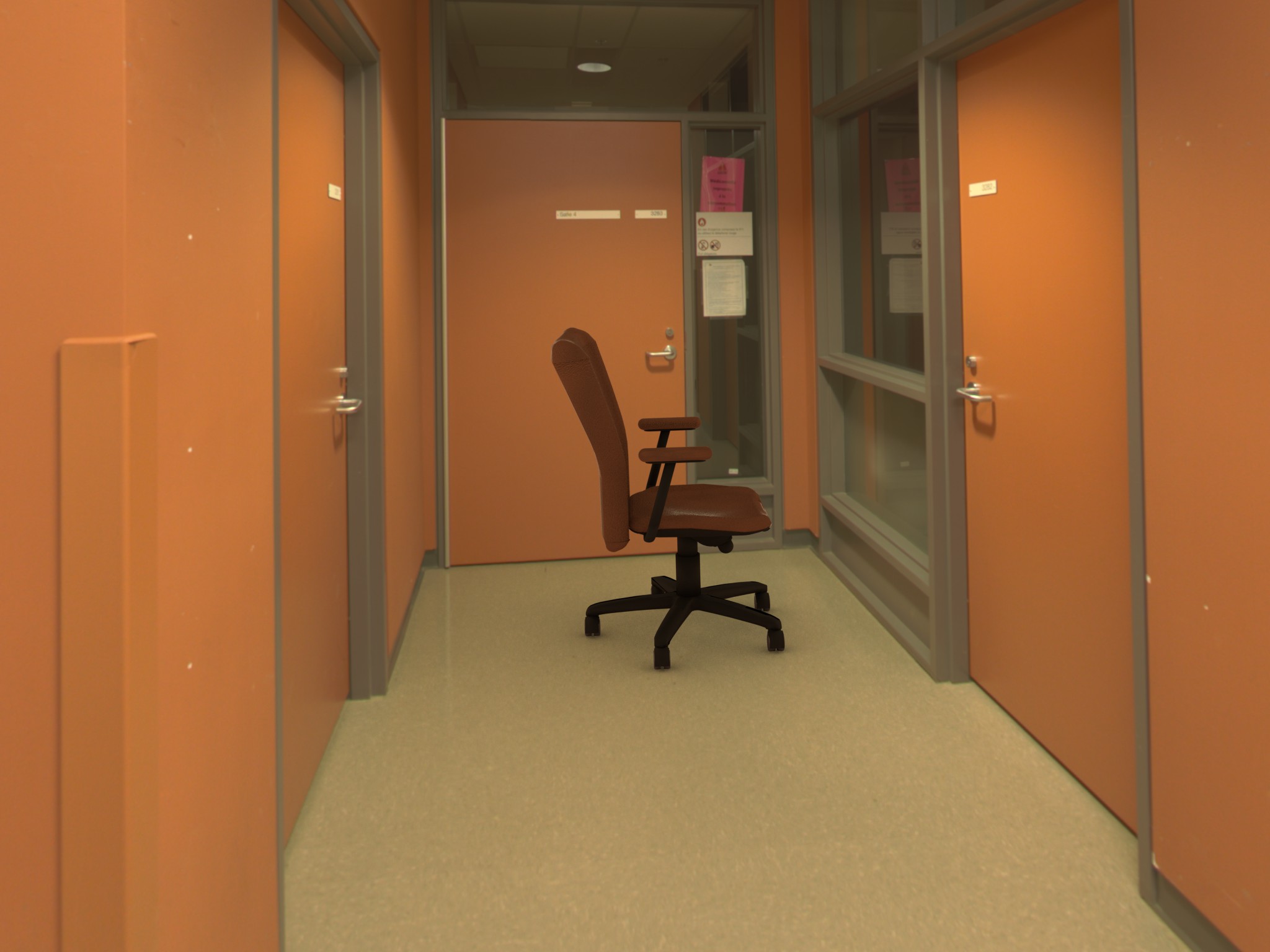} & 
\includegraphics[width=.195\linewidth]{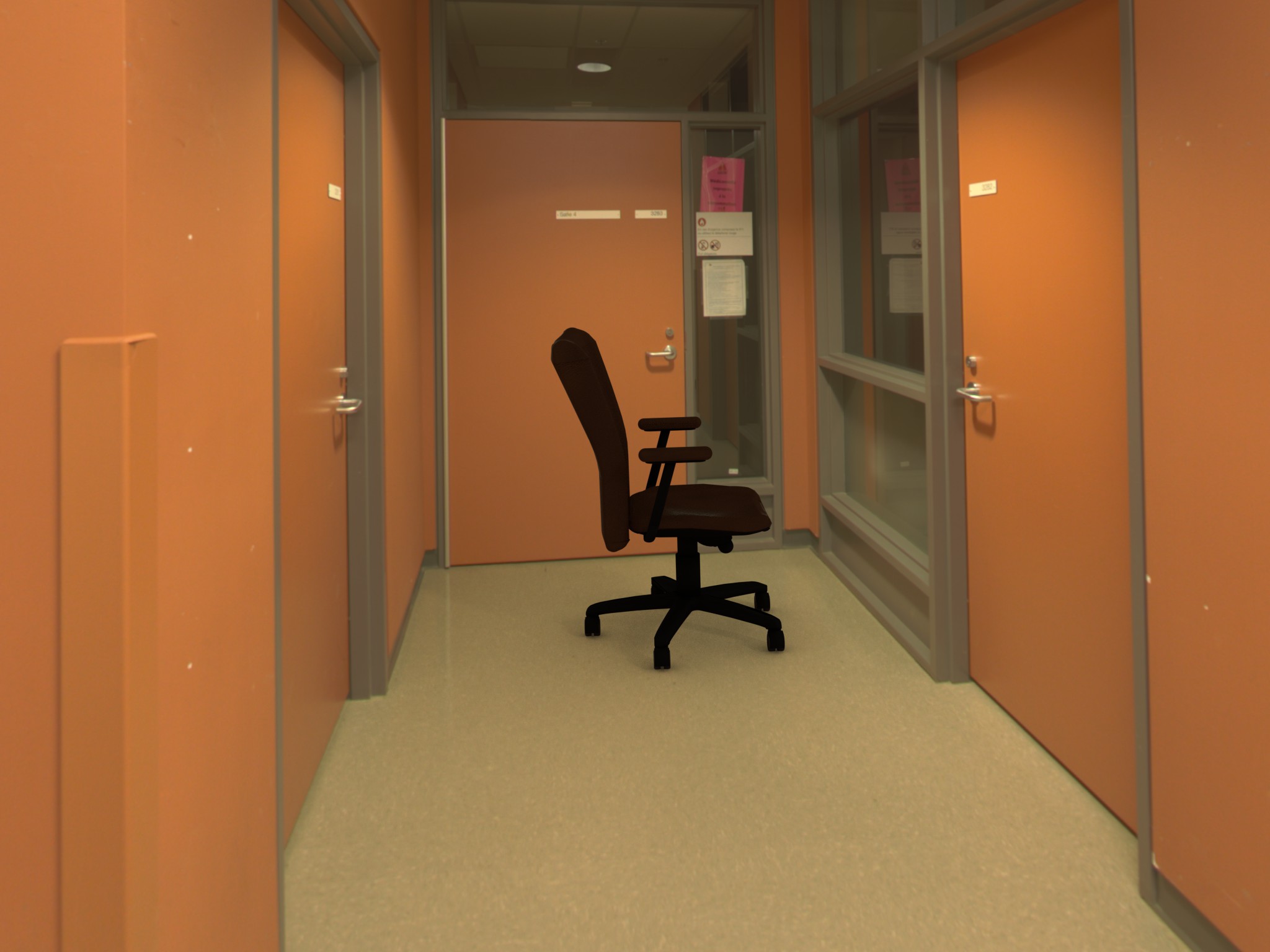} \\
%
(a) Ground truth lighting &
(b) Our HDR network & 
(c) HDR network, intensity tuned & 
(d) \cite{khan-siggraph-06} & 
(e) \cite{karsch-tog-14}
\end{tabular}
\caption[]{Comparison of (b) our method and (c) our method with a single intensity factor humanly tuned with (a) ground truth lighting, (c) \cite{khan-siggraph-06} and (d) \cite{karsch-tog-14} on virtual object relighting. While our results sometimes visually differ from ground truth, they yield realistic object insertion results. In contrast, Khan et al.~\shortcite{khan-siggraph-06} do not estimate HDR lighting, so renders look flat. Since Karsch et al.~\shortcite{karsch-tog-14} rely on intrinsic decomposition, geometry estimation and inverse lighting, we found the method to be quite sensitive to errors in any one of these steps. Therefore, renders are often much too bright or dark. More results available in the supplementary material.}
\label{f:results-comparison}
\end{figure*}

Fig.~\ref{f:results-comparison} presents virtual objects inserted in crops extracted from our HDR test set, so that they can be compared with relighting with their ground truth lighting. The network predictions yields convincing relighting results that are very close to ground truth.

\paragraph{Global intensity scaling}

Recovering the \emph{absolute} illumination intensities from an uncalibrated LDR image is an ill-posed problem, since many combinations of light intensities and camera parameters (shutter speed, ISO, etc.) may result in the exact same image. Thus, an object lit with our network estimate is sometimes too dark or too bright. Given that the network recovers correct \emph{relative} illumination---that is, the ratio between intensities in different parts of the panorama is accurate---fixing this issue boils down to the selection of a \emph{single}, global intensity scaling parameter. This parameter can be easily specified and is often one of the controls most production compositors offer users to ensure that their renders are properly exposed (even when relighting with ground truth IBLs). 

In addition to the automatically estimated results in fig.~\ref{f:results-comparison}, we also provide a set of results (fig.~\ref{f:results-comparison}-(c)) where this global intensity scaling has been manually specified. Many more results can be found in the supplementary material, including an example of the effect of this scaling parameter on the appearance of a composite. While our automatic estimate is reasonable in many cases, slightly tuning this scale factor can produce compelling results in almost all cases. Note that all other results in this paper, including comparisons with previous methods, use our \emph{automatic} light estimates.

\begin{figure*}[!t]
\centering
\footnotesize
\setlength{\tabcolsep}{1pt}
\begin{tabular}{cccc}
\includegraphics[width=.244\linewidth]{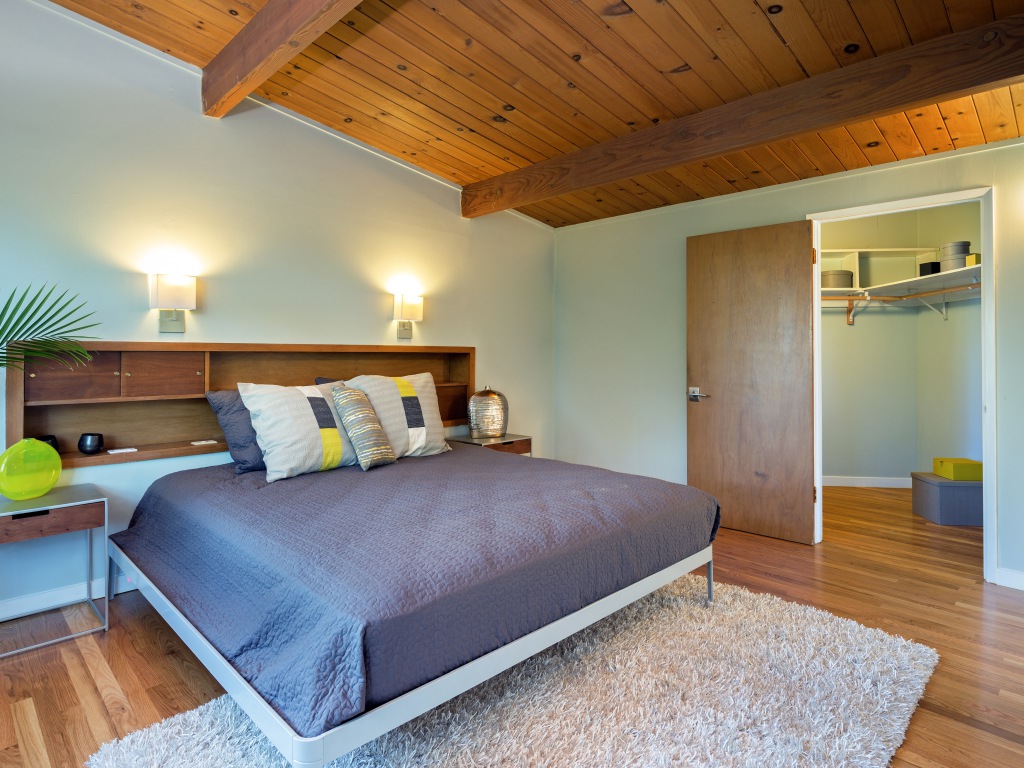} & 
\includegraphics[width=.244\linewidth]{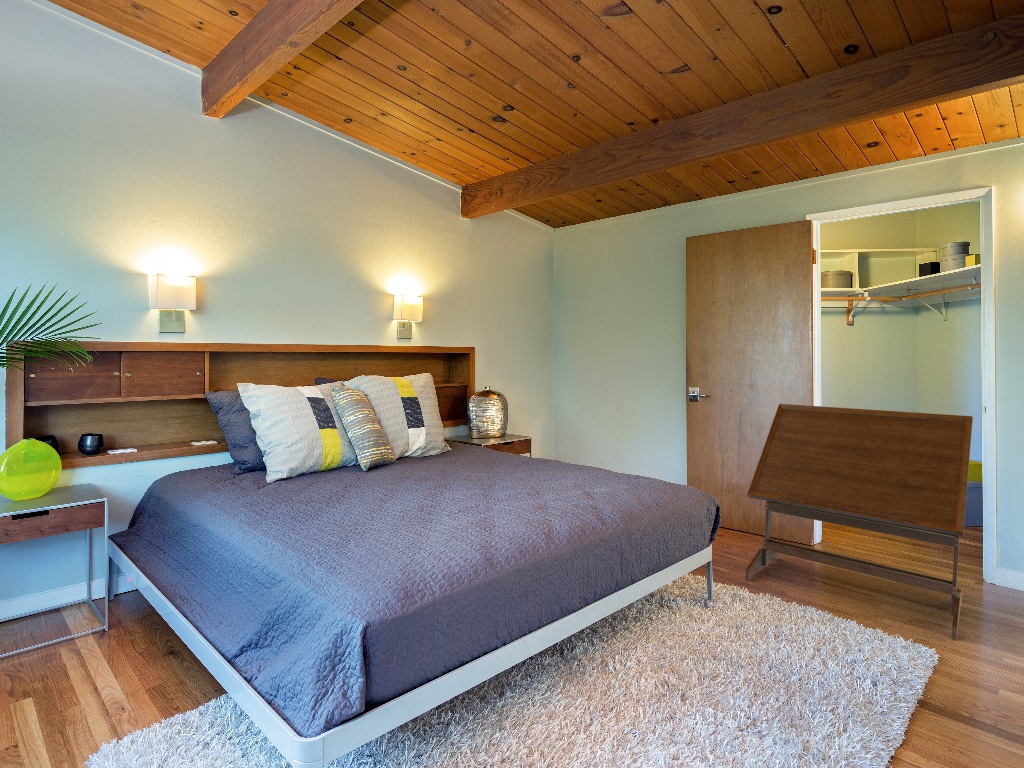} & 
\hspace{.4em}
\includegraphics[width=.244\linewidth]{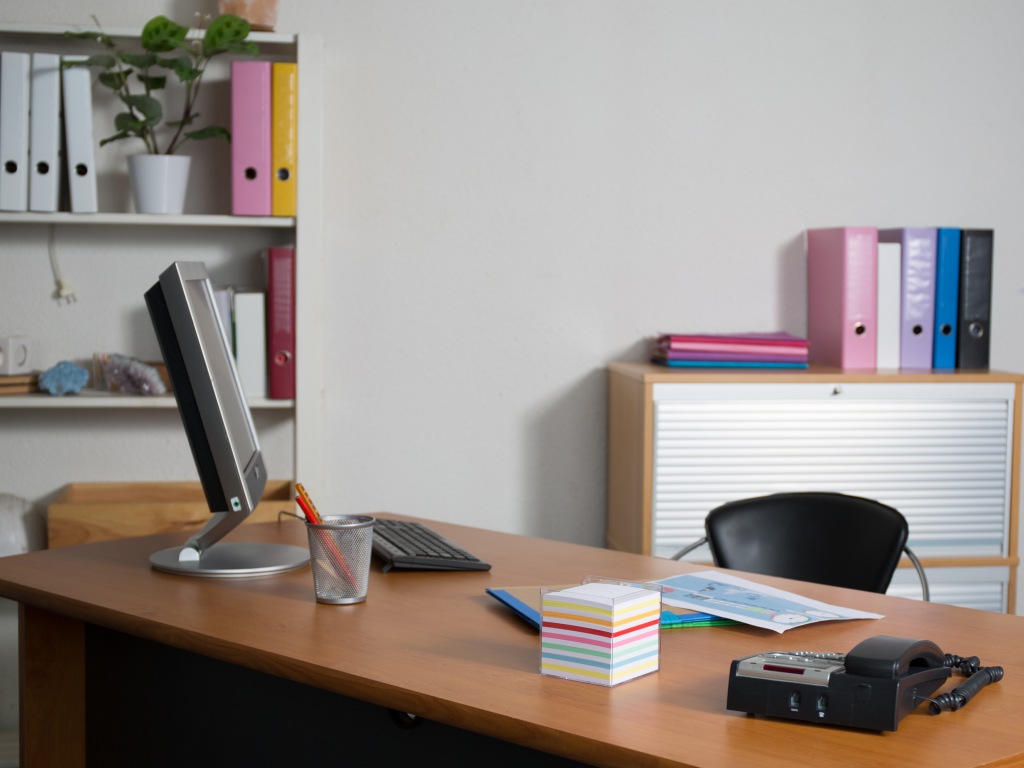} & 
\includegraphics[width=.244\linewidth]{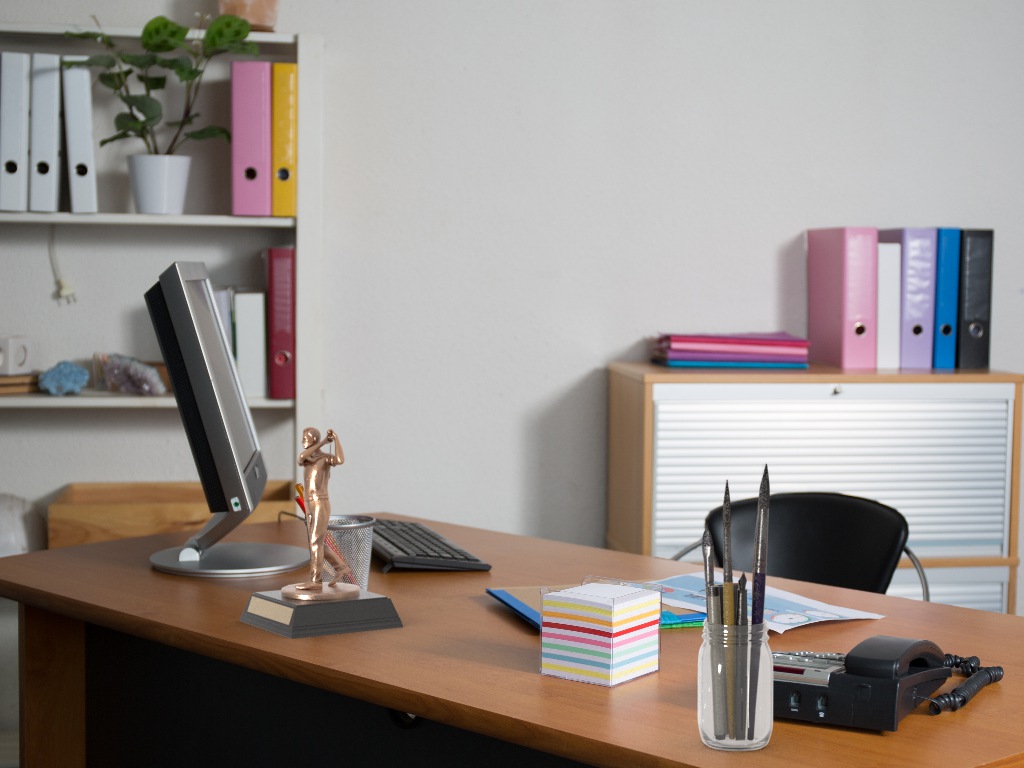}\vspace{.4em} \\
\includegraphics[width=.244\linewidth]{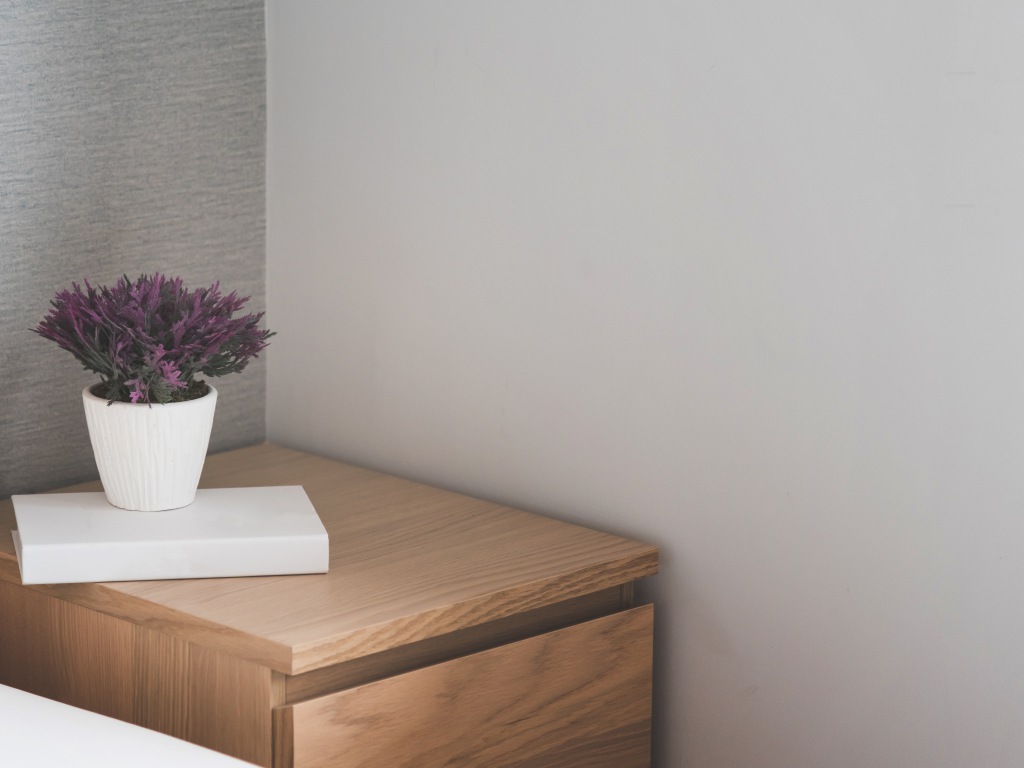} & 
\includegraphics[width=.244\linewidth]{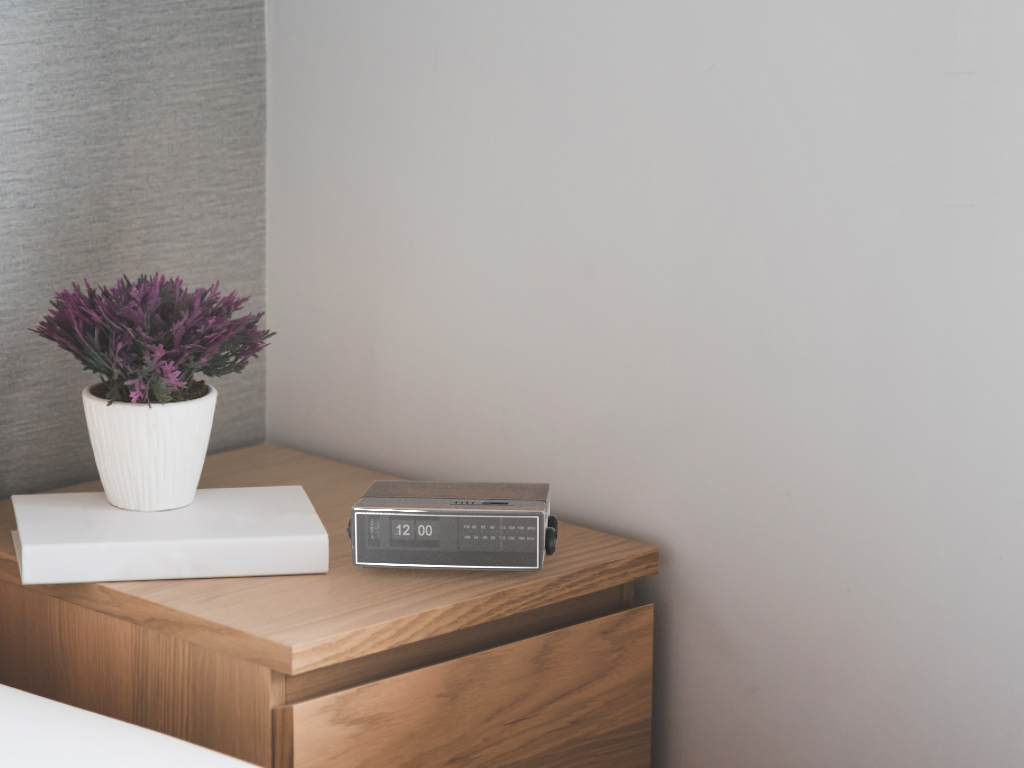} & 
\hspace{.4em}
\includegraphics[width=.244\linewidth]{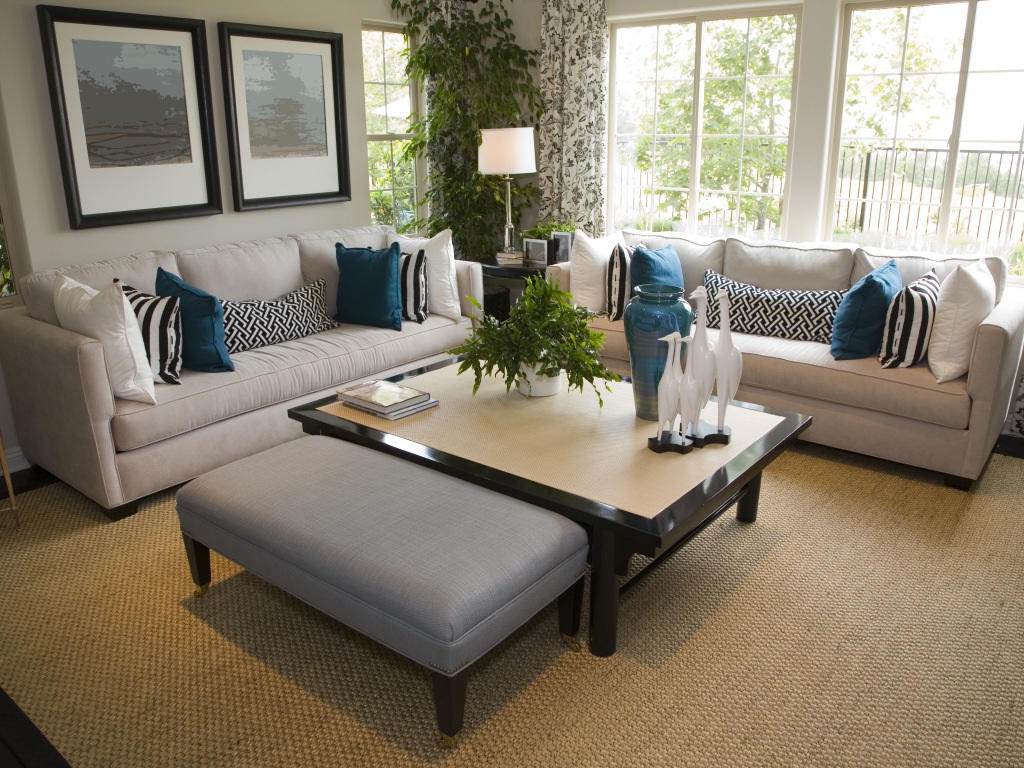} & 
\includegraphics[width=.244\linewidth]{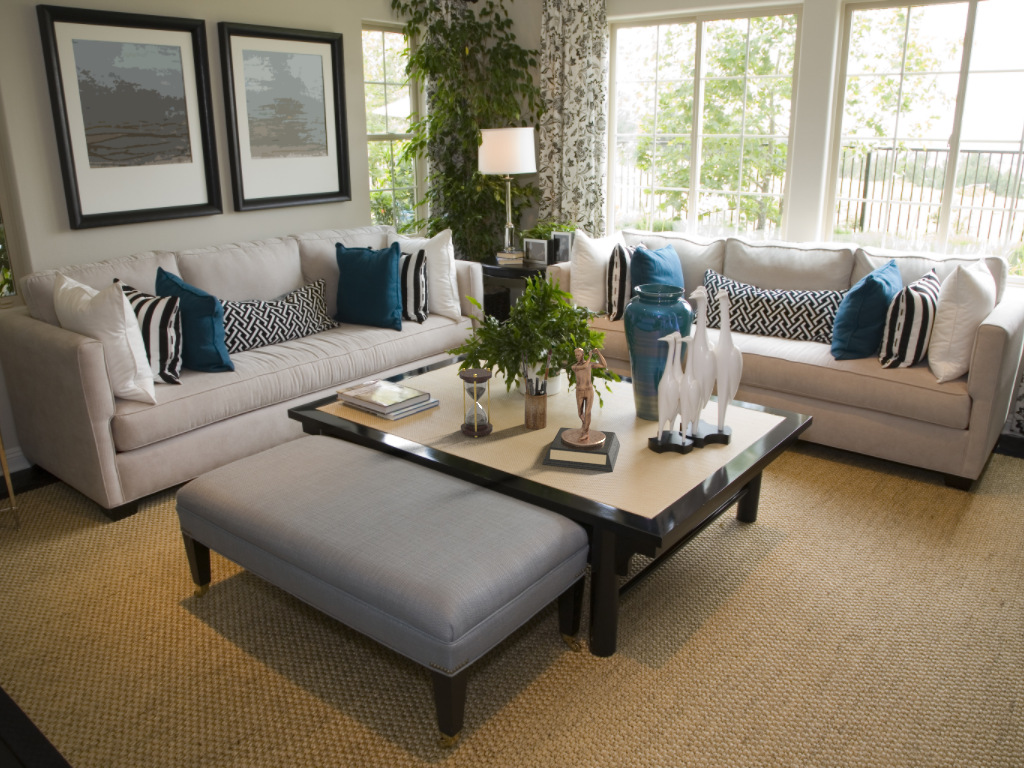}\vspace{.4em} \\
\includegraphics[width=.244\linewidth]{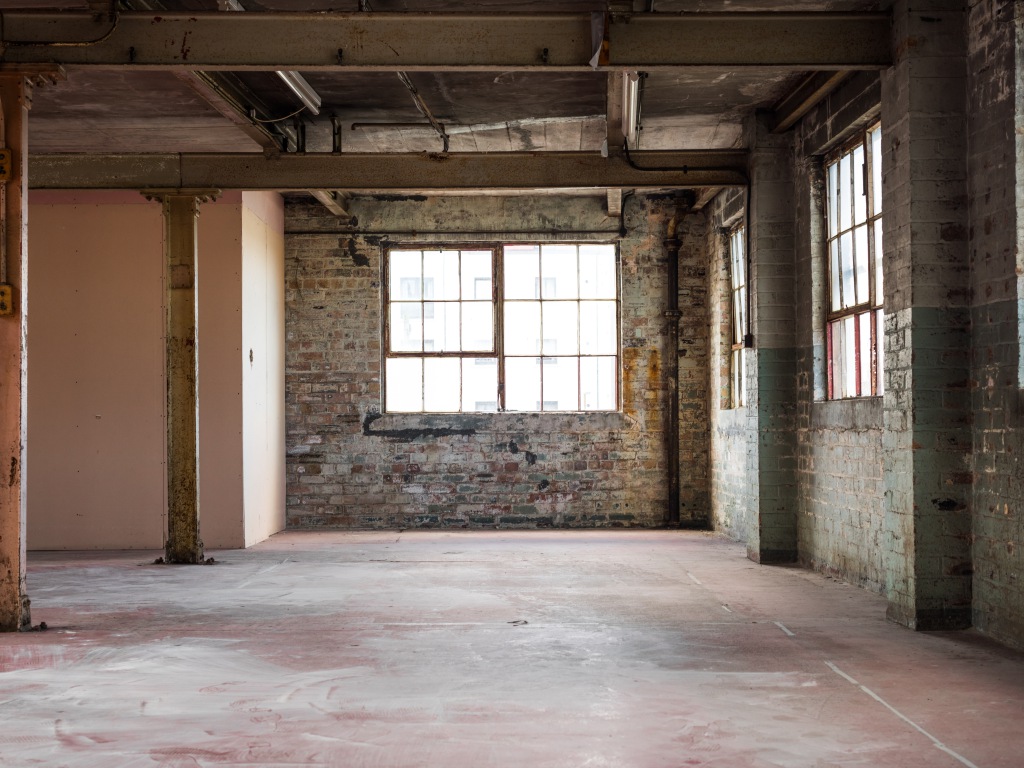} & 
\includegraphics[width=.244\linewidth]{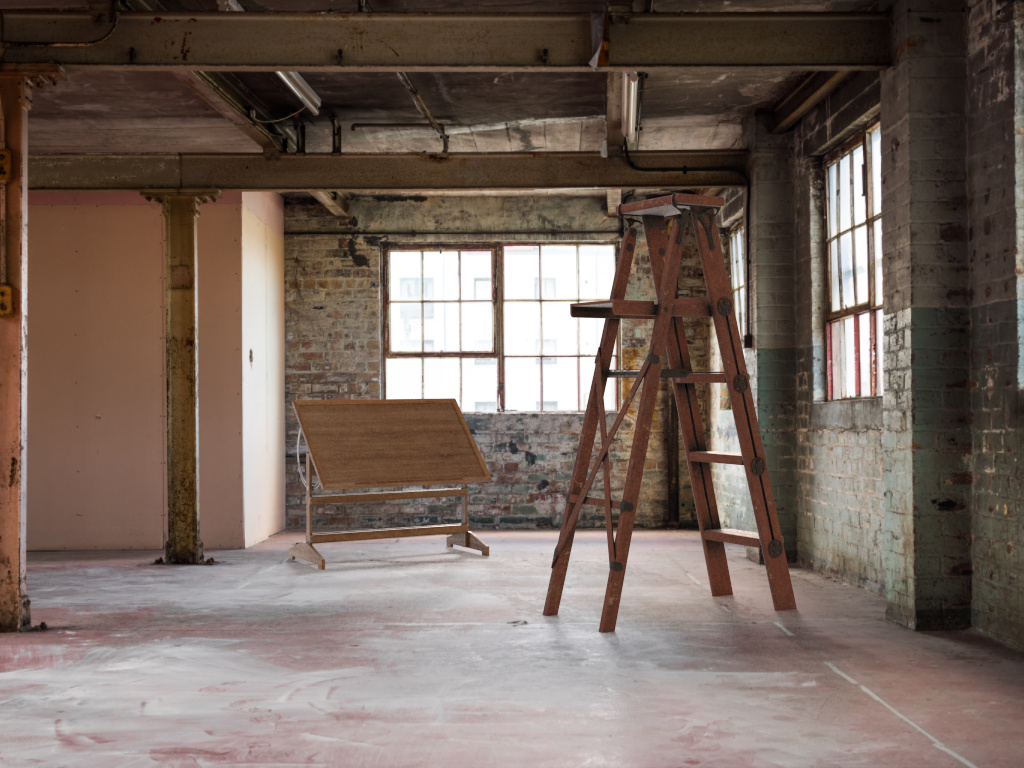} & 
\hspace{.4em}
\includegraphics[width=.244\linewidth]{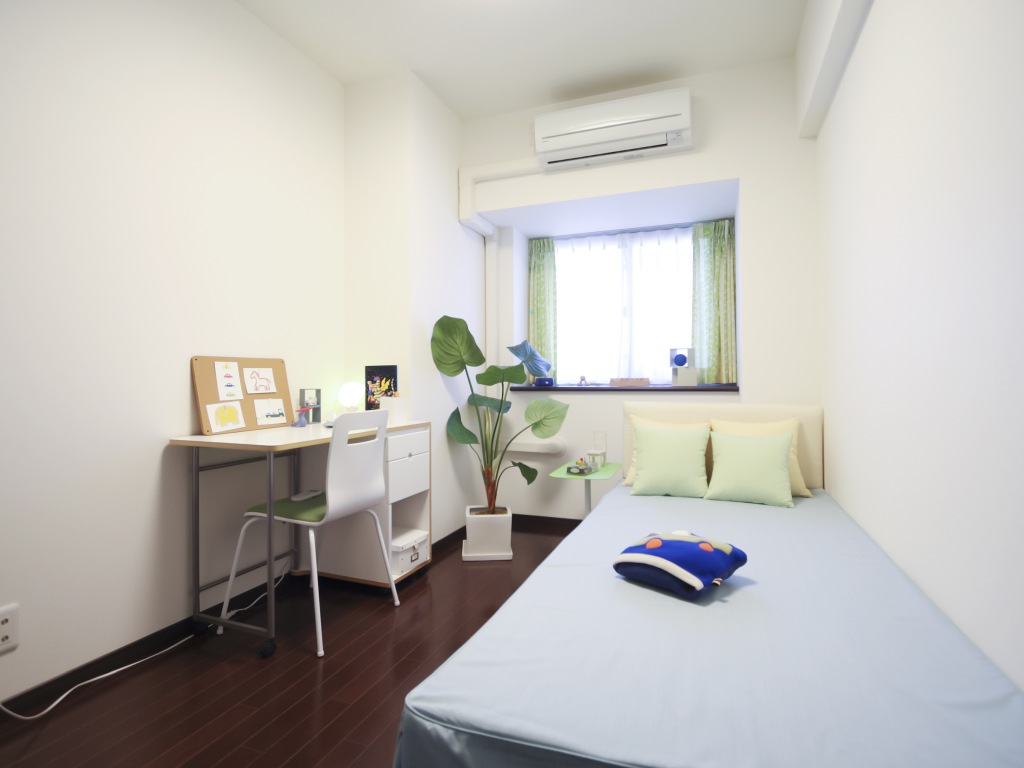} & 
\includegraphics[width=.244\linewidth]{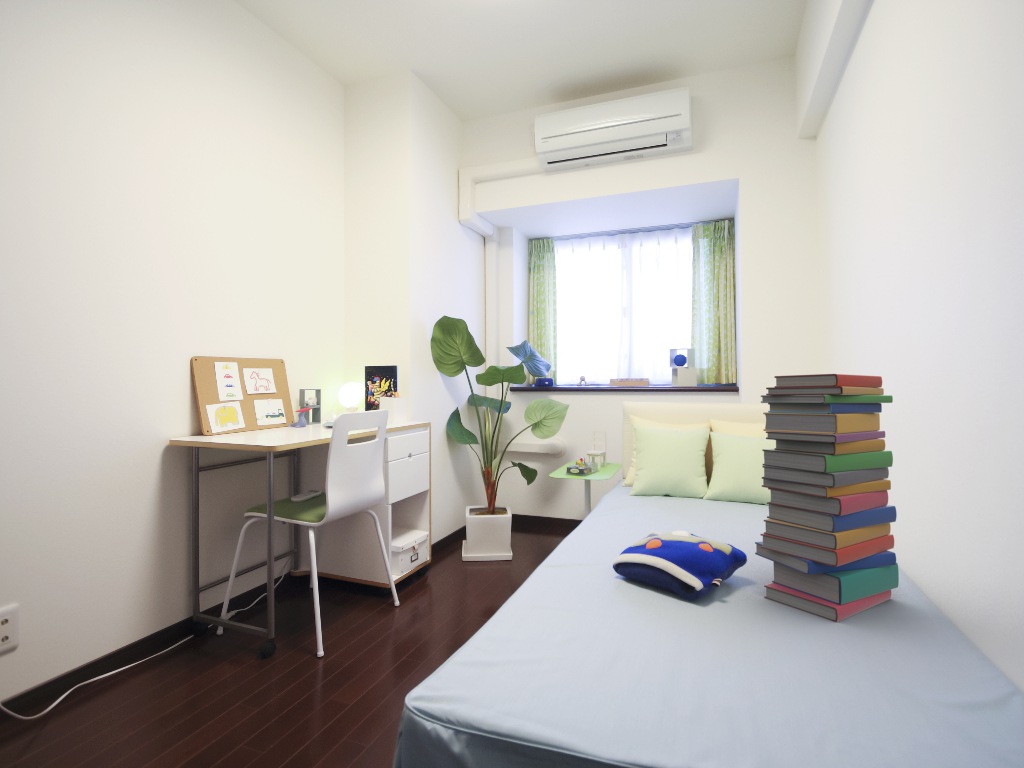}\vspace{.4em}\\
(a) Input photo & 
(b) Relit by our estimate & 
\hspace{.5em}
(c) Input photo & 
(d) Relit by our estimate \\
\end{tabular}
\caption{Object relighting on a variety of generic stock photos downloaded from the Internet. In all cases, light estimation is performed completely automatically by our HDR network, the output of which is directly used by the rendering engine to relight the virtual objects.}
\label{f:results-renders-stock}
\end{figure*}

Finally, we also provide qualitative examples of objects inserted in generic stock photos downloaded from the Internet in fig.~\ref{f:results-renders-stock}. The network is able to obtain robust illumination estimates, even for images with varying fields-of-view and viewpoints, uncontrolled capture settings, and unknown post-processing---all factors that lie outside our HDR training set.

\begin{figure}
\centering
\footnotesize
\setlength{\tabcolsep}{1pt}
\begin{tabular}{cc}
\includegraphics[width=.49\linewidth]{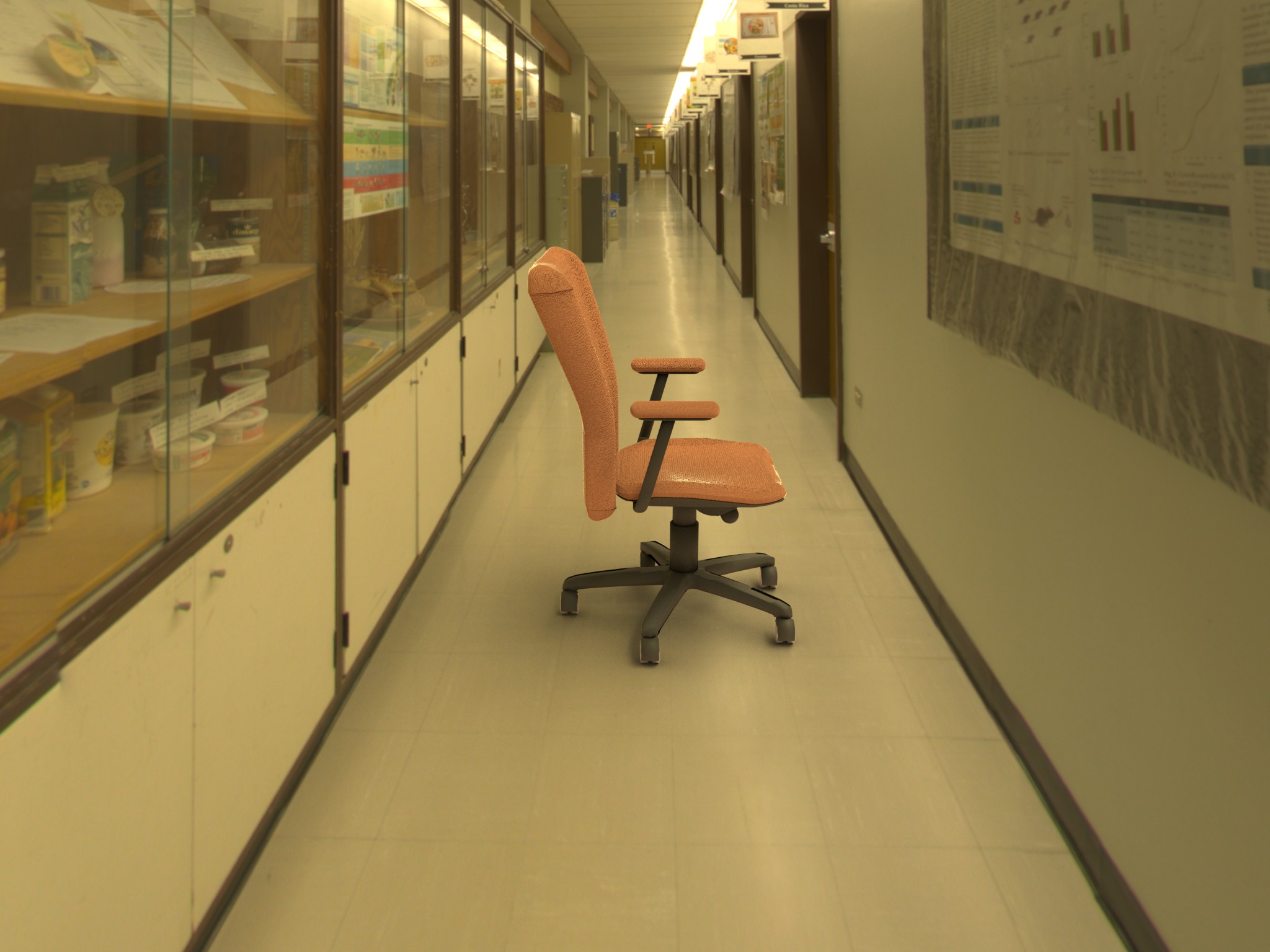} & 
\includegraphics[width=.49\linewidth]{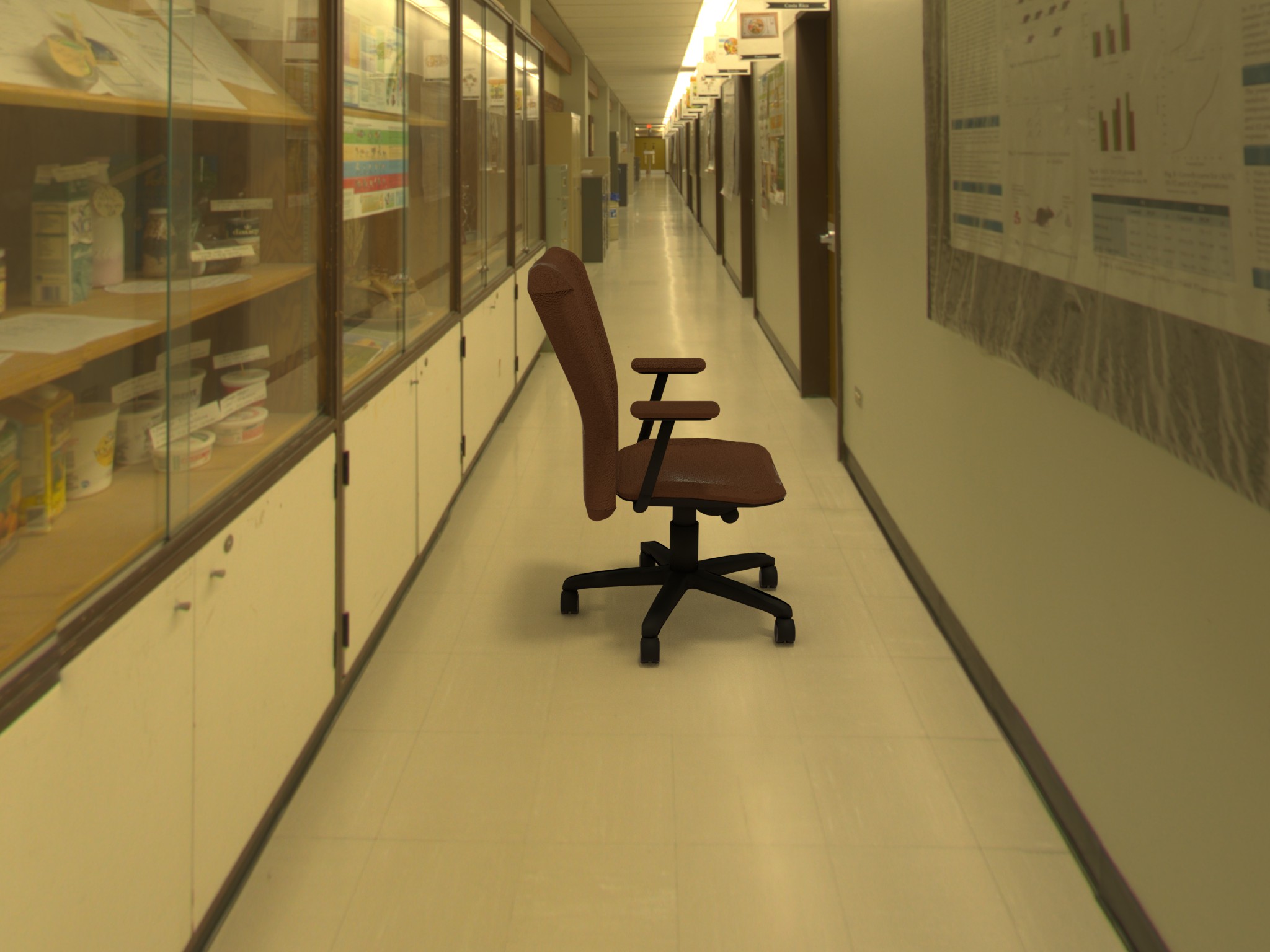} \\
\includegraphics[width=.49\linewidth]{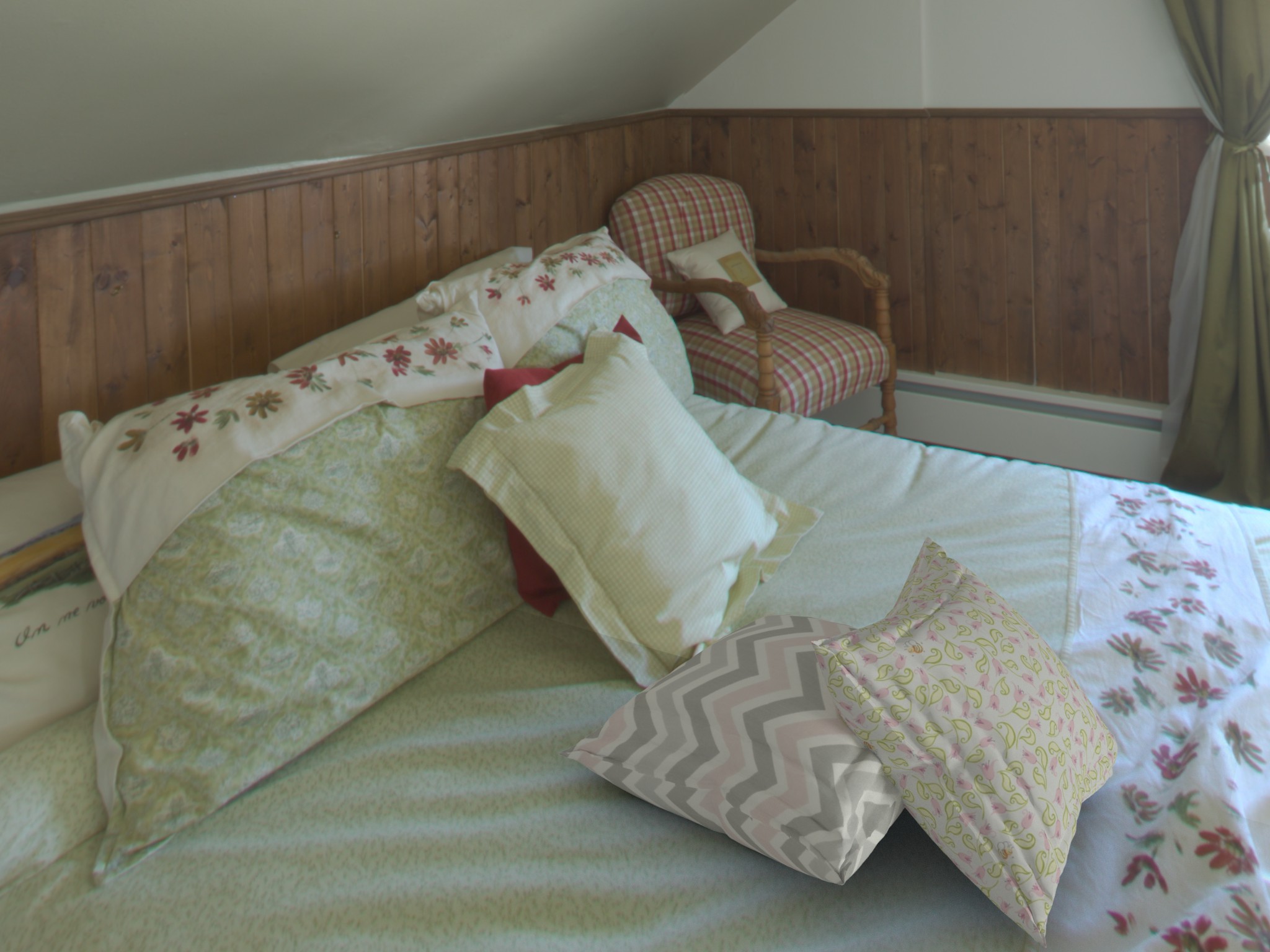} & 
\includegraphics[width=.49\linewidth]{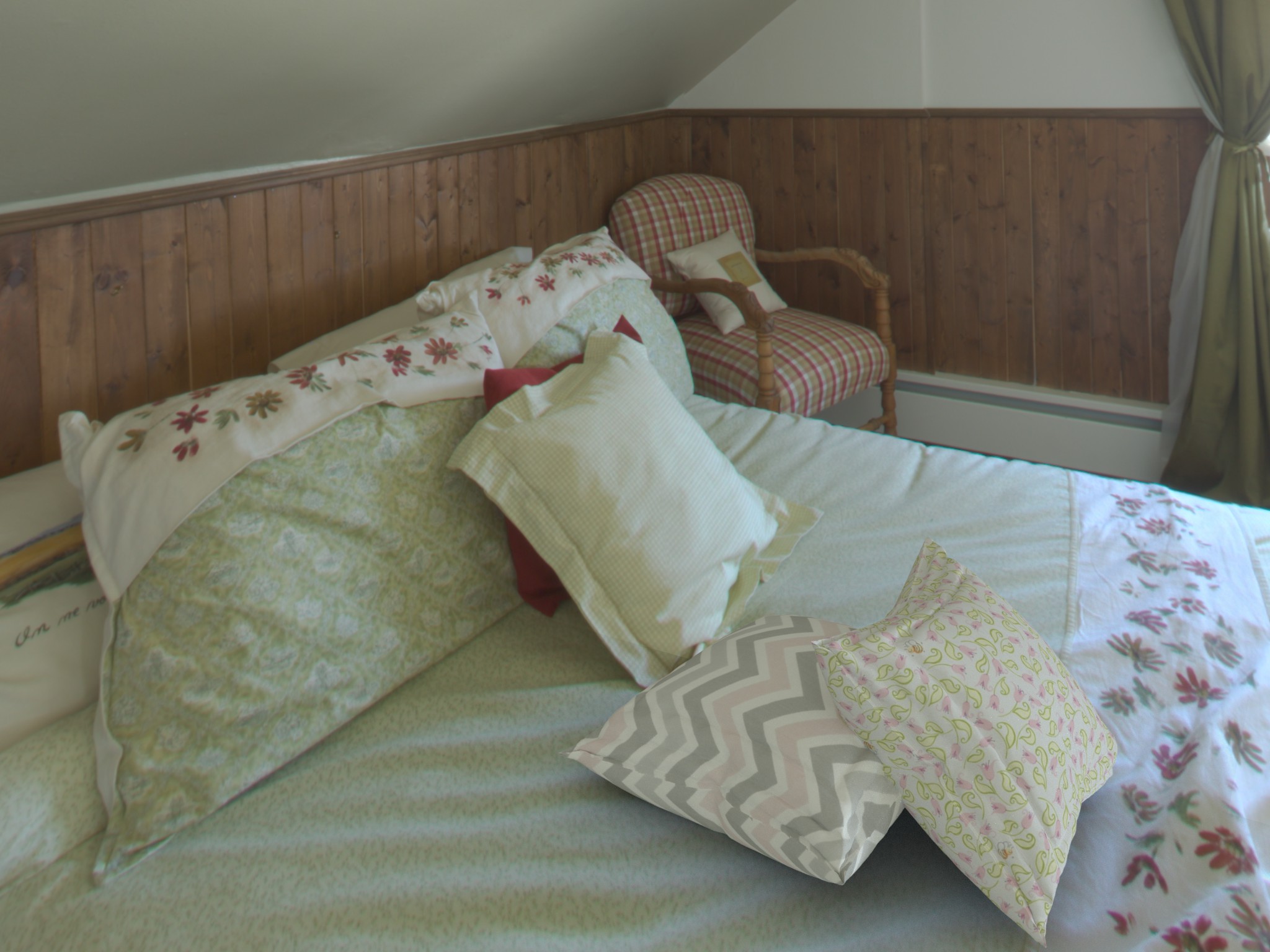} \\
\end{tabular}
\caption[]{For each row, the virtual objects in the image are either lit by the ground truth or by the output of our HDR network. Can you guess which is which? Answers below. \\
\rotatebox{180}{First row: left is GT, second row: right is GT. Did you get it right?}}
\label{f:results-userstudy}
\end{figure}

\begin{figure}
\centering
\includegraphics[width=0.9\columnwidth]{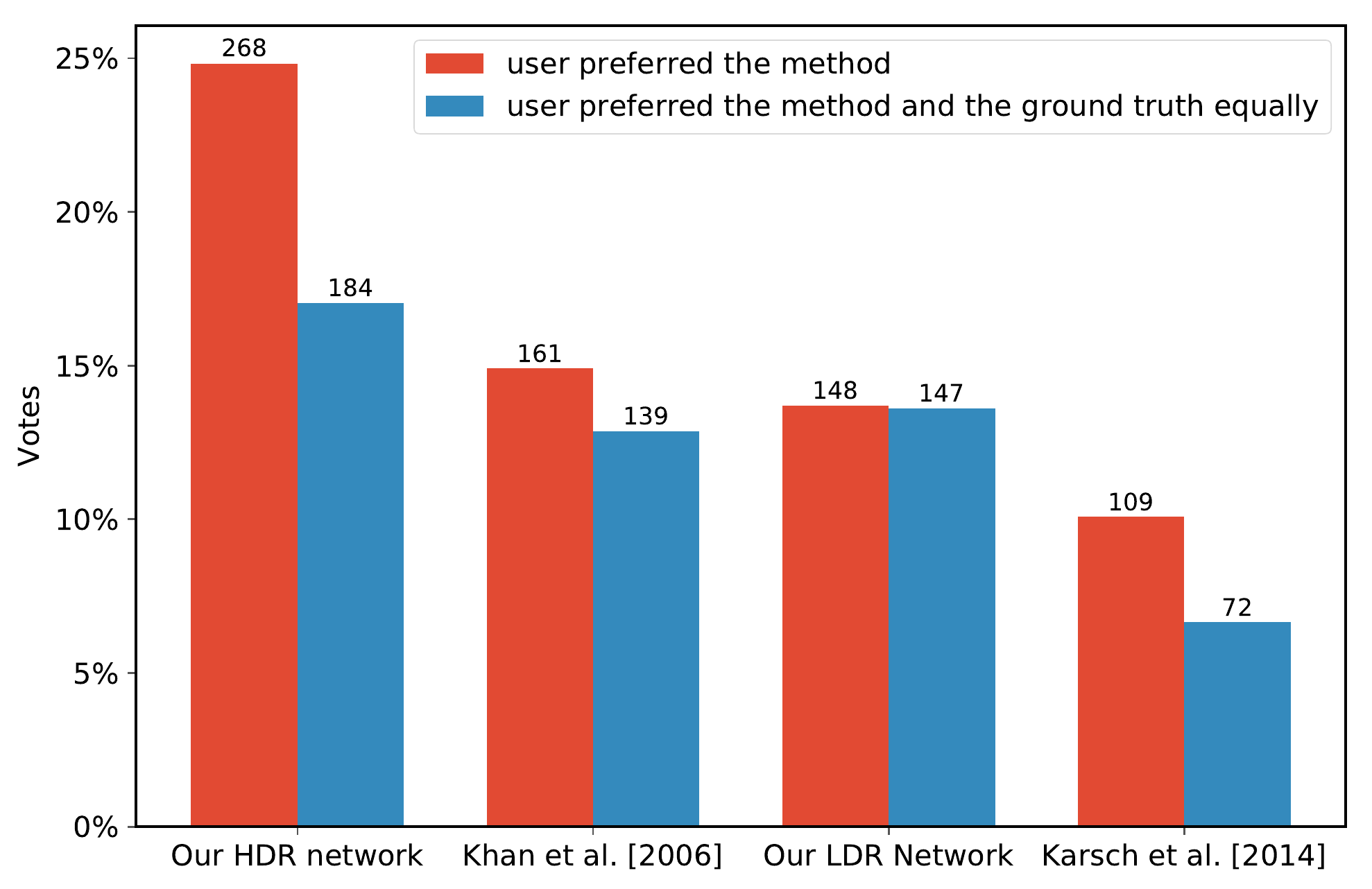}
\caption{Each method that participated in the user study is shown as a column, where blue votes indicate that user preferred the method instead of the ground truth, whereas red votes indicate that the user did not have a strong preference w.r.t the ground truth and the method.}
\label{f:results-us-bars}
\end{figure}

\subsection{Comparison with previous work}

We compare our approach with that of Khan et al.~\shortcite{khan-siggraph-06} and Karsch et al.~\shortcite{karsch-tog-14}, and show comparative relighting results in fig.~\ref{f:results-comparison}. Khan et al.~\shortcite{khan-siggraph-06} estimate the illumination conditions by projecting the background image directly on a sphere and flip it to get the whole environment map. This fails to estimate the proper dynamic range and position of light sources. In contrast, our approach produces robust estimates of lighting direction and intensity even when the light is not visible in the image. 

Karsch et al.~\shortcite{karsch-tog-14} use a light classifier to detect in-view lights, estimate out-of-view light locations by matching the background image to a database of panoramas, and estimate light intensities using a rendering-based optimization. We used the authors' original code to estimate both in-view and out-of-view lighting and these are shown in fig.~\ref{f:results-comparison}-(e). Their panorama matching is based on image appearance features that are not necessarily correlated with scene illumination. As a results, while their technique sometimes retrieves good matches, it may retrieve arbitrarily bad matches. In this case, adjusting the light intensities may not converge to satisfying answers since the light sources are not allowed to move in their optimization. In addition, their inverse lighting approach relies on reconstructing the depth and the diffuse albedo of the scene. Both of these are challenging problems, and errors in these estimates lead to errors in their lighting predictions. In contrast, our method learns a direct mapping between image appearance and scene illumination and yields robust, accurate results. 

\subsection{User study}

In addition to inferring scene illumination, we are also interested in using these estimates for graphics applications like 3D object insertion. This begs the questions, \emph{how realistic do synthetic objects lit by our estimates look when they are composited into input images?} We assess this axis of performance via a perceptual user study. We prepared 20 scenes from our HDR test set, and inserted a variety of different virtual objects in each of them. We generated a reference composite by relighting objects into these images using their ground truth illumination (obtained by warping the HDR panorama from which the image was extracted). We compared these results with objects that were relit using light probes estimated from four different methods: 1) our HDR network (\emph{without} artist tuning, see sec.~\ref{ss:hdreval}), 2) our LDR network, 3) \cite{khan-siggraph-06}, and 4) \cite{karsch-tog-14}. For each technique, we showed users a pair of images --- the reference image rendered with the ground truth illumination and the result rendered with one of the methods to be compared -- and asked them to indicate which image looks the most realistic. The realism of these composites is significantly affected by the geometric alignment of the objects and scenes, quality of the object geometries and materials, and rendering settings; our forced choice A/B test allows us to somewhat remove these factors and isolate the effect of the lighting estimation on the realism of the composite. Note that this test is only possible because we have high-quality ground truth illumination from our HDR dataset.

Two examples of the type of comparison asked of users is shown in fig.~\ref{f:results-userstudy}. The users were given the option to choose the more realistic image from the given two choices, and and additional third option to indicate that they are both equally plausible. Aggregated user study results are given in fig.~\ref{f:results-us-bars}. In total, we gathered responses from $105$ unique participants, with $1080$ comparisons with respect to the ground truth for each method. The results indicate that our HDR network rendering was considered as or more realistic than the ground truth result in $41.85\%$ of the responses, which is a significant improvement over Khan et al.~\shortcite{khan-siggraph-06} ($27.78\%$) and Karsch et al.~\shortcite{karsch-tog-14} ($16.76\%$). Note that even our LDR network ($27.32\%$) is comparable to Khan et al. and a significant improvement over Karsch et al.~\shortcite{karsch-tog-14}.

Beyond the significant improvement our automatic methods provides over these previous methods, note that our results can be further improved by simply tuning the global intensity scaling (see fig.~\ref{f:results-comparison}(c)). This is not possible for either Khan et al. (which cannot recover distinct lights to create effects like shadows) and Karsch et al. (which often gets the direction of the lights wrong and requires significant effort to correct).



\begin{figure}[!t]
\centering
\footnotesize
\setlength{\tabcolsep}{1pt}
\begin{tabular}{ccc}
\includegraphics[height=1.55cm]{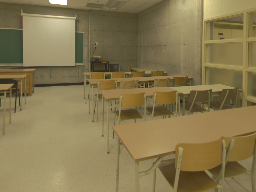} & 
\includegraphics[height=1.55cm]{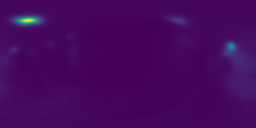} & 
\includegraphics[height=1.55cm]{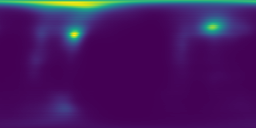} \\
\includegraphics[height=1.55cm]{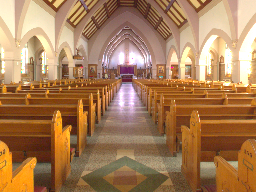} & 
\includegraphics[height=1.55cm]{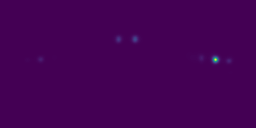} & 
\includegraphics[height=1.55cm]{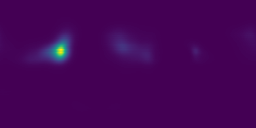} \\
\includegraphics[height=1.55cm]{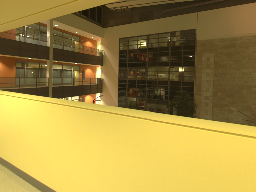} & 
\includegraphics[height=1.55cm]{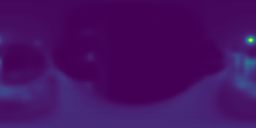} & 
\includegraphics[height=1.55cm]{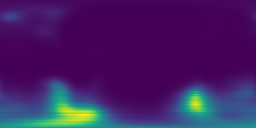} \\

(a) Input photo & 
(b) Ground truth &
(c) Our prediction 
\end{tabular}
\caption[]{Typical failure cases include incorrect spatial extent and orientation of light sources (often blurring them, top), scenes with complex geometry (middle), and images with strong local illumination variations (the lower half of the bottom image is lit, while the upper half is not).}
\label{f:failure-cases}
\vspace{1em}
\end{figure}

\section{Discussion}

Our main contribution in this paper is to frame illumination estimation from indoor scenes as an end-to-end learning problem. The major benefit of our approach is that it learns a direct mapping from image to lighting. Therefore, it does not make any assumption about the scene (other than it is captured indoors), it does not need to explicitly estimate other scene properties such as scene or camera geometry, nor does it need to use inverse rendering methods which are slow to optimize and prone to local minima. To make this learning possible, we introduce two additional contributions: 1) a method for detecting light sources in LDR panoramas which outperforms the state of the art; and 2) a panorama warping operation that allows us to adapt the lighting conditions for different cropped photos extracted from the panoramas. Together these steps allow us to automatically create labeled data from the SUN360 dataset~\cite{xiao-cvpr-12}, and train an LDR illumination prediction network. We fine-tuned this network using a new dataset of HDR environment maps to predict HDR scene illumination. Both quantitative and perceptual evaluations of our method show that it is significantly better than the state-of-the-art. 

\paragraph{Limitations and future work} 

While our network is good predicting at light locations, it sometimes has issues inferring the spatial extent and orientation of light sources, particularly for out-of-view lights. This might be partly caused by our in-network filtering. As a result, large area lights might be detected as smaller lights. More typically, sharp light sources get blurred out and do not create the kind of crisp cast shadows that are characteristic of them. Finally, the network can fail on images with complex, ambiguous geometric or photometric cues. Fig.~\ref{f:failure-cases} illustrates some of these scenarios. Generally speaking, the network is often better at recovering the light source locations than intensity, since the pre-training step (making the network retrieving light positions) has a much larger training set than the fine-tuning step. Also relevant is the exposure of the input LDR image, to which the light intensity estimator is much more sensitive than the light location estimator.

Our network was trained to predict light intensities and we used a simple scheme to assign color. Needless to say, light color plays an important role in image appearance, and we would like to robustly infer it too. We trained our network to predict one lighting solution for an input image. However, as noted previously, indoor illumination is localized in nature, and might vary even within an image (fig.~\ref{f:failure-cases}, bottom). Our long-term goal is to be able to recover this spatially-varying lighting distribution---a challenging problem that will certainly require datasets annotated with geometry and illumination.

Finally, while our work looks at only problem of lighting estimation, this problem is closely related to other scene inference tasks like geometric reconstruction and intrinsic images. Inferring all these properties jointly could benefit each individual task.

\section*{Acknowledgements}

The authors would like to thank Yannick Hold-Geoffroy for his help in setting up the renders and the user study. We would also like to thank Henrique Weber for his help with EnvyDepth, and Jean-Michel Fortin for his work on HDR data capture.

Parts of this work were done while Marc-Andr\'e Gardner was an intern at Adobe Research. This work was partially supported by the REPARTI Strategic Network and the FRQNT New Researcher Grant 2016NC189939. We gratefully acknowledge the support of Nvidia with the donation of the GPUs used for this research, funding from Adobe to cover the cost of HDR dataset acquisition, as well as a generous gift from Adobe to J.-F. Lalonde. 

\bibliographystyle{ACM-Reference-Format}
\bibliography{template}


\begin{thebibliography}{34}


\ifx \showCODEN    \undefined \def \showCODEN     #1{\unskip}     \fi
\ifx \showDOI      \undefined \def \showDOI       #1{#1}\fi
\ifx \showISBNx    \undefined \def \showISBNx     #1{\unskip}     \fi
\ifx \showISBNxiii \undefined \def \showISBNxiii  #1{\unskip}     \fi
\ifx \showISSN     \undefined \def \showISSN      #1{\unskip}     \fi
\ifx \showLCCN     \undefined \def \showLCCN      #1{\unskip}     \fi
\ifx \shownote     \undefined \def \shownote      #1{#1}          \fi
\ifx \showarticletitle \undefined \def \showarticletitle #1{#1}   \fi
\ifx \showURL      \undefined \def \showURL       {\relax}        \fi
\providecommand\bibfield[2]{#2}
\providecommand\bibinfo[2]{#2}
\providecommand\natexlab[1]{#1}
\providecommand\showeprint[2][]{arXiv:#2}

\bibitem[\protect\citeauthoryear{Bansal, Russell, and Gupta}{Bansal
  et~al\mbox{.}}{2016}]%
        {bansal2016marr}
\bibfield{author}{\bibinfo{person}{Aayush Bansal}, \bibinfo{person}{Bryan
  Russell}, {and} \bibinfo{person}{Abhinav Gupta}.}
  \bibinfo{year}{2016}\natexlab{}.
\newblock \showarticletitle{Marr Revisited: {2D}-{3D} Model Alignment via
  Surface Normal Prediction}.
\newblock \bibinfo{journal}{{\em IEEE Conference on Computer Vision and Pattern
  Recognition\/}} (\bibinfo{year}{2016}).
\newblock


\bibitem[\protect\citeauthoryear{Banterle, Callieri, Dellepiane, Corsini,
  Pellacini, and Scopigno}{Banterle et~al\mbox{.}}{2013}]%
        {banterle-cgf-13}
\bibfield{author}{\bibinfo{person}{Francesco Banterle}, \bibinfo{person}{Marco
  Callieri}, \bibinfo{person}{Matteo Dellepiane}, \bibinfo{person}{Massimiliano
  Corsini}, \bibinfo{person}{Fabio Pellacini}, {and} \bibinfo{person}{Roberto
  Scopigno}.} \bibinfo{year}{2013}\natexlab{}.
\newblock \showarticletitle{Envy{D}epth: An interface for recovering local
  natural illumination from environment maps}.
\newblock \bibinfo{journal}{{\em Computer Graphics Forum\/}}
  \bibinfo{volume}{32}, \bibinfo{number}{7} (\bibinfo{year}{2013}),
  \bibinfo{pages}{411--420}.
\newblock


\bibitem[\protect\citeauthoryear{Barron and Malik}{Barron and Malik}{2013a}]%
        {barron2013rgbd}
\bibfield{author}{\bibinfo{person}{Jonathan Barron} {and}
  \bibinfo{person}{Jitendra Malik}.} \bibinfo{year}{2013}\natexlab{a}.
\newblock \showarticletitle{Intrinsic Scene Properties from a Single RGB-D
  Image}.
\newblock \bibinfo{journal}{{\em IEEE Conference on Computer Vision and Pattern
  Recognition\/}} (\bibinfo{year}{2013}).
\newblock


\bibitem[\protect\citeauthoryear{Barron and Malik}{Barron and Malik}{2013b}]%
        {barron-pami-15}
\bibfield{author}{\bibinfo{person}{Jonathan Barron} {and}
  \bibinfo{person}{Jitendra Malik}.} \bibinfo{year}{2013}\natexlab{b}.
\newblock \showarticletitle{Shape, illumination, and reflectance from shading}.
\newblock \bibinfo{journal}{{\em IEEE Transactions on Pattern Analysis and
  Machine Intelligence\/}} \bibinfo{volume}{37}, \bibinfo{number}{8}
  (\bibinfo{year}{2013}), \bibinfo{pages}{1670--1687}.
\newblock


\bibitem[\protect\citeauthoryear{Bell, Upchurch, Snavely, and Bala}{Bell
  et~al\mbox{.}}{2015}]%
        {bell2015minc}
\bibfield{author}{\bibinfo{person}{Sean Bell}, \bibinfo{person}{Paul Upchurch},
  \bibinfo{person}{Noah Snavely}, {and} \bibinfo{person}{Kavita Bala}.}
  \bibinfo{year}{2015}\natexlab{}.
\newblock \showarticletitle{Material Recognition in the Wild with the Materials
  in Context Database}.
\newblock \bibinfo{journal}{{\em IEEE Conference on Computer Vision and Pattern
  Recognition\/}} (\bibinfo{year}{2015}).
\newblock


\bibitem[\protect\citeauthoryear{Bitouk, Kumar, Dhillon, Belhumeur, and
  Nayar}{Bitouk et~al\mbox{.}}{2008}]%
        {bitouk-sig-08}
\bibfield{author}{\bibinfo{person}{Dmitri Bitouk}, \bibinfo{person}{Neeraj
  Kumar}, \bibinfo{person}{Samreen Dhillon}, \bibinfo{person}{Peter Belhumeur},
  {and} \bibinfo{person}{Shree~K. Nayar}.} \bibinfo{year}{2008}\natexlab{}.
\newblock \showarticletitle{Face Swapping: Automatically Replacing Faces in
  Photographs}.
\newblock \bibinfo{journal}{{\em ACM Transactions on Graphics\/}}
  \bibinfo{volume}{27}, \bibinfo{number}{3}, Article \bibinfo{articleno}{39}
  (\bibinfo{date}{Aug.} \bibinfo{year}{2008}), \bibinfo{numpages}{8}~pages.
\newblock


\bibitem[\protect\citeauthoryear{Clevert, Unterthiner, and Hochreiter}{Clevert
  et~al\mbox{.}}{2016}]%
        {clevert-iclr-16}
\bibfield{author}{\bibinfo{person}{Djork-Arn{\'{e}} Clevert},
  \bibinfo{person}{Thomas Unterthiner}, {and} \bibinfo{person}{Sepp
  Hochreiter}.} \bibinfo{year}{2016}\natexlab{}.
\newblock \showarticletitle{Fast and Accurate Deep Network Learning by
  Exponential Linear Units ({ELUs})}. In \bibinfo{booktitle}{{\em International
  Conference on Learning Representations}}.
\newblock


\bibitem[\protect\citeauthoryear{Dalal and Triggs}{Dalal and Triggs}{2005}]%
        {dalal-cvpr-05}
\bibfield{author}{\bibinfo{person}{Navneet Dalal} {and} \bibinfo{person}{Bill
  Triggs}.} \bibinfo{year}{2005}\natexlab{}.
\newblock \showarticletitle{Histograms of Oriented Gradients for Human
  Detection}. In \bibinfo{booktitle}{{\em IEEE Conference on Computer Vision
  and Pattern Recognition}}. \bibinfo{pages}{886--893}.
\newblock


\bibitem[\protect\citeauthoryear{Debevec}{Debevec}{1997}]%
        {debevec-siggraph-97}
\bibfield{author}{\bibinfo{person}{Paul Debevec}.}
  \bibinfo{year}{1997}\natexlab{}.
\newblock \showarticletitle{Recovering High Dynamic Range Radiance Maps from
  Photographs}. In \bibinfo{booktitle}{{\em ACM SIGGRAPH 1997}}.
  \bibinfo{pages}{1--10}.
\newblock


\bibitem[\protect\citeauthoryear{Debevec}{Debevec}{1998}]%
        {debevec-sig-98}
\bibfield{author}{\bibinfo{person}{Paul Debevec}.}
  \bibinfo{year}{1998}\natexlab{}.
\newblock \showarticletitle{Rendering Synthetic Objects into Real Scenes :
  Bridging Traditional and Image-based Graphics with Global Illumination and
  High Dynamic Range Photography}. In \bibinfo{booktitle}{{\em Proceedings of
  ACM SIGGRAPH}}.
\newblock


\bibitem[\protect\citeauthoryear{Debevec, Graham, Busch, and Bolas}{Debevec
  et~al\mbox{.}}{2012}]%
        {debevec-sslp-12}
\bibfield{author}{\bibinfo{person}{Paul Debevec}, \bibinfo{person}{Paul
  Graham}, \bibinfo{person}{Jay Busch}, {and} \bibinfo{person}{Mark Bolas}.}
  \bibinfo{year}{2012}\natexlab{}.
\newblock \showarticletitle{A Single-shot Light Probe}. In
  \bibinfo{booktitle}{{\em ACM SIGGRAPH 2012 Talks}}. \bibinfo{publisher}{ACM},
  \bibinfo{address}{New York, NY, USA}, \bibinfo{pages}{10:1--10:1}.
\newblock


\bibitem[\protect\citeauthoryear{Eigen and Fergus}{Eigen and Fergus}{2015}]%
        {eigen-iccv-15}
\bibfield{author}{\bibinfo{person}{David Eigen} {and} \bibinfo{person}{Rob
  Fergus}.} \bibinfo{year}{2015}\natexlab{}.
\newblock \showarticletitle{Predicting Depth, Surface Normals and Semantic
  Labels with a Common Multi-Scale Convolutional Architecture}.
\newblock \bibinfo{journal}{{\em International Conference on Computer
  Vision\/}} (\bibinfo{year}{2015}).
\newblock


\bibitem[\protect\citeauthoryear{Felzenszwalb, Girshick, McAllester, and
  Ramanan}{Felzenszwalb et~al\mbox{.}}{2010}]%
        {felzenszwalb-pami-10}
\bibfield{author}{\bibinfo{person}{Pedro~F. Felzenszwalb},
  \bibinfo{person}{Ross~B. Girshick}, \bibinfo{person}{David McAllester}, {and}
  \bibinfo{person}{Deva Ramanan}.} \bibinfo{year}{2010}\natexlab{}.
\newblock \showarticletitle{Object Detection with Discriminative Trained Part
  Based Models}.
\newblock \bibinfo{journal}{{\em IEEE Transactions on Pattern Analysis and
  Machine Intelligence\/}} \bibinfo{volume}{32}, \bibinfo{number}{9}
  (\bibinfo{year}{2010}), \bibinfo{pages}{1627--1645}.
\newblock


\bibitem[\protect\citeauthoryear{Georgoulis, Rematas, Ritschel, Fritz, Gool,
  and Tuytelaars}{Georgoulis et~al\mbox{.}}{2016}]%
        {georgoulis-arxiv-16a}
\bibfield{author}{\bibinfo{person}{Stamatios Georgoulis},
  \bibinfo{person}{Konstantinos Rematas}, \bibinfo{person}{Tobias Ritschel},
  \bibinfo{person}{Mario Fritz}, \bibinfo{person}{Luc J.~Van Gool}, {and}
  \bibinfo{person}{Tinne Tuytelaars}.} \bibinfo{year}{2016}\natexlab{}.
\newblock \showarticletitle{DeLight-Net: Decomposing Reflectance Maps into
  Specular Materials and Natural Illumination}.
\newblock \bibinfo{journal}{{\em CoRR\/}}  \bibinfo{volume}{abs/1603.08240}
  (\bibinfo{year}{2016}).
\newblock


\bibitem[\protect\citeauthoryear{He, Zhang, Ren, and Sun}{He
  et~al\mbox{.}}{2016}]%
        {he-cvpr-16}
\bibfield{author}{\bibinfo{person}{Kaiming He}, \bibinfo{person}{Xiangyu
  Zhang}, \bibinfo{person}{Shaoqing Ren}, {and} \bibinfo{person}{Jian Sun}.}
  \bibinfo{year}{2016}\natexlab{}.
\newblock \showarticletitle{Deep Residual Learning for Image Recognition}. In
  \bibinfo{booktitle}{{\em IEEE Conference on Computer Vision and Pattern
  Recognition}}.
\newblock


\bibitem[\protect\citeauthoryear{Hold{-}Geoffroy, Sunkavalli, Hadap,
  Gambaretto, and Lalonde}{Hold{-}Geoffroy et~al\mbox{.}}{2017}]%
        {holdgeoffroy-cvpr-17}
\bibfield{author}{\bibinfo{person}{Yannick Hold{-}Geoffroy},
  \bibinfo{person}{Kalyan Sunkavalli}, \bibinfo{person}{Sunil Hadap},
  \bibinfo{person}{Emiliano Gambaretto}, {and}
  \bibinfo{person}{Jean{-}Fran{\c{c}}ois Lalonde}.}
  \bibinfo{year}{2017}\natexlab{}.
\newblock \showarticletitle{Deep Outdoor Illumination Estimation}. In
  \bibinfo{booktitle}{{\em IEEE Conference on Computer Vision and Pattern
  Recognition}}.
\newblock


\bibitem[\protect\citeauthoryear{Ho{\v{s}}ek and Wilkie}{Ho{\v{s}}ek and
  Wilkie}{2012}]%
        {Hosek2012}
\bibfield{author}{\bibinfo{person}{Luk{\'{a}}{\v{s}} Ho{\v{s}}ek} {and}
  \bibinfo{person}{Alexander Wilkie}.} \bibinfo{year}{2012}\natexlab{}.
\newblock \showarticletitle{{An analytic model for full spectral sky-dome
  radiance}}.
\newblock \bibinfo{journal}{{\em ACM Transactions on Graphics\/}}
  \bibinfo{volume}{31}, \bibinfo{number}{4} (\bibinfo{year}{2012}),
  \bibinfo{pages}{1--9}.
\newblock
\showISBNx{0730-0301}
\showISSN{07300301}
\showURL{%
\url{http://www.scopus.com/inward/record.url?eid=2-s2.0-84872254894}}


\bibitem[\protect\citeauthoryear{Karsch, Hedau, Forsyth, and Hoiem}{Karsch
  et~al\mbox{.}}{2011}]%
        {karsch-sig-11}
\bibfield{author}{\bibinfo{person}{Kevin Karsch}, \bibinfo{person}{Varsha
  Hedau}, \bibinfo{person}{David Forsyth}, {and} \bibinfo{person}{Derek
  Hoiem}.} \bibinfo{year}{2011}\natexlab{}.
\newblock \showarticletitle{Rendering synthetic objects into legacy
  photographs}.
\newblock \bibinfo{journal}{{\em ACM Transactions on Graphics\/}}
  \bibinfo{volume}{30}, \bibinfo{number}{6} (\bibinfo{year}{2011}),
  \bibinfo{pages}{1}.
\newblock


\bibitem[\protect\citeauthoryear{Karsch, Sunkavalli, Hadap, Carr, Jin, Fonte,
  Sittig, and Forsyth}{Karsch et~al\mbox{.}}{2014}]%
        {karsch-tog-14}
\bibfield{author}{\bibinfo{person}{Kevin Karsch}, \bibinfo{person}{Kalyan
  Sunkavalli}, \bibinfo{person}{Sunil Hadap}, \bibinfo{person}{Nathan Carr},
  \bibinfo{person}{Hailin Jin}, \bibinfo{person}{Rafael Fonte},
  \bibinfo{person}{Michael Sittig}, {and} \bibinfo{person}{David Forsyth}.}
  \bibinfo{year}{2014}\natexlab{}.
\newblock \showarticletitle{Automatic Scene Inference for 3D Object
  Compositing}.
\newblock \bibinfo{journal}{{\em ACM Transactions on Graphics\/}}
  \bibinfo{number}{3} (\bibinfo{year}{2014}), \bibinfo{pages}{32:1--32:15}.
\newblock


\bibitem[\protect\citeauthoryear{Khan, Reinhard, Fleming, and
  B{\"{u}}lthoff}{Khan et~al\mbox{.}}{2006}]%
        {khan-siggraph-06}
\bibfield{author}{\bibinfo{person}{Erum~Arif Khan}, \bibinfo{person}{Erik
  Reinhard}, \bibinfo{person}{Roland~W. Fleming}, {and}
  \bibinfo{person}{Heinrich~H. B{\"{u}}lthoff}.}
  \bibinfo{year}{2006}\natexlab{}.
\newblock \showarticletitle{Image-based material editing}.
\newblock \bibinfo{journal}{{\em ACM Transactions on Graphics\/}}
  \bibinfo{volume}{25}, \bibinfo{number}{3} (\bibinfo{year}{2006}),
  \bibinfo{pages}{654}.
\newblock


\bibitem[\protect\citeauthoryear{Kingma and Ba}{Kingma and Ba}{2014}]%
        {kingma2014adam}
\bibfield{author}{\bibinfo{person}{Diederik Kingma} {and}
  \bibinfo{person}{Jimmy Ba}.} \bibinfo{year}{2014}\natexlab{}.
\newblock \showarticletitle{Adam: A method for stochastic optimization}.
\newblock \bibinfo{journal}{{\em arXiv preprint arXiv:1412.6980\/}}
  (\bibinfo{year}{2014}).
\newblock


\bibitem[\protect\citeauthoryear{Kr\"{a}henb\"{u}hl and
  Koltun}{Kr\"{a}henb\"{u}hl and Koltun}{2012}]%
        {krahenbuhl-nips-12}
\bibfield{author}{\bibinfo{person}{Philipp Kr\"{a}henb\"{u}hl} {and}
  \bibinfo{person}{Vladlen Koltun}.} \bibinfo{year}{2012}\natexlab{}.
\newblock \showarticletitle{Efficient Inference in Fully Connected {CRF}s with
  Gaussian Edge Potentials}. In \bibinfo{booktitle}{{\em Neural Information
  Processing Systems}}.
\newblock


\bibitem[\protect\citeauthoryear{Lalonde, Hoiem, Efros, Rother, Winn, and
  Criminisi}{Lalonde et~al\mbox{.}}{2007}]%
        {lalonde-sig-07}
\bibfield{author}{\bibinfo{person}{Jean-Fran\c{c}ois Lalonde},
  \bibinfo{person}{Derek Hoiem}, \bibinfo{person}{Alexei~A. Efros},
  \bibinfo{person}{Carsten Rother}, \bibinfo{person}{John Winn}, {and}
  \bibinfo{person}{Antonio Criminisi}.} \bibinfo{year}{2007}\natexlab{}.
\newblock \showarticletitle{Photo Clip Art}.
\newblock \bibinfo{journal}{{\em ACM Transactions on Graphics\/}}
  \bibinfo{volume}{26}, \bibinfo{number}{3}, Article \bibinfo{articleno}{3}
  (\bibinfo{date}{July} \bibinfo{year}{2007}).
\newblock


\bibitem[\protect\citeauthoryear{Lalonde, Narasimhan, and Efros}{Lalonde
  et~al\mbox{.}}{2010}]%
        {lalonde-ijcv-10}
\bibfield{author}{\bibinfo{person}{Jean-Fran\c{c}ois Lalonde},
  \bibinfo{person}{Srinivasa~G. Narasimhan}, {and} \bibinfo{person}{Alexei~A.
  Efros}.} \bibinfo{year}{2010}\natexlab{}.
\newblock \showarticletitle{What do the sun and the sky tell us about the
  camera?}
\newblock \bibinfo{journal}{{\em International Journal on Computer Vision\/}}
  \bibinfo{volume}{88}, \bibinfo{number}{1} (\bibinfo{date}{May}
  \bibinfo{year}{2010}), \bibinfo{pages}{24--51}.
\newblock


\bibitem[\protect\citeauthoryear{Lombardi and Nishino}{Lombardi and
  Nishino}{2016}]%
        {lombardi2016reflectance}
\bibfield{author}{\bibinfo{person}{Stephen Lombardi} {and} \bibinfo{person}{Ko
  Nishino}.} \bibinfo{year}{2016}\natexlab{}.
\newblock \showarticletitle{Reflectance and Illumination Recovery in the Wild}.
\newblock \bibinfo{journal}{{\em IEEE Transactions on Pattern Analysis and
  Machine Intelligence\/}} \bibinfo{volume}{38}, \bibinfo{number}{1}
  (\bibinfo{year}{2016}), \bibinfo{pages}{129--141}.
\newblock


\bibitem[\protect\citeauthoryear{Lopez-Moreno, Hadap, Reinhard, and
  Gutierrez}{Lopez-Moreno et~al\mbox{.}}{2010}]%
        {moreno-cng-10}
\bibfield{author}{\bibinfo{person}{Jorge Lopez-Moreno}, \bibinfo{person}{Sunil
  Hadap}, \bibinfo{person}{Erik Reinhard}, {and} \bibinfo{person}{Diego
  Gutierrez}.} \bibinfo{year}{2010}\natexlab{}.
\newblock \showarticletitle{Compositing images through light source detection}.
\newblock \bibinfo{journal}{{\em Computers {\&} Graphics\/}}
  \bibinfo{volume}{34}, \bibinfo{number}{6} (\bibinfo{year}{2010}),
  \bibinfo{pages}{698--707}.
\newblock


\bibitem[\protect\citeauthoryear{Ramamoorthi and Hanrahan}{Ramamoorthi and
  Hanrahan}{2001}]%
        {ramamoorthi-sig-01}
\bibfield{author}{\bibinfo{person}{Ravi Ramamoorthi} {and} \bibinfo{person}{Pat
  Hanrahan}.} \bibinfo{year}{2001}\natexlab{}.
\newblock \showarticletitle{A Signal-processing Framework for Inverse
  Rendering}. In \bibinfo{booktitle}{{\em Proceedings of the 28th Annual
  Conference on Computer Graphics and Interactive Techniques}} {\em
  (\bibinfo{series}{SIGGRAPH '01})}. \bibinfo{pages}{117--128}.
\newblock


\bibitem[\protect\citeauthoryear{Reinhard, Heidrich, Debevec, Pattanaik, Ward,
  and Myszkowski}{Reinhard et~al\mbox{.}}{2010}]%
        {reinhard-book-10}
\bibfield{author}{\bibinfo{person}{Erik Reinhard}, \bibinfo{person}{Wolfgang
  Heidrich}, \bibinfo{person}{Paul Debevec}, \bibinfo{person}{Sumanta
  Pattanaik}, \bibinfo{person}{Greg Ward}, {and} \bibinfo{person}{Karol
  Myszkowski}.} \bibinfo{year}{2010}\natexlab{}.
\newblock \bibinfo{booktitle}{{\em High Dynamic Range Imaging\/}
  (\bibinfo{edition}{2} ed.)}.
\newblock \bibinfo{publisher}{Morgan Kaufman}.
\newblock


\bibitem[\protect\citeauthoryear{Rematas, Ritschel, Fritz, Gavves, and
  Tuytelaars}{Rematas et~al\mbox{.}}{2016}]%
        {rematas-cvpr-16}
\bibfield{author}{\bibinfo{person}{Konstantinos Rematas},
  \bibinfo{person}{Tobias Ritschel}, \bibinfo{person}{Mario Fritz},
  \bibinfo{person}{Efstratios Gavves}, {and} \bibinfo{person}{Tinne
  Tuytelaars}.} \bibinfo{year}{2016}\natexlab{}.
\newblock \showarticletitle{Deep Reflectance Maps}. In \bibinfo{booktitle}{{\em
  IEEE Conference on Computer Vision and Pattern Recognition}}.
\newblock


\bibitem[\protect\citeauthoryear{Valgaerts, Wu, Bruhn, Seidel, and
  Theobalt}{Valgaerts et~al\mbox{.}}{2012}]%
        {valgaerts-tog-12}
\bibfield{author}{\bibinfo{person}{Levi Valgaerts}, \bibinfo{person}{Chenglei
  Wu}, \bibinfo{person}{Andr{\'e}s Bruhn}, \bibinfo{person}{Hans-Peter Seidel},
  {and} \bibinfo{person}{Christian Theobalt}.} \bibinfo{year}{2012}\natexlab{}.
\newblock \showarticletitle{Lightweight Binocular Facial Performance Capture
  Under Uncontrolled Lighting}.
\newblock \bibinfo{journal}{{\em ACM Transactions on Graphics\/}}
  \bibinfo{volume}{31}, \bibinfo{number}{6}, Article \bibinfo{articleno}{187}
  (\bibinfo{date}{Nov.} \bibinfo{year}{2012}), \bibinfo{numpages}{11}~pages.
\newblock


\bibitem[\protect\citeauthoryear{Wu, Wilburn, Matsushita, and Theobalt}{Wu
  et~al\mbox{.}}{2011}]%
        {wu-cvpr-11}
\bibfield{author}{\bibinfo{person}{Chenglei Wu}, \bibinfo{person}{B. Wilburn},
  \bibinfo{person}{Y. Matsushita}, {and} \bibinfo{person}{C. Theobalt}.}
  \bibinfo{year}{2011}\natexlab{}.
\newblock \showarticletitle{High-quality Shape from Multi-view Stereo and
  Shading Under General Illumination}. In \bibinfo{booktitle}{{\em IEEE
  Conference on Computer Vision and Pattern Recognition}}.
\newblock


\bibitem[\protect\citeauthoryear{Xiao, Ehinger, Oliva, and Torralba}{Xiao
  et~al\mbox{.}}{2012}]%
        {xiao-cvpr-12}
\bibfield{author}{\bibinfo{person}{Jianxiong Xiao}, \bibinfo{person}{Krista~A.
  Ehinger}, \bibinfo{person}{Aude Oliva}, {and} \bibinfo{person}{Antonio
  Torralba}.} \bibinfo{year}{2012}\natexlab{}.
\newblock \showarticletitle{Recognizing scene viewpoint using panoramic place
  representation}. In \bibinfo{booktitle}{{\em IEEE Conference on Computer
  Vision and Pattern Recognition}}.
\newblock


\bibitem[\protect\citeauthoryear{Zhang, Cohen, and Curless}{Zhang
  et~al\mbox{.}}{2016}]%
        {zhang-siga-16}
\bibfield{author}{\bibinfo{person}{Edward Zhang}, \bibinfo{person}{Michael~F.
  Cohen}, {and} \bibinfo{person}{Brian Curless}.}
  \bibinfo{year}{2016}\natexlab{}.
\newblock \showarticletitle{Emptying, Refurnishing, and Relighting Indoor
  Spaces}.
\newblock \bibinfo{journal}{{\em ACM Transactions on Graphics\/}}
  \bibinfo{volume}{35}, \bibinfo{number}{6} (\bibinfo{year}{2016}).
\newblock


\bibitem[\protect\citeauthoryear{Zhou, Kr\"{a}henb\"{u}hl, and Efros}{Zhou
  et~al\mbox{.}}{2015}]%
        {zhou2015intrinsic}
\bibfield{author}{\bibinfo{person}{Tinghui Zhou}, \bibinfo{person}{Philipp
  Kr\"{a}henb\"{u}hl}, {and} \bibinfo{person}{Alexei~A. Efros}.}
  \bibinfo{year}{2015}\natexlab{}.
\newblock \showarticletitle{Learning Data-driven Reflectance Priors for
  Intrinsic Image Decomposition}.
\newblock \bibinfo{journal}{{\em International Conference on Computer
  Vision\/}} (\bibinfo{year}{2015}).
\newblock


\end{thebibliography}
\end{document}